\documentclass{article} 
\usepackage{arxiv,times}

\usepackage{math_commands}
\usepackage[T1]{fontenc}
\usepackage[clock]{ifsym}

\usepackage[utf8]{inputenc} 
\usepackage[T1]{fontenc}    

\usepackage{amsfonts}       
\usepackage{amsmath}
\usepackage{amssymb}
\usepackage{amsthm}
\usepackage{array}
\usepackage{bm}
\usepackage{booktabs}       
\usepackage{caption}
\usepackage{colortbl}
\usepackage{empheq}
\usepackage{enumitem}
\usepackage{etoolbox}
\usepackage{float}
\usepackage{makecell}
\usepackage{mathrsfs}
\usepackage{mdframed}
\usepackage{microtype}      
\usepackage{multirow}
\usepackage{nccmath}
\usepackage{nicefrac}       
\usepackage{pifont}
\usepackage{ragged2e}
\usepackage{rotating}
\usepackage{subcaption}
\usepackage{tabularx}
\usepackage{tikz}
\usepackage{times}
\usepackage{url}            
\usepackage{wrapfig}
\usepackage{xcolor, color, colortbl}
\usepackage{xspace}
\usepackage{thm-restate}

\newcommand{\benchname}{\textsc{TiC}-DataComp}
\newcommand{\cyfcc}{\textsc{TiC}-YFCC} 
\newcommand{\credcaps}{\textsc{TiC}-RedCaps}
\newcommand{\datacomp}{DataComp}
\newcommand{\mmedium}{\texttt{medium}}
\newcommand{\mlarge}{\texttt{large}}
\newcommand{\mxlarge}{\texttt{xlarge}}
\newcommand{\retreival}{{Retrieval}}
\newcommand{\imagenet}{{Net}}
\newcommand{\Oracle}{{Oracle}}
\newcommand{\Cumulative}{{Cumulative}}
\newcommand{\Sequential}{{Sequential}}
\newcommand{\Restart}{{Restart}}
\newcommand{\Patching}{{Patching}}
\newcommand{\Full}{{All}}
\newcommand{\Equal}{{Equal}}
\newcommand{\Exponential}{{Exp}}
\newcommand{\LwF}{LwF}
\newcommand{\constcosine}{{Const-Cosine}}

\definecolor{colorYes}{RGB}{51,160,44}
\definecolor{colorNo}{RGB}{228,26,28}
\newcommand{\xmark}{\textcolor{colorNo}{\ding{55}}}
\newcommand{\cmark}{\textcolor{colorYes}{\ding{51}}}

\usepackage{hyperref}
\usepackage{url}

\usepackage[capitalize]{cleveref}
\crefname{section}{Sec.}{Secs.}
\Crefname{section}{Section}{Sections}
\Crefname{table}{Table}{Tables}
\crefname{table}{Tab.}{Tabs.}

\title{{\normalsize \showclock{0}{45}} TiC-CLIP: Continual Training of CLIP Models} 

\author{Saurabh Garg$^\ddagger$\thanks{Work done during an internship at Apple.}\And
Mehrdad Farajtabar$^\dagger$ \And
Hadi Pouransari$^\dagger$\And
Raviteja Vemulapalli$^\dagger$ \And
Sachin Mehta$^\dagger$ \And
Oncel Tuzel$^\dagger$ \And
Vaishaal Shankar$^\dagger$ \And
Fartash Faghri$^\dagger$ \AND
\vspace*{-0.7cm}\\
$^\dagger$Apple $^\ddagger$Carnegie Mellon University \\
\texttt{sgarg2@andrew.cmu.edu}, \texttt{fartash@apple.com}
}

\usepackage[textwidth=2.5cm]{todonotes}

\newcommand{\update}[1]{\textcolor{black}{#1}}

\iclrfinalcopy 
\begin{document}

\maketitle

\vspace{-10pt}

\begin{abstract}
    Keeping large foundation models up to date on latest data is inherently expensive. To avoid the prohibitive costs of constantly retraining, it is imperative to \emph{continually} train these models. 
This problem is exacerbated by the lack of any large scale continual learning benchmarks or baselines.
We introduce the first set of web-scale
Time-Continual (TiC) benchmarks 
for 
training vision-language models: \benchname{}, \cyfcc{}, and \credcaps{}. \benchname{}, our largest dataset, contains over 12.7B timestamped image-text pairs spanning 9 years (2014--2022).
We first use our benchmarks to curate various \emph{dynamic} evaluations to measure temporal robustness of existing models. We show OpenAI's CLIP (trained on data up to 2020) loses $\approx 8\%$ zero-shot accuracy on our curated retrieval task from 2021--2022
compared with more recently trained models in OpenCLIP repository. 
We then study how to efficiently train models on time-continuous data.
We demonstrate that a simple rehearsal-based approach that continues training from the last checkpoint 
and replays old data reduces compute by $2.5\times$ when compared to the standard practice of retraining from scratch\footnote{Code is available at \url{https://github.com/apple/ml-tic-clip}.}. 

\vspace{-10pt}

\end{abstract}


\vspace{-5pt}
\section{Introduction}
\vspace{-3pt}

Large multimodal foundation models~\citep{bommasani2021opportunities} have
offered unprecedented advancements in image-generation %
and zero-shot generalization,
and have led to a paradigm shift in 
multimodal learning, e.g., CLIP~\citep{radford2021learning}, Flamingo~\citep{alayrac2022flamingo},
and Stable Diffusion~\citep{rombach2022high}. 
These foundation models are typically trained on large web-scale datasets which are fixed and \emph{static} in nature. For example, CLIP's training data contains
400 million image-text pairs, and Stable Diffusion was trained on LAION-2B dataset~\citep{schuhmann2022laion}.  
In reality, however, these models must operate in a \emph{dynamic} environment, where the world is in a state of constant change. For instance, the internet continually evolves, with petabytes of new data being added daily~\citep{wenzek2019ccnet,fbblog}.
It remains unclear how legacy models, e.g., OpenAI's CLIP models which 
were trained on internet-scale data up until 2020, work on future data
and whether they even require any re-training 
to adapt to time-evolving data.

\begin{figure}[t!]
    \centering
    \begin{subfigure}{0.63\linewidth}
        \centering
        \includegraphics[width=\linewidth]{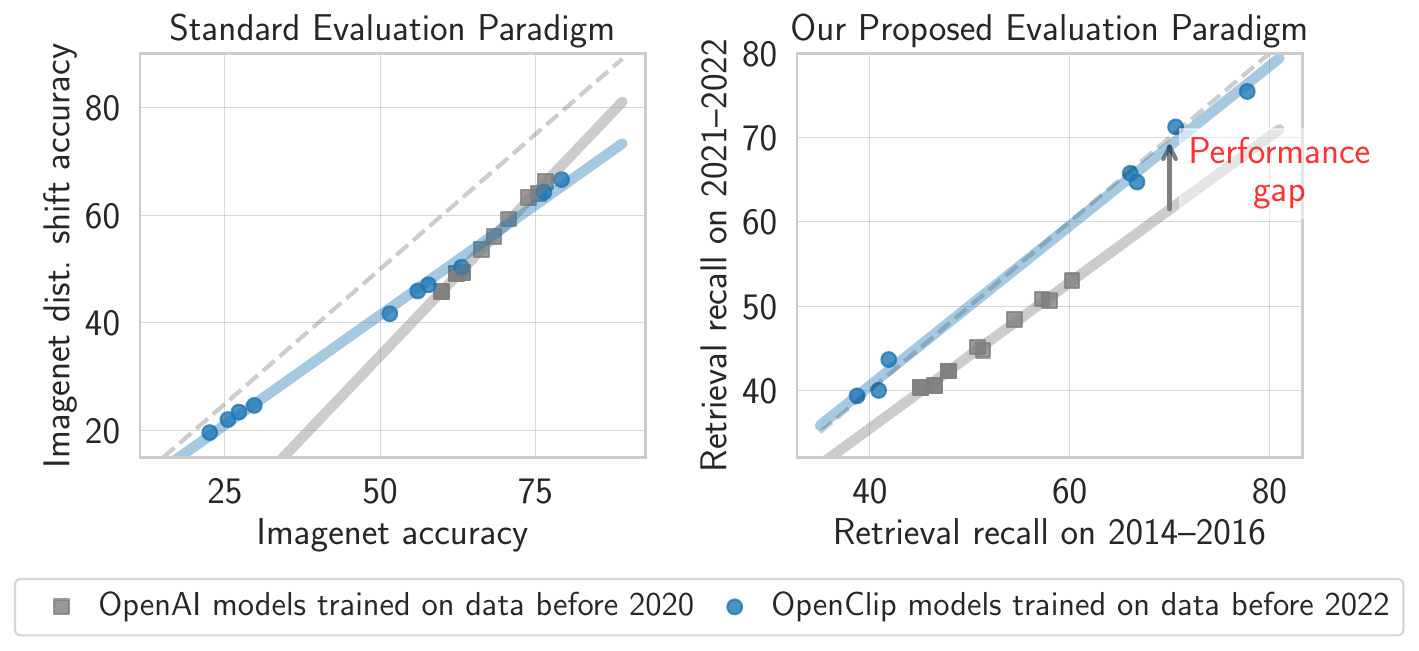}
        \label{fig:eval_openai}
    \end{subfigure}
    \begin{subfigure}{0.35\linewidth}
        \centering
        \includegraphics[width=\linewidth]{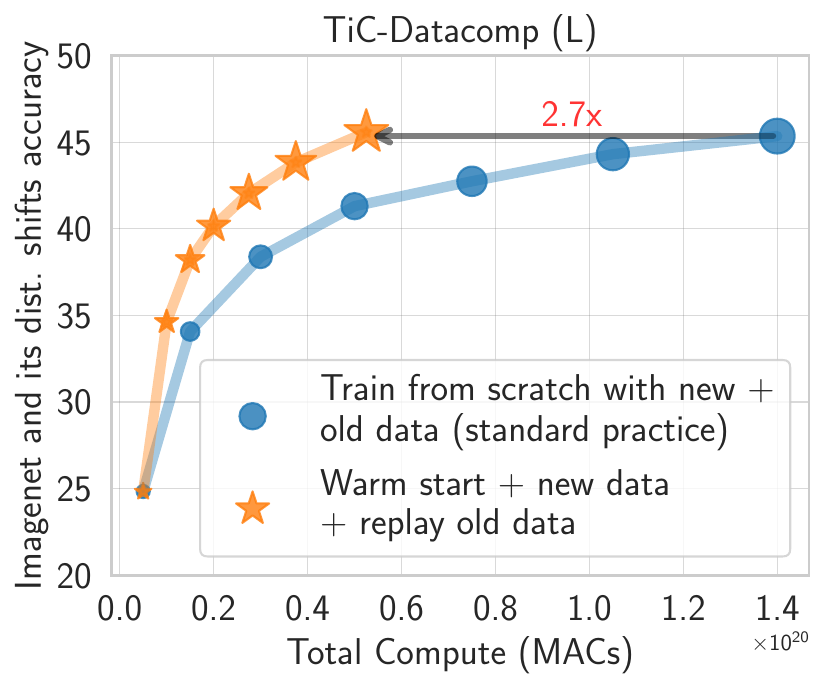}
        \label{fig:CT_large}
    \end{subfigure}
    \vspace{-13pt}
    \caption{
    \emph{(Left, Middle)} 
    \textbf{OpenAI models show less zero-shot robustness on retrieval task from 2021--2022.} OpenCLIP models and OpenAI models have similar robustness on standard benchmarks.
    However, OpenAI models show less robustness on our retrieval task when compared with recent models in OpenCLIP repository, highlighting susceptibility to a time-evolving data distribution
    \emph{(Right)} \textbf{Simple continual training baseline is computationally efficient and competitive to retraining from scratch.} Different points denote models trained sequentially on our \benchname{} (L) as data arrives over time. Warm start training with previous checkpoint and replaying all old data, performs similar to \Oracle{} which trains from scratch every time new data arrives, by using $2.7\times$ less compute. }
    \vspace{-13pt}
    \label{fig:intro}
\end{figure}

We begin by comparing robustness of OpenAI's CLIP models to 
others in OpenCLIP repository that are trained on 
more recently curated web-datasets (e.g., LAION-5B, DataComp) containing 
data up until 2022~\citep{ilharco2021openclip}. 
Since there is no existing benchmark to understand robustness to time-evolving vision-language data, 
we curate \emph{dynamic} classification and retrieval tasks for years 2014--2022   
and evaluate different CLIP models (see \secref{subsec:dynamic_eval_datasets} 
for our evaluation tasks). 
We make an intriguing observation that
OpenAI models exhibit a significant gap in retrieval performance
on data from 2021--2022 compared with 2014--2016
whereas OpenCLIP models retain their performance.
In contrast, standard evaluations such as accuracy on ImageNet distribution shifts paint an incomplete picture that OpenAI's CLIP models are 
slightly more robust than OpenCLIP models (\figref{fig:intro}). 
Our findings not only demonstrate the critical need for models to adapt 
and evolve alongside dynamic data distributions, but
also underscores the limitations of relying solely on static benchmarks (e.g. ImageNet).

One naive but common practice for adapting to time-evolving data 
is to train a new CLIP model from \emph{scratch} every time we obtain a new pool of image-text data. 
This practice has its rationale: 
initiating training from a pre-existing model can make it difficult to change the model's behavior in light of new data~\citep{ash2020warm, achille2018critical, liu2023knowledge}.  
However, training foundation models from scratch demands significant computational resources and is often infeasible to repeat frequently.
For example, ViT-g-14 in \citet{schuhmann2022laion, cherti2022reproducible} was trained for 240K A100 GPU hours which is approximately one month on 400 GPUs.
The prevailing training guidelines centered around  scaling laws for CLIP training 
have only looked at training from scratch~\citep{cherti2023reproducible}. 
This leads to a pivotal question: \emph{How can we continuously update models as the data distribution evolves over time given computational constraints?}

There exists a vast literature on continual learning, 
with a focus on adapting models to dynamic environments~\citep{parisi2019continual,hadsell2020embracing,de2021continual}.
Traditionally, this field concentrated on synthetic incremental benchmarks that 
lack natural evolution between tasks, and hence, continual learning methods 
are seldom used in real-world scenarios~\citep{cossu2022class,lin2021clear}. 
In contrast, recent works focusing on continual 
learning methods for CLIP models, 
primarily target improving performance on a single or a sequence 
of disjoint downstream tasks~\citep{ding2022don,zhou2023learning,zheng2023preventing, ilharco2022patching}.
While some recent works have started to 
address these problems, existing benchmarks 
are comparatively much smaller in scale, or lack paired image-text data~\citep{ni2023continual,lin2021clear}. 
Simply put, there is a scarcity of work focusing on continual training of CLIP models on naturally evolving data with time at web-scale.

We take the first step towards \textbf{Time-Continual (\textsc{TiC})} training of CLIP models 
where data distribution evolves naturally over time (overview
in \cref{fig:setup-figure}). We introduce 
\benchname{}, a new benchmark for Time-Continual training of CLIP models, 
which we create by appending ``crawl time'' information to  existing CommonPool dataset~\citep{gadre2023datacomp}. 
We also repurpose other 
web-scale datasets gathered from diverse sources, such as Reddit and Flickr.
Specifically, we curate \cyfcc\ and \credcaps\ 
by leveraging time information available in YFCC~\citep{thomee2016yfcc100m} 
and Redcaps~\citep{desai2021redcaps} respectively.
The primary objective of our study on this benchmark is to develop continual learning 
methods that operate within a constrained computational budget 
(say $C$) each time a fresh batch of data becomes available.
These methods compete with an \Oracle, which 
starts training from scratch every time new data 
arrives, utilizing a cumulative computational budget.

To assess models trained in our \textsc{TiC}-CLIP
framework, we
evaluate models on our proposed dynamic evaluation tasks that evolve with time along with 
28 standard classification and retrieval tasks including ImageNet~\citep{krizhevsky2012imagenet}, 
ImageNet distributions shifts, 
and Flickr~\citep{plummer2015flickr30k},
in a zero-shot manner following the work of \citet{gadre2023datacomp, radford2021learning}.

Finally, we develop continual learning methods on our benchmarks and perform
over two hundred experiments
with different baselines 
that utilize previous checkpoints (e.g., warm start, patching, and distillation), 
replay buffers, and learning rate schedules.
Our findings highlight a key takeaway: \Cumulative{} method that warm starts training with the latest 
checkpoint and replays all old data, achieves performance competitive to an \Oracle{} 
while being $2.7\times$ computationally more efficient. 
Additionally, our experiments demonstrate interesting trade-offs
between buffer sizes for static and dynamic performance 
and provide valuable insights into learning rate schedules for sequential training.
Our results span over various dataset scales (from 11M samples to 3B)
and highlight trends with different methods that are largely consistent across scales.

To make our benchmarks accessible, we publicly release the code and the time information we
collect on top of existing datasets \href{https://github.com/apple/ml-tic-clip}{here}. 
Our work is just an initial step towards continual training of foundation models, 
and we believe our research would spur more attention to this understudied area. %


\vspace{-5pt}
\section{TiC-CLIP: Benchmarks and Experimental Protocol} \label{sec:benchmarks}
\vspace{-3pt}

\begin{figure*}[!t]
    \centering
    \includegraphics[width=\linewidth]{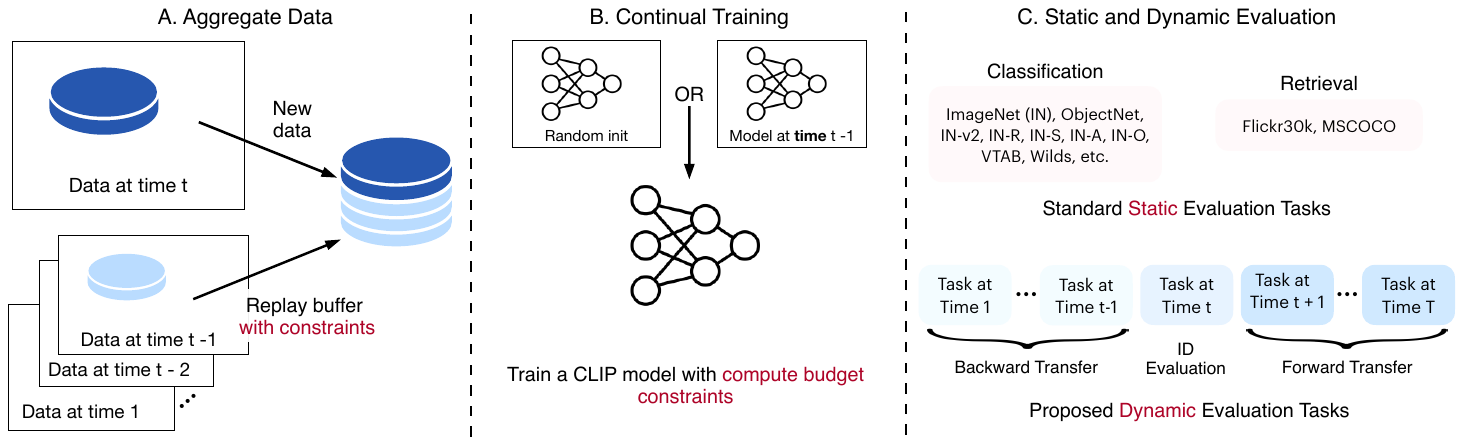}
    \vspace{-15pt}
    \caption{\textbf{Experimental protocol on our proposed continual benchmarks.} \emph{(A)}  Combine
    new and old data
    given buffer constraints. \emph{(B)} Continually train a model with a compute budget (say $C$) either by starting with previous checkpoint or from scratch. \emph{(C)} Evaluate models on standard datasets and our proposed dynamic datasets.
    Comparison with other benchmarks in
    \cref{app:cl_benchmarks}. 
    }
    \label{fig:setup-figure}
    \vspace{-10pt}
\end{figure*}

In this section, we introduce our benchmark (\cref{fig:setup-figure}) focusing on the training of a vision-language foundation model with the
Contrastive Language Image Pretraining (CLIP)~\citep{radford2021learning}) objective.  
Notably, we train on image-text data that arrives sequentially unlike the 
conventional image-text datasets which are static (e.g.  
WiT in CLIP, DataComp in \citet{gadre2023datacomp}).
We curate \benchname{}, 
\cyfcc{}, and \credcaps{} that are image-text pairs sourced from the internet
which we augment with auxiliary time information.  We also introduce dynamic 
evaluation tasks to assess performance of our continually trained models on 
data evolving with time.
The goal of a learner is 
to train a \emph{deployable} model at each step
as new data becomes available with a fixed compute budget.

\vspace{-3pt}
\subsection{Benchmark Design: How we Create Time-Continual Datasets?} \label{subsec:benchmark_design}
\vspace{-3pt}

To instantiate continual training of CLIP, we extend existing image-text 
datasets with time information collected from the original source of the 
datasets.  Our largest dataset is \benchname{} which contains 
12.7 billion image-text pairs with ``crawl-time'' metadata.  We create this 
   dataset on top of the existing DataComp benchmark~\citep{gadre2023datacomp}. 
We also create \cyfcc{} and \credcaps{} on top of existing 
YFCC15M~\citep{thomee2016yfcc100m, radford2021learning} and 
Redcaps~\citep{desai2021redcaps} datasets to highlight that our findings are 
broadly applicable to carefully curated datasets from diverse sources such as 
Reddit and Flickr. 
While time-related metadata is absent in the DataComp benchmark, it is 
available in the original releases of YFCC and Redcaps. Nevertheless, to the 
best of our knowledge, no prior work utilizes such time information 
for continual training of CLIP models.
We show dataset statistics for all datasets, e.g., number of examples in each year in 
\appref{app:dataset_stats}.

\textbf{\benchname{} {} {}} We collect timestamps for the 
CommonPool dataset introduced in \datacomp{} which contains 12.7B
image-text  pairs (not including 0.1B inaccessible ones).
This dataset stands as the largest public 
image-text dataset to date.  
The source of \datacomp{} is Common Crawl, which periodically 
releases web-crawled data snapshots, typically on a monthly basis since 2014 
with new and updated webpages.
To construct \benchname{}, we augment each image-text pair in 
\datacomp{} with their \emph{first} timestamp.
We followed the same 
construction process as \datacomp{} but retained only the 
image-text pair found in the earliest snapshot during the deduplication stage.  
This process provides timestamps at the granularity of 
months, spanning years 2014--2022. See \appref{app:tic_datacomp_construction} for details on the 
construction process.
We note that while this augmented time information may contain some noise, on 
average, we find it to be a reasonably accurate proxy for the upload time of 
web pages (see \appref{app:tic_datacomp_construction}).

Although our benchmark contains time information at the granularity of 
months, we limit our experiments to granularity of years by consolidating data for all months in a year.
Similar to \datacomp, our benchmark has an inclusive design, 
accommodating participants with varying levels of computational resources. In 
particular, we experiment with \mmedium, \mlarge, and \mxlarge{} sizes from 
CommonPool.
\cite{gadre2023datacomp} leverage different filtering strategies to select the training subset. 
We are concerned that filtering techniques bias the selected training data.
In \appref{app:tic_datacomp_filtering}, we provide preliminary evidence 
that ``Bestpool'' filtering that uses off-the-shelf CLIP models, indeed biases the selected data to old time steps.
Nevertheless, to highlight significance of our findings even for state-of-the filtering techniques, we experiment with both Bestpool and Basic filtering (no CLIP filtering) at \mxlarge{} scale.
For \mlarge{} and \mmedium{} scales, we only experiment with Basic filtering.

\textbf{\cyfcc{} {} {}} We experiment with the 15M subset of 
YFCC100M~\citep{thomee2016yfcc100m}, namely YFCC15M, selected by 
OpenAI~\citep{radford2021learning}. This filtering retains only images with 
natural text in captions. YFCC100M contains data from years 2008--2014 and 
was originally released with upload timestamps. We 
use this information to create continual splits at the granularity of 
years. 

\textbf{\credcaps{} {} {}} RedCaps contains 12M image-caption pairs 
from manually curated set of subreddits across 
$2011$--$2020$~\citep{desai2021redcaps}.  We use the creation timestamps of the posts to create splits for continual learning. Similar to the other two 
datasets, we experiment at the granularity of years.

\vspace*{-6pt}
\subsection{Evaluation Testbed} \label{subsec:dynamic_eval_datasets}
\vspace*{-4pt}

\begin{figure*}[!t]
    \centering
    \includegraphics[width=\linewidth]{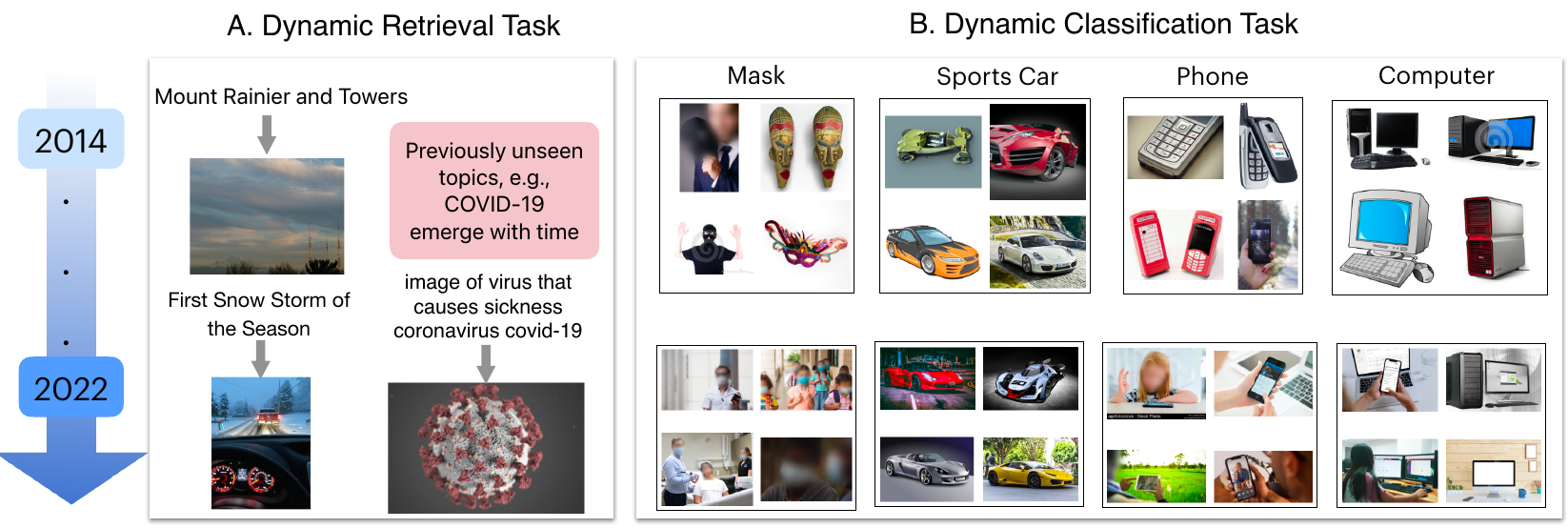}
    \vspace{-15pt}
    \caption{\textbf{Distribution of examples changes from 2014 to 2022 in our dynamic evaluation tasks.} 
    \emph{(Left)} Samples for text to image retrieval. For new timestamps, images from novel concepts appear (e.g., COVID-19). \emph{(Right)} Samples from our classification task for 4 categories. We observe that not only objects evolve over time but also images from recent timestamps are captured more in the wild.}
    \vspace{-10pt}
    \label{fig:eval_datasets}
\end{figure*}

\textbf{Dynamic tasks{} {} {}} We leverage the temporal information in 
our benchmarks to create \emph{dynamic evaluation} tasks.
Here, the test data 
comprises samples varying over years as the world evolved.
For our largest dataset which is 
\benchname{}, we create dynamic tasks for both retrieval and classification as 
described below. (examples in \Cref{fig:eval_datasets}
and additional examples in \appref{app:evaluation_examples}): 

I. \emph{Dynamic retrieval task}: To create a retrieval task, we sample 
a batch of IID image-text pairs 
from different timestamps and evaluate text retrieval performance given the 
corresponding image (similarly, image retrieval given the corresponding text). We 
refer to the dataset as \benchname{}-\retreival{}.

II. \emph{Dynamic classification task}: We also create a classification dataset 
\benchname{}-\imagenet{} with ImageNet classes from CommonPool and 
augmented with timestamps.
Inspired by LAIONNet~\citep{shirali2023makes},
we first filter examples where the 
corresponding caption contains one and only one of the synsets of ImageNet.  
Then we only retain examples where the similarity between ImageNet synset 
definition and the caption exceeds a threshold of $0.5$. We evaluate the 
similarity using an off-the-shelf sentence embedding 
model~\citep{reimers-2019-sentence-bert}.  Crucially, unlike LAIONNet, we do 
not filter the image-text pairs with CLIP similarity scores to avoid biasing 
the selection process. 
We describe the construction process in more details in 
\appref{app:evaluation_examples}.
On \benchname{}-\imagenet{}, 
we report average accuracy over all classes and over selected nodes (e.g., 
motor vehicles) at each time step. 

Similarly, we create 
retrieval tasks for \cyfcc{} and \credcaps{}.
Note that we remove the extracted image-text pairs for dynamic retrieval and 
classification tasks from the training sets.
Evaluations on dynamic tasks are done 
in a zero shot manner. 

\textbf{Static tasks{} {} {}}  We also evaluate models on numerous 
classification and retrieval tasks in a zero-shot manner as in 
\citet{radford2021learning}. 
In particular, we consider 28 standard tasks: 27 image classification tasks, e.g., 
ImageNet
and its 6 distribution shifts (e.g., 
ImageNetv2, ImageNet-R, ImageNet-Sketch, and Objectnet), datasets from VTAB and 
Flickr30k retrieval task. We refer to these as \emph{static evaluation} tasks.  
We list all the datasets in \appref{app:static_datasets}. 

\textbf{Evaluation metrics{} {} {}}
We define metrics for classification tasks and retrieval tasks based on \emph{accuracy} and \emph{Recall@1}, respectively. Let $T$ represent the number of time 
steps for which we have data.
For each training 
method, we generate a total of $T$ models, each corresponding to the end of 
training at a particular time step. 
For static datasets (e.g., ImageNet), we report average performance of $T$ 
models. However, when dealing with dynamic evaluation datasets, we assess the 
performance of each of the $T$ models on evaluation datasets collected at all 
time steps. Consequently, for each model and a dynamic evaluation task, we 
obtain $T$ performance values. We represent these values using the performance 
matrix $\calE$, where each entry $\calE_{i,j}$ signifies the performance of the 
model obtained after observing training data at time step $i$ when evaluated on 
a dataset from time step $j$.  The performance matrix $\calE$ can also be 
succinctly summarized using three standard metrics commonly employed in 
continual learning  evaluations~\citep{lin2021clear,diaz2018don}:
\vspace*{-2.5mm}
\begin{itemize}[leftmargin=*]
\item \emph{In-domain performance}: average performance at each training time step (i.e., the diagonal of $\calE$)%
\item \emph{Backward transfer}: average on time steps before each training step (i.e., the lower triangular of $\calE$)%
\item \emph{Forward transfer}: average on time steps following each 
    training step (i.e., the upper triangular of $\calE$)%
\end{itemize}
\vspace*{-1.5mm}

Sometimes, the metrics described above can cause the backward transfer metric to be influenced by later evaluation time steps, biasing the backward transfer metric (refer to \appref{app:updated_metrics} for details). Therefore, in \appref{app:updated_metrics}, we present results using revised metrics that mitigate this issue.

While the static tasks capture performance on standard benchmarks, dynamic 
tasks capture problems due to distribution shift (for forward transfer) and 
forgetting (for backward transfer). The goal in our benchmark is to 
develop continual learning methods that maximize performance on static tasks 
while simultaneously optimizing for performance on dynamic tasks.

\vspace{-6pt}
\subsection{Experimental Protocol For Training} \label{subsec:exp_protocol}
\vspace{-3pt}

\textbf{Streaming protocol {} {}} We follow a streaming protocol, where data is 
progressively revealed to the learner
in large batches
with the objective of achieving 
a deployable model as early as possible after each batch arrives.
We conduct experiments with data streaming at the granularity of years
and our benchmark supports future research at the granularity of months.
Additionally, as the amount of data from earlier time 
steps is limited (see \appref{app:dataset_stats}), we aggregate data from the earlier time steps into a single larger batch and timestamp it by the latest year in the range.
After this aggregation, we have 7 time steps for \benchname{} 
(2016--2022) and 4 for both \cyfcc{} (2011--2014) and \credcaps{} (2017--2020).  
While the number of image-text pairs revealed at each time step are of similar 
orders of magnitude, the exact number does vary across steps
and we do not 
artificially alter the sizes. 

\textbf{Memory budget {} {}} We allow methods to use the last model 
checkpoint at each step as the cost of keeping one checkpoint per month is 
often negligible. In contrast, the cost of retaining 
old data can be high and might not be permitted due to data expiration 
policies. Thus, along with studying methods that retain all old data, we also 
explore strategies that restrict data persistence (see 
\secref{sec:methods} for details).

\textbf{Compute budget{} {} {}} To ensure a fair comparison between methods, we establish a consistent total compute budget, quantified in terms of Multiply-Accumulate Operations (MACs), and allocate it evenly for training at every time step.
Unless specified otherwise, for all methods except 
\Oracle{} and \LwF{},
we use the same compute budget. For 
experiments on \benchname, we refer to compute configurations in 
\textsc{DataComp} for overall compute. For \credcaps{} and \cyfcc{}, we use 
compute of order \mmedium\ scale in \benchname. 
Compute budget details are in \appref{app:compute_constraints}.

\vspace{-6pt}
\subsection{Analyzing Distribution Shifts in the Constructed Benchmarks}
\vspace{-3pt}

\textbf{\benchname{} analysis through the lens of constructed evaluation tasks {} {}}
First, we qualitatively analyze the examples in our retrieval and classification dataset (\figref{fig:eval_datasets}). 
We observe that over time, in the retrieval task, 
new concepts like COVID-19 emerge. Likewise, certain
ImageNet classes evolve, such as the shift from ``masquerad'' masks to ``surgical/protective'' 
masks in their definitions. Moreover, as time evolves, we observe that  
image quality improves and more images tend to appear in the wild 
in contrast to centered white background images.
Next, we compare performance of 
OpenAI and OpenCLIP models on our datasets.
Here, we only present the main findings, 
and delegate a detailed discussion to \appref{app:dist_shift_analysis}.  
We observe a 
significant performance gap between 
OpenAI and OpenCLIP models 
on our dynamic retrieval task (\figref{fig:intro}). 
This gap widens notably on retrieval queries where captions mention COVID-19.
On the other hand,  OpenAI and 
OpenCLIP models exhibit similar robustness for 
retrieval on data coming from Flickr highlighting 
that data from some domains do not exhibit shifts 
that cause performance drops.
For our classification task, we observe a very small drop ($\approx 1\%$) when averaged across all categories.
However, we observe a substantial gap 
on specific subtrees in ImageNet. For example,  
classes in ``motor vehicle'' subtree 
show an approximate $4\%$ performance drop, when comparing OpenAI 
and OpenCLIP models. These findings highlight that while overall ImageNet classes 
may remain timeless, certain categories tend to evolve faster than others.  
Our qualitative and quantitative analysis on \benchname{} 
clearly highlights evolution of distributions and captures different properties 
than standard benchmarks.

\textbf{Quantitative analysis on \cyfcc{} {} {}}
We analyze \cyfcc{} using off-the-shelf sentence and image encoders.
We first embed images from different time steps with an 
OpenAI CLIP encoder and then compute Frechet Inception Distance (FID; \citet{Seitzer2020FID}). 
As time progresses, we observe that FID distance increases 
with respect to data from first time step (\figref{fig:yfcc_dist_shift} in \appref{app:dist_shift_analysis}). 
Similarly, we use pretrained sentence transformer to 
extract top-5 categories from Wordnet Nouns  
for each caption. We observe that the TV distance 
over distribution of WordNet Nouns evolves over time
when compared to data from the first time step. More details in \appref{app:dist_shift_analysis}.

\vspace{-8pt}
\section{TiC-CLIP: How to Continually Train CLIP Models?} \label{sec:methods}
\vspace{-3pt}
\begin{wrapfigure}{r}{0.5\textwidth}
    \vspace*{-15pt}
    \captionof{table}{Table summarizing our methods. $D$: data size in each step, $T$ total time steps,
$t$: current time step, $C$: compute budget (iterations).} 
    \vspace*{-8pt}
\resizebox{0.5\textwidth}{!}{%
\begin{tabular}{lccc|c}
\toprule[1.2pt]
\multirow{2}{*}{\textbf{Method}} & \multicolumn{3}{c}{\textbf{Each Step}} & \textbf{Total}\\
\cmidrule{2-4}
 & \textbf{Train Size} & \textbf{Init.} & \textbf{Compute} & \textbf{Compute}\\
\midrule
\Cumulative{}-\Full{}                              & $tD$   & Last       & $C$   & $TC$  \\
\Cumulative{}-\Exponential{}                       & $2D$   & Last       & $C$   & $TC$  \\
\Cumulative{}-\Equal{}                             & $2D$   & Last       & $C$   & $TC$  \\
\Sequential{}                                      & $D$    & Last       & $C$   & $TC$  \\
\Restart{}                                         & $tD$   & Rand       & $C$   & $TC$  \\
\Patching{}                                        & $D$    & Last Patch & $C$   & $TC$  \\
\rowcolor{gray!10} \LwF{}                          & $D$    & Last       & $1.2 \times  C$  & $1.2 \times TC$  \\
\rowcolor{gray!10} \Oracle{}$^{**}$                & $tD$   & Rand       & $tC$  & $\frac{(T+1)T}{2}C$\\
\bottomrule[1.2pt]
\end{tabular}
}    
    \label{tab:methods}
    \vspace*{-18pt}
\end{wrapfigure}

In this section, we lay out different methods specifically focus on the following questions (\cref{tab:methods}):
(i) How to utilize/replay data from previous time steps;
(ii) How to leverage previously trained model checkpoints?
(iii) What should be the training/optimization procedure?

Data replay methods initialized from the last checkpoint 
demonstrate strong performance on standard continual learning benchmarks (\cref{sec:related}).  
We consider replay methods with/without initialization from last 
checkpoint(s):

I. \textbf{\Oracle{}}: Train a CLIP model from scratch (i.e., random initialization) on all image-text data received till time 
$t$ using a large compute budget of $t\times C$.
\Oracle{} represents a \emph{prohibitively expensive} 
method that is the most common practice in training large-scale foundation models.  The goal of other methods is to perform as close as possible to 
the \Oracle{} within their limited budget.

II. \textbf{\Cumulative{}}: Train each model initialized from last checkpoint on the union of all data up to $t$ with compute budget $C$.
This method is 
analogous to experience replay~\citep{robins1995catastrophic,hayes2019memory} but with substantially 
larger buffers than common in the continual learning literature.
Given a fixed buffer size for each past step, we observe minimal to no 
difference between random subsampling and other strategies. 
After 
sampling the replay data, we randomly shuffle it together with new data for training.
We consider the following strategies for sampling buffer sizes per step:
\vspace*{-2mm}
\begin{itemize}[leftmargin=*]
    \itemsep0em
    \item \textbf{-\Full{}}: Replay all previous data.
    \item \textbf{-\Exponential{}}:
    Replay a buffer of size $D$ and reduce the amount of old data by half at each step.
    \update{For example, at 3-rd time step, 
    we retain $D/2, D/2$ of old data and at 4-th, we retain $D/4, D/4, D/2$ of old data. 
    Along with $D$ data from current step, this method trains on at most $2D$ data in each step.}
    \item \textbf{-\Equal{}}: Replay a buffer of size $D$ but split the buffer equally among all
    previous years. \update{For example, at 4-th step, we retain $D/3, D/3, D/3$ of old data.
    Along with $D$ data from current time step, this method trains on at most $2D$ data in each step.}
\end{itemize}
\vspace*{-2mm}

III. \textbf{\Sequential{}}: Train \emph{only} on the new data starting from the best checkpoint of the previous time step. \Sequential{} is similar to \Cumulative{} but without any replay buffer.  

IV. \textbf{\Restart{}}: Train each model from scratch (i.e., random initialization) on all the data till time $t$ for compute budget $C$.
\Restart{} is similar to the \Oracle{} but with compute budget $C$ at each time step and
similar to \Sequential{} but with random initialization.
As such, \Restart{} helps us understand the \emph{forward transfer}
and \emph{loss of plasticity} in our benchmark~\citep{ash2020warm, dohare2023loss}.

V. \textbf{\Patching}: We use sequential patching from \citet{ilharco2022patching}. Initialize from a patched model of last step and train only on the new data.
To obtain a patched model at each time step, we apply weight interpolation with the patched model (if any) trained at time step $t-1$ and the model trained at time step $t$.
We tune the mixing coefficients by optimizing average retrieval performance on previous tasks.

VI. \textbf{\LwF{}}: Train only on the new data with a KL divergence penalty between the image-text similarity matrix of last checkpoint and current model on each batch~\citep{li2017learning,ding2022don}. See \appref{app:other_methods} for results with other continual learning methods, e.g., EWC~\citep{kirkpatrick2017overcoming}.

\textbf{Learning rate schedule {} {}}
The defacto Learning Rate (LR) schedule for training CLIP models is an initial linear increase to a maximum value, i.e., warm up, followed by a cosine decay
~\citep{radford2021learning, gadre2023datacomp}. 
We default to using a cosine LR schedule for each sequential run, resulting in a cyclic schedule and observe a significant increase in training loss early in subsequent runs when the LR is high.
However, as training progresses, we observe that the increased loss decreases at a faster rate (when compared to training from scratch) allowing us to train with cyclic schedules.  
We discuss this more and  explore an alternate learning rate schedule in \appref{sec:const_cosine_schedule}.

\textbf{}\textbf{Other Training details and hyperparameters {} {} {}} 
Unless specified otherwise, we closely follow the original CLIP training recipe~\citep{radford2021learning}. 
We train the CLIP variant with ViT-B/16 as the image encoder~\citep{dosovitskiy2020image}. 
All training and hyperparameters can be found in \appref{app:hyperparam_details}.

\vspace{-6pt}
\section{Experiments and Main Results}
\vspace{-4pt}

\begin{table}[t]
    \centering
    \caption{\textbf{Zero shot performance on our time-continual benchmarks.} 
    $^*$ and $^{**}$ denote methods that violate the compute budget.
    For static tasks, we tabulate accuracy of the models obtained on the final timestamp. 
    For dynamic tasks, we tabulate forward/backward transfer 
    and ID performance on retrieval tasks (\secref{subsec:exp_protocol}). 
    For \benchname{} (XL), we include results with Bestpool 
    filtering (basic filtering in \tabref{tab:basic_xlarge_result}).  
    For all metrics, higher is better. %
    }
    \vspace*{-5pt}
\resizebox{1.0\textwidth}{!}{%
\begin{tabular}{lcccccccccc}
\toprule[1.2pt]
\multirow{3}{*}{\textbf{Benchmark}} & \multirow{3}{*}{\textbf{Method}} &   \multirow{3}{*}{\parbox{4em}{\centering \textbf{Compute} (MACs)}} & \multicolumn{4}{c}{\textbf{Static Tasks}} &  {} & \multicolumn{3}{c}{\textbf{Dynamic Retrieval Tasks}} \\[3pt]
{} & {} & {} & \multirow{2}{*}{\parbox{4em}{ImageNet}} & \multirow{2}{*}{\parbox{4em}{ImageNet dist. shift}} & \multirow{2}{*}{\parbox{4em}{Flickr30k}} & \multirow{2}{*}{\parbox{5.5em}{\centering Average over 28 datasets}} & {} & \multirow{2}{*}{\parbox{4em}{\centering Backward Transfer}} & \multirow{2}{*}{\parbox{4em}{\centering ID Performance}} & \multirow{2}{*}{\parbox{4em}{\centering Forward Transfer}}  \\ 
{} & {} & \\
\midrule
\multirow{8}{*}{\textbf{\cyfcc{}}} & \Restart{} &  $3.4 \times 10^{18}$  & $5.2$ & $3.6$ & $3.0$ & $12.9$ &  & $13.2$ & $41.4$ & $18.6$
\\
& \Sequential{} & $3.4 \times 10^{18}$  &  $17.3$ & $10.5$ & $15.9$ & $21.9$ &  & $42.2$ & $48.4$ & $23.7$ \\
& \Patching{} & $3.4 \times 10^{18}$  & $18.9$ & $11.3$ & $18.5$ & $23.3$ &  & $44.7$ & $53.4$ & $24.5$ \\
& \Cumulative{}-\Exponential{} &   $3.4 \times 10^{18}$  & $24.1$ & $14.3$ & $20.4$ & $25.9$ &  & $60.4$ & $60.1$ & $27.1$ \\
& \Cumulative{}-\Equal &  $3.4 \times 10^{18}$ & $23.9$ & $13.8$ & $20.5$ & $26.3$ &  & $60.4$ & $60.4$ & $27.1$ \\
&  \Cumulative{}-\Full{} &   $3.4 \times 10^{18}$  & $\mathbf{29.3}$ & $\mathbf{17.6}$ & $\mathbf{26.8}$ & $\mathbf{29.6}$ &  & $\mathbf{66.4}$ & $\mathbf{60.2}$ & $\mathbf{27.6}$\\
\rowcolor{gray!10} & \LwF{}$^{*}$ & $4.1 \times 10^{18}$  & $16.9$ & $9.8$ & $14.7$ & $21.2$ &  & $36.6$ & $56.0$ & $23.2$ \\ 
\rowcolor{gray!10} &  \Cumulative{}-\Full{}$^{*}$ &   $3.6\times 10^{18}$ & $\mathbf{29.2}$ & $\mathbf{17.5}$ & $\mathbf{27.4}$ & $\mathbf{29.3}$ &  & $\mathbf{66.8}$ & $\mathbf{60.3}$ & $\mathbf{27.6}$ \\
\rowcolor{gray!10} & \Oracle{}$^{**}$ &  $8.5\times 10^{18}$ & $\mathbf{29.2}$ & $\mathbf{17.0}$ & $\mathbf{25.9}$ & $\mathbf{29.0}$ &  & $\mathbf{66.1}$ & $\mathbf{61.8}$ & $\mathbf{26.9}$
 \\
\midrule 
\multirow{8}{*}{\textbf{\credcaps{}}} 
& \Restart{} & $3.4 \times 10^{18}$ & $11.7$ & $8.5$ & $3.7$ & $18.4$ &  & $21.3$ & $25.4$ & $22.4$ \\
& \Sequential{} &  $3.4 \times 10^{18}$ & $19.3$ & $13.7$ & $6.2$ & $25.8$ &  & $33.0$ & $33.6$ & $27.5$ \\
& \Patching{} &  $3.4 \times 10^{18}$ & $21.3$ & $15.2$ & $7.7$ & $26.8$ &  & $34.8$ & $34.8$ & $27.8$ \\
& \Cumulative{}-\Exponential{} &  $3.4 \times 10^{18}$ & $27.3$ & $19.1$ & $10.5$ & $30.0$ &  & $44.5$ & $42.0$ & $32.6$ \\
& \Cumulative{}-\Equal &  $3.4 \times 10^{18}$ & $27.8$ & $19.4$ & $10.0$ & $30.5$ &  & $44.4$ & $42.0$ & $32.6$\\
&  \Cumulative{}-\Full{} & $3.4 \times 10^{18}$ &  $\mathbf{32.2}$ & $18.7$ & $14.5$ & ${31.7}$ &  & $\mathbf{48.9}$ & $\mathbf{43.2}$ & $\mathbf{33.4}$ \\
\rowcolor{gray!10} & \LwF{}$^{*}$ &  $4.1 \times 10^{18}$ & $21.6$ & $14.8$ & $8.2$ & $27.3$ &  & $35.4$ & $36.0$ & $28.4$ \\
\rowcolor{gray!10} &  \Cumulative{}-\Full{}$^{*}$ &  $3.6\times 10^{18}$ & $\mathbf{32.9}$ & $\mathbf{23.7}$ & $\mathbf{14.1}$ & $\mathbf{32.9}$ &  & $\mathbf{49.0}$ & $\mathbf{43.4}$ & $\mathbf{33.4}$ \\
\rowcolor{gray!10} & \Oracle{}$^{**}$ &  $8.5\times 10^{18}$  & $\mathbf{32.7}$ & $\mathbf{22.7}$ & $\mathbf{14.3}$ & $\mathbf{32.3}$ &  & $\mathbf{48.5}$ & $\mathbf{43.1}$ & $\mathbf{33.4}$\\
\midrule 
\multirow{6}{*}{\textbf{\benchname} (M)} 
& \Sequential{} &  $3.0 \times 10^{18}$  &$19.2$ & $16.4$ & $16.4$ & $26.0$ &  & $25.7$ & $26.4$ & $14.9$ \\
& \Patching{} &  $3.0 \times 10^{18}$  & $19.3$ & $16.8$ & $18.5$ & $26.4$ &  & $26.9$ & $25.4$ & $14.5$ \\
& \Cumulative{}-\Exponential{} &  $3.0 \times 10^{18}$ & $22.1$ & $18.4$ & $20.4$ & $28.8$ &  & $31.7$ & $27.1$ & $\mathbf{15.2}$ \\
& \Cumulative{}-\Equal &  $3.0 \times 10^{18}$  & $22.1$ & $18.4$ & $19.2$ & $28.0$ &  & $31.8$ & $26.8$ & $15.1$\\
&  \Cumulative{}-\Full{} &   $3.0 \times 10^{18}$ &$24.0$ & $20.2$ & $20.9$ & $30.0$ &  & $33.8$ & $26.4$ & $15.1$\\
\rowcolor{gray!10}  & \LwF{}$^{*}$ &  $3.8 \times 10^{18}$ &  $19.2$ & $16.5$ & $17.7$ & $27.0$ &  & $25.6$ & $26.6$ & $14.9$
 \\
\rowcolor{gray!10} &  \Cumulative{}-\Full{}$^{*}$ &  $3.9 \times 10^{18}$ & $\mathbf{30.0}$ & $\mathbf{25.0}$ & $\mathbf{28.6}$ & $\mathbf{35.1}$ &  & $\mathbf{36.7}$ & $\mathbf{28.3}$ & $\mathbf{15.5}$
\\
\rowcolor{gray!10} & \Oracle{}$^{**}$ &  $1.2 \times 10^{19}$ & $25.5$ & $21.2$ & $23.3$ & $30.8$ &  & $34.9$ & $27.8$ & $\mathbf{15.6}$\\
\midrule 
\multirow{5}{*}{\textbf{\benchname} (L)} 
& \Sequential{} & $2.7 \times 10^{19}$ & $44.7$ & $37.4$ & $48.4$ & $45.7$ &  & $52.6$ & $\mathbf{58.4}$ & $41.1$ \\
& \Patching{} & $2.7 \times 10^{19}$ & $45.8$ & $38.9$ & $49.7$ & $46.9$ &  & $55.2$ & $57.5$ & $40.9$ \\
& \Cumulative{}-\Exponential{} & $2.7 \times 10^{19}$ & $47.3$ & $39.6$ & $50.8$ & $47.6$ &  & $60.4$ & $\mathbf{58.4}$ & $\mathbf{41.4}$\\
& \Cumulative{}-\Equal & $2.7 \times 10^{19}$ & $47.7$ & $40.3$ & $51.8$ & $47.7$ &  & $60.9$ & $\mathbf{58.2}$ & $\mathbf{41.4}$\\
&  \Cumulative{}-\Full{} & $2.7 \times 10^{19}$ &    $48.9$ & $41.3$ & $50.9$ & $48.0$ &  & $62.1$ & $57.3$ & $41.2$\\
\rowcolor{gray!10} &  \Cumulative{}-\Full{}$^{*}$ &  $4.1\times 10^{19}$ & $53.0$ & $\mathbf{44.3}$ & $\mathbf{54.4}$ & $\mathbf{51.3}$ &  & $63.0$ & $57.8$ & $41.2$ \\
\rowcolor{gray!10}  &  \Oracle{}$^{**}$ & $1.1 \times 10^{20}$ & $\mathbf{53.6}$ & $44.0$ & $53.9$ & $50.4$ &  & $\mathbf{64.3}$ & $\mathbf{58.6}$ & $\mathbf{41.8}$ \\
\midrule 
\multirow{3}{*}{\textbf{\benchname} (XL)} & \Sequential{} &  $2.7\times 10^{20}$ & $66.5$ & $54.2$ & $61.2$ & $61.0$ &  & $63.1$ & $68.9$ & $56.8$\\
&  \Cumulative{}-\Full{} &  $2.7\times 10^{20}$ & $71.6$ & $58.8$ & $65.1$ & ${64.8}$ &  & $\mathbf{70.7}$ & $\mathbf{68.5}$ & $\mathbf{57.1}$\\
\rowcolor{gray!10} &  \Cumulative{}-\Full{}$^{*}$ &  $3.5\times 10^{20}$ & $\mathbf{72.8}$ & $60.4$ & $66.5$ & $\mathbf{66.7}$ &  & $\mathbf{71.0}$ & $\mathbf{68.6}$ & $\mathbf{57.1}$ \\
\rowcolor{gray!10} &  \Oracle{}$^{**}$ &  $1.1 \times 10^{21}$ & $\mathbf{73.3}$ & $\mathbf{61.3}$ & $\mathbf{68.0}$ & $65.8$ & & - & - & - \\
\bottomrule[1.2pt]
\end{tabular}
}    
    \label{tab:main_results}
    \vspace*{-8pt}
\end{table}

Our main results are in \tabref{tab:main_results} and more detailed plots on each dataset are in \appref{app:detailed_results}. 
Recall, our goal is compete   
with an \Oracle{} that re-trains from scratch 
every time new data is observed, both on dynamic and static tasks, 
while being computationally efficient. 
Here, we summarize our key findings:

\textbf{\Cumulative{}-\Full{} saves up to $4\times$ the cost.} 
On dynamic evaluation tasks,  we observe that \Cumulative{}-\Full{}
where we replay all the past data, achieves performance close to the \Oracle{} (within 1\%) 
using significantly less compute ($4\times$ less on \benchname{} and $2.5\times$ less on \cyfcc{} and \credcaps{}). 
On static tasks, the gap remains small at small scales but grows to
4.7\%  on \mlarge{},
1.8\% on \mxlarge{} Bestpool,
and 4\% on \mxlarge{} Basic (see \tabref{tab:main_results} and \tabref{tab:basic_xlarge_result}).
In these cases, training \Cumulative{} models with
slightly extra compute bridges the gap 
while remaining at least $2.7\times$ more computationally efficient (see rows with $^*$  in \tabref{tab:main_results}).
This highlights that with unconstrained access to past data, 
we can simply train sequentially and save significant computational resources.

\begin{figure}[t]
    \centering
    \begin{subfigure}{0.58\linewidth}
        \centering
        \includegraphics[width=\linewidth]{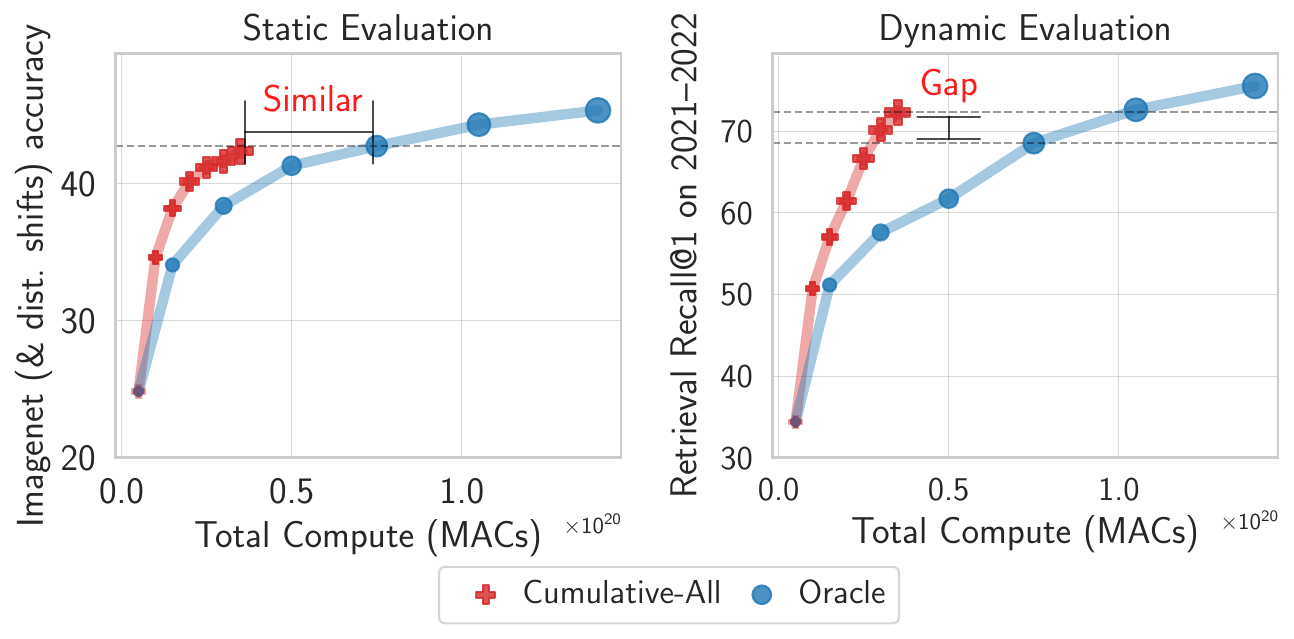}
    \end{subfigure}
    \hfill
    \rule{0.8pt}{4cm} 
    \hfill
    \begin{subfigure}{0.38\linewidth}
        \centering
        \includegraphics[width=\linewidth]{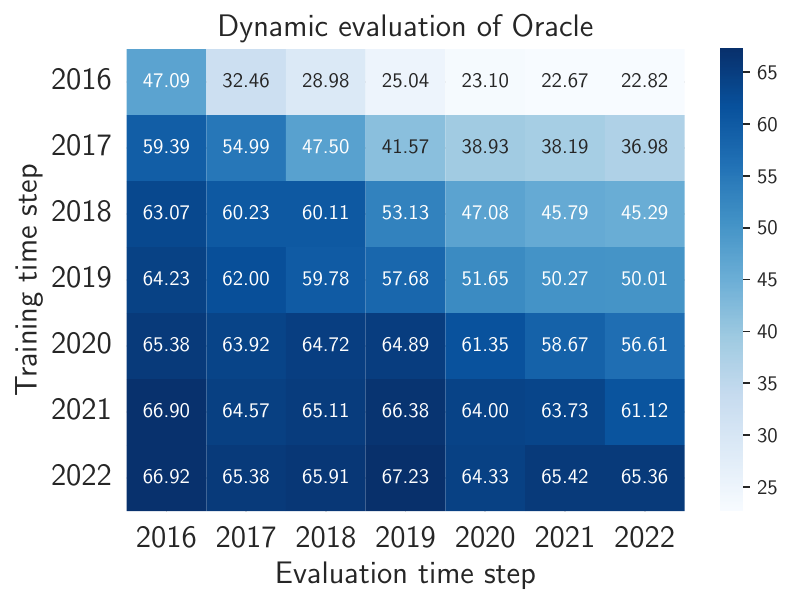}
    \end{subfigure}
    
    \vspace*{-4pt}
    \caption{\emph{(Left)} \textbf{Dynamic and static evaluations rank
    models differently}.
    Models with similar performance on static datasets, have $> 6\%$ difference on retrieval task from 2021-2022 \benchname{} (L).
    Different points denote models trained sequentially over time. 
       \emph{(Right)} \textbf{Performance of \Oracle{} on future time steps
       drops highlighting distribution shift in dataset}.
       Each row evaluates the \Oracle{} trained on \benchname{} (L) at a particular time step across all dynamic retrieval tasks.
    }
    \label{fig:static_dynamic_comp}
    \vspace*{-15pt}
\end{figure}

\textbf{At scale, \Sequential{} has strong forward transfer but lacks on static tasks.}
On \cyfcc{} and \credcaps{}, which are at the smallest scale, we observe a significant gap ($>10\%$)
between \Sequential{} (with no data replay) and \Oracle{}
 on all tasks.
On the other hand, on all scales in \benchname{}, \Sequential{} 
shows strong performance on forward transfer and ID dynamic evaluations. 
However, on static tasks and backward transfer evaluations, 
\Sequential{} significantly underperforms the \Oracle{}.

\textbf{\Patching{} and \LwF{} improve over \Sequential{} but lag behind \Cumulative{}-\Full{}.}
On static tasks, \LwF{} improves over \Sequential{} by 2\%,
while on dynamic tasks, 
\LwF{} improves backward transfer by 7\% on \benchname{} (M).
However, its computation cost is higher than even 
\Cumulative{}-\Full$^*$ which outperforms \LwF{} on all tasks. 
\Patching{} improves over 
\Sequential{} on backward transfer on all datasets 
(e.g., 5\% boost on \benchname{} L)
highlighting that \Patching{}
combines benefits of previously patched model and the 
new \Sequential{} model without additional computation cost.
However, such benefits do not show up on static tasks. 
These results 
hint that to continuously 
improve on static tasks with time, 
replaying old data as in \Cumulative{}-\Full{} plays a crucial role.

\textbf{-\Exponential{} and -\Equal{}
significantly reduce replay %
buffer size and maintain static task performance and backward transfer.}
Recall, that -\Exponential{} and -\Equal{} reduce the replay buffer size to 
a maximum $2D$ of old data.
In particular, at the last time step, -\Exponential{} and -\Equal{} 
reduce the buffer size by $3.5\times$ for \benchname{} datasets.  
While reducing the buffer sizes, these methods still achieve performance 
close to \Cumulative{}-\Full{} (within $2\%$) on both static and dynamic tasks, 
with -\Equal{} consistently better than -\Exponential{} strategy. 
As we go to large scale, e.g., from \mmedium{} to \mlarge{}, the gap between 
these methods and \Cumulative{}-\Full{} reduces. 
These findings demonstrate that even a small amount of replay data 
from old time steps stays competitive with replaying all data and significantly
improves over no replay at all.

\textbf{Warm up helps training on data from first time step, but hurts on subsequent time steps.}
Cosine LR is commonly coupled with an initial warm-up that linearly
increases the LR from zero to maximum LR.
We investigate the effectiveness of warm-up in first versus subsequent time steps.
Surprisingly, we observe that not using warmup for subsequent 
training runs is \emph{strictly} more beneficial
than using warm up
on both static and dynamic tasks. In particular, 
on \benchname{} (L), we observe about $1.5\%$ improvement in ImageNet accuracy and $4.3\%$ improvement on ID dynamic retrieval when not using warmup with \Cumulative{} (see \appref{app:lr_results}).
Moreover, we also ablate over not using warm up for the first training run and observe a drop of 
approximately $4.8\%$ accuracy in the first time step on \benchname{} (L). 
Hence, we default to using warmup when training on the first time step 
and not using it on the subsequent time steps with all methods 
except for training on \benchname{} (XL) where we 
add a smaller warm up (10\% of the warm up iterations used in first step) to stabilize training.

\textbf{Same maximum LR works best across all runs when using cosine schedule.} 
We ablate on \benchname{} (M) to investigate
how to change LR after training 
on data from the first time step. 
Unlike conventional pretraining and finetuning settings where LR is typically decreased for subsequent training, 
we observe that decaying maximum LR for subsequent steps in our setup hurts on static and dynamic tasks 
and consequently, we use same 
maximum LR across our runs (see \appref{app:lr_results}). %

\textbf{Filtering strategy changes the ordering of performance on static and dynamic retrieval tasks.}
We observe that while bestpool filtering  models
outperform basic filterining models 
on \benchname{} (XL) by 6\%
on static tasks, they underperform by over $5\%$
on dynamic retrieval task (see \figref{fig:bestpool_basic}). 

\textbf{Dynamic tasks provide complimentary information for model selection compared to static tasks.} 
Choosing models solely based on static task performance may 
inadvertently select models that underperform on dynamic tasks. 
For example, \Cumulative{} models that show relatively modest improvements on static tasks 
continue to improve by $>6\%$ for retrieval on 2021-2022 (\figref{fig:static_dynamic_comp}).

\begin{wrapfigure}{r}{0.4\textwidth}
    \vspace{-12pt}
\captionof{table}{
ImageNet continual training.
\Cumulative{}-\Full{} remains close to \Oracle{}.  
} 
    \vspace*{-7pt}
\centering
    
\resizebox{0.4\textwidth}{!}{%
\begin{tabular}{lc|ccc}
\toprule[1.2pt]
\multirow{2}{*}{\textbf{Method}} & \multicolumn{4}{c}{\textbf{Number of splits}} \\
\cmidrule{2-5}
& 1 (\Oracle{}) & 2 & 4 & 8\\
\midrule
\Cumulative{}-\Full{}  & $80.9$ & $80.8$ & $80.6$ & $80.0$  \\
\bottomrule[1.2pt]
\end{tabular}
}
    \label{tab:imagenet_results}
    \vspace*{-8pt}
\end{wrapfigure}

\textbf{\Cumulative-\Full{} remains competitive to \Oracle{} even on ImageNet on up to 8 splits.}
CLIP models are often trained for fewer epochs and are typically not trained until they reach an ``overfitting'' regime. 
Here, we investigate how \Cumulative-\Full{}  performs when compared to \Oracle{} when training is done for longer. Specifically, we assess \Cumulative-\Full{} on 2, 4 and 8 IID splits including the full dataset (see \appref{app:imagenet_exp} for details). 
\tabref{tab:imagenet_results} summmarizes our key findings.
Notably, even with up to 8 splits, the difference in accuracy between \Oracle{} and \Cumulative-\Full{} remains below 0.9\%. These results underscore the feasibility of continual training with \Cumulative-\Full{} even on ImageNet.

\vspace{-8pt}
\section{Related Work}
\vspace{-6pt}
\label{sec:related}

\textbf{Benchmarks for continual learning{} {} } Traditionally, the continual learning community has focused on
domain, class, and task incremental benchmarks~\citep{hsu2018re,van2019three, zhou2023pycil} with artificial task boundaries (e.g., 
Split-CIFAR, Perm-MNIST).
These benchmarks are often task-specific and 
present minimal or no meaningful evolution between 
adjacent tasks. 
Consequently, continual learning methods 
are often confined to these benchmarks
and seldom scale to practical real-world scenarios \citep{cossu2022class,lin2021clear}.
On the other hand, continual learning methods for CLIP models
are primarily aimed at fine-tuning 
to improve performance on a single or on a sequence 
of disjoint downstream tasks~\citep{thengane2022clip,zheng2023preventing, ilharco2022patching}. Existing large-scale benchmarks for training CLIP models, 
e.g., Datacomp~\citep{gadre2023datacomp} and LAION-5B~\citep{schuhmann2022laion}, are curated to investigate methods and scaling laws to 
train state-of-the-art
CLIP models in a single training run. In our work, we augment these existing 
datasets with temporal information to create benchmarks for continual pertaining of CLIP models. 

\textbf{Continual learning methods {} {}}
Common methods can be categorized into three categories: i) regularization,
ii) replay, and iii) architecture-based methods. Regularization 
methods add a penalty to keep the fine-tuned model close to its initialization and often incur additional memory/compute costs \citep{kirkpatrick2017overcoming,
mirzadeh2020dropout,
mirzadeh2020understanding,farajtabar2020orthogonal}. 
Data replay methods retain 
all or a subset of the 
prior data 
for subsequent training~\citep{lopez2017gradient,rebuffi2017icarl,chaudhry2018efficient}.  
Simple replay-based baselines surpass various methods on standard benchmarks~\citep{lomonaco2022cvpr,balaji2020effectiveness,prabhu2020gdumb}.  
Lastly, architecture-based methods expand the model as new tasks arrive, 
limiting their applicability in evolving 
environments without clear task boundaries~\citep{schwarz2018progress,rusu2016progressive}. 
In this work, we compare popular continual learning methods 
 with simple alternatives for continually pretraining of CLIP.

\vspace{-7pt}
\section{Conclusion and Future Work}
\vspace{-5pt}

We view \benchname{} as the initial stride toward the continual 
training of large-scale vision-language foundation models.
We believe that our benchmark, alongside the preliminary results obtained 
using simple baselines will foster future
research for large-scale continual-learning. 
There are several pivotal directions for future work:
(i) Compare our baselines on continually streaming data at finer 
granularity, e.g., streaming data at the monthly level;
(ii) Investigate alternate learning rate schedules (e.g., \constcosine{} as in \appref{sec:const_cosine_schedule}) that 
are forward looking, and are better suited to continual learning;
(iii) Better data filtering techniques that are more inclusive of future data;
(iv)  Expand our problem setup to encompass the training of other large-scale 
foundation models.

\bibliography{references}
\bibliographystyle{arxiv}

\appendix
\newpage

\section{Continual Learning Benchmarks and Methods}
\label{app:cl_benchmarks}

We introduce a large-scale image-text benchmark with web scale streaming image 
text pairs specially developed for studying how efficiently one can get a fresh 
CLIP model with new incoming batches of data.  \Cref{tab:cl_dataset_comparison} 
compares the proposed benchmark with existing datasets for continual learning.  
Note that this table is not aimed to be an exhaustive list of all CL datasets, 
but the most popular benchmarks in each domain. For language modeling tasks we 
report the number of examples/documents as the number of samples and for 
detection tasks we report the number of labeled objects/bounding boxes.

\begin{table}[H]
    \centering
        \caption{Comparison with continual learning benchmarks.}
    \label{tab:cl_dataset_comparison}
    \vspace{-5pt}
    \resizebox{0.99\textwidth}{!}{%
\begin{tabular}{lcccccc}
\toprule[1.1pt]
\textbf{Benchmark}                           & \textbf{\# Samples} & \textbf{Years}  & \textbf{Time-Continual} & \textbf{Image-Text}& \textbf{Task}                 \\
\midrule[1.1pt]
Split-MNIST~\citep{goodfellow2013empirical}&  60K                & 1998                 &  \xmark                 &   \xmark           & Classification                \\
Perm-MNIST~\citep{goodfellow2013empirical} &  60K                & 1998                      &  \xmark                 &   \xmark           & Classification                \\
Rot-MNIST~\citep{lopez2017gradient} &  60K                & 1998                     &  \xmark                 &   \xmark           & Classification                \\
Split-CIFAR-100~\citep{zenke2017continual}    &        50K          & 2008                   &  \xmark                 &   \xmark           & Classification                \\
Split-MINI-ImageNet~\citep{chaudhry2019tiny}    &  50K                &   2009                   &  \xmark                 &   \xmark           & Classification                \\
Split-ImageNet~\citep{wen2020batchensemble}    &  1.2M                & 2009                         &  \xmark                 &   \xmark           & Classification                \\
Split-ImageNet-R~\citep{wang2022dualprompt}    &  30K                & 2019                          &  \xmark                 &   \xmark           & Classification                \\
CORe50~\citep{lomonaco2017core50}                   &  165K                & 2017        &  \xmark                &   \xmark            & Detection \\
CLAD~\citep{verwimp2023clad}                   &  23K                & 2021          &  \xmark                 &   \xmark            & Detection \\
WANDERLUST~\citep{wang2021wanderlust}                   &  326K     &     2021              &  \cmark                &   \xmark            & Detection \\
Inc-PASCAL~\citep{michieli2019incremental}                   &  11K                & 2012        &  \xmark                 &   \xmark            & Segmentation \\
Inc-ADE20K~\citep{cermelli2020modeling}                   &  20K                & 2012
&  \xmark                 &   \xmark            & Segmentation \\
StreamingQA~\citep{liska2022streamingqa}                   & 100K                 &    2007–2020      &  \cmark                 &   \xmark           &     Question Answering            \\
TemporalWiki~\citep{jang2022temporalwiki}                   &   32M              &     2021     &  \cmark                 &   \xmark           & Language Modeling                 \\
CKL~\citep{jang2021towards}                   &     30K            &  2019-2021        &  \xmark                 &   \xmark           & Language Modeling                 \\
CTrL~\citep{veniat2020efficient}                   &     300K             &    1998-2017     &  \xmark                 &   \xmark           & Classification                \\
CLOC~\citep{cai2021online}                   &  39M                & 2006-2014         &  \cmark                 &   \xmark           & Classification                \\
CLEAR~\citep{lin2021clear}                   &  7.8M               & 2004-2014                   &  \cmark                 &   \xmark           & Classification                \\
NEVIS~\citep{bornschein2022nevis}            &  8M                 & 1992-2021                   &  \cmark                 &   \xmark           & Classification                \\
Mod-X~\citep{ni2023continual}            &  156K                 & 2014                  &  \xmark                 &   \cmark           &  Retrieval \\
CLiMB~\citep{srinivasan2022climb}            &  1.3M                 & 2013-2021              &  \xmark                 &   \cmark           & Classification \\
\midrule
\cyfcc                                       & 15M                 & 2008-2014                       &  \cmark                 &   \cmark           & Retrieval / ZS Classification \\
\credcaps                                    & 12M                 & 2011-2020                       &  \cmark                 &   \cmark           & Retrieval / ZS Classification \\
\benchname                                   & 100M/1B/12B         & 2014-2022                       &  \cmark                 &   \cmark           & Retrieval / ZS Classification \\
\bottomrule[1.1pt]
\end{tabular}
    }
    \vspace{-5pt}
    
\end{table}

\vspace{-6pt}
\subsection{\update{Extended Related Work}}
\vspace{-3pt}

Neural networks trained on new data suffer from 
catastrophic forgetting of prior 
knowledge~\citep{sutton1986two,goodfellow2013empirical}.
Addressing the continual learning challenge, researchers have primarily honed 
in on methods tailored for small-scale benchmarks, specifically focusing on 
domain, class, or task incremental benchmarks~\citep{hsu2018re,van2019three}. 
Continual learning of foundation models would significantly reduce the costs 
and increase quick adaptability. While some recent works have started to 
introduce continual learning benchmarks, they are not naturally 
time-continual and are comparatively much smaller in 
scale~\citep{ni2023continual,srinivasan2022climb}.
While evaluations on these benchmarks often neglect the consideration of 
``training time'', it becomes a pivotal factor when scaling continual 
learning approaches to scenarios involving the training of foundation models 
such as {CLIP}.

In our study, we abstain from comparing with continual learning methods that  notably prolong the ``training time''. Methods such as 
GEM~\citep{lopez2017gradient, chaudhry2018efficient},  
and IMM~\citep{lee2017overcoming}, which 
compute gradients for two models in each training iteration, essentially double
the training duration.
For completeness, we include a comparison with 
LWF~\citep{li2017learning, ding2022don}  and 
EWC~\citep{kirkpatrick2017overcoming}. 
While these methods increase computation cost over standard training
due to an additional forward pass, the increase in computation cost is relatively much smaller than methods that compute additional gradients.
Our LWF implementation is motivated by \citet{ding2022don} which focuses on continual fine-tuning CLIP models on classification tasks by adapting LwF to CLIP models. 
Instead, for setups where additional compute resources are available, we 
run our \Cumulative{}-\Full{} approach for slightly longer. \Cumulative-\Full{} narrows 
the gap with \Oracle{} (refer to \tabref{tab:main_results}). Given that data 
storage costs are substantially lower than computational costs at scale, we 
advocate for taking computational efficiency into consideration in future 
endeavors.

\vspace{-6pt}
\subsection{\update{Discussion and comparison with CLOC Benchmark}}
\vspace{-3pt}

\update{\citet{cai2021online} provide interesting discussion/analysis for continual learning at a large number of steps. However, our study differs from \citet{cai2021online} in several crucial respects: 
(i) Training Methodology: We employ noisy supervision using contrastive loss between image-text pairs, as opposed to the cross-entropy loss used by \citet{cai2021online}. 
(ii) Scale of Experiments: Our experiments on the TiC-DataComp dataset are orders of magnitude larger, scaling up by 200$\times$.}

\update{These differences introduce unique challenges. The use of contrastive loss (i) necessitates a tailored approach to designing our evaluation studies. The significantly larger scale of our experiments (ii) poses challenges in collecting timestamped data and understanding if and how distribution shifts impact learning at this scale. }

\section{Additional Experimental Results}

\vspace{-5pt}
\subsection{Detailed Results on Our Benchmarks} \label{app:detailed_results}
\vspace{-5pt}

\begin{figure}[H]
    \centering
    \begin{subfigure}[b]{\textwidth}
        \includegraphics[width=\linewidth]{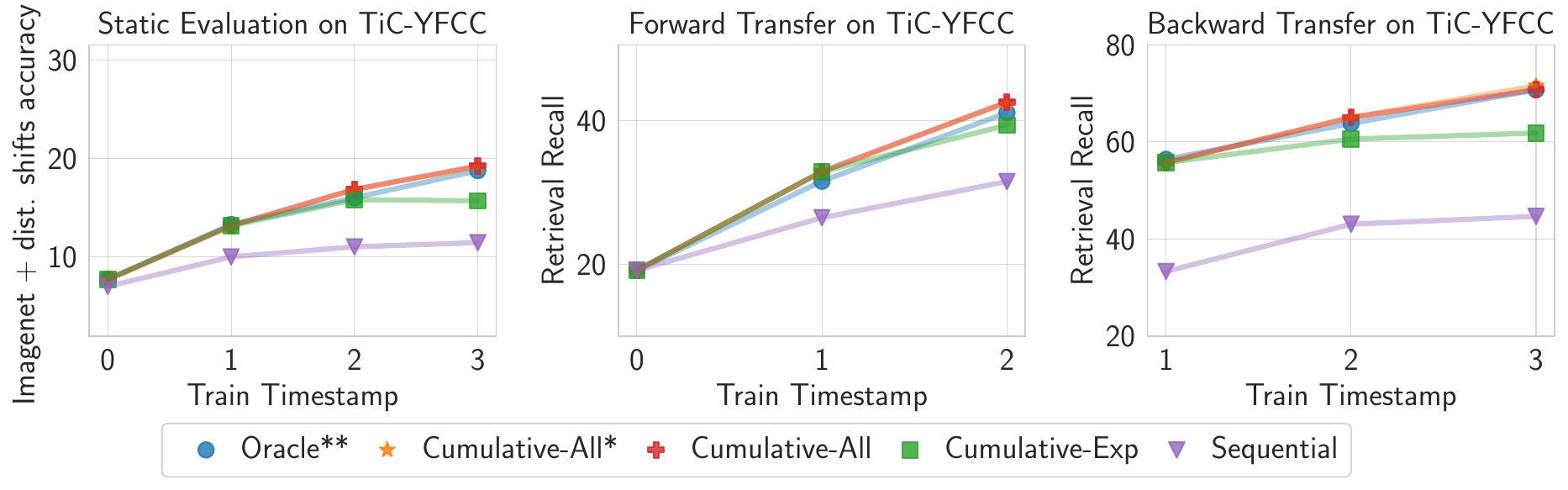}
        \caption{ \cyfcc{}.}
    \end{subfigure}

    \vspace{3pt}
    \begin{subfigure}[b]{\textwidth}
        \includegraphics[width=\linewidth]{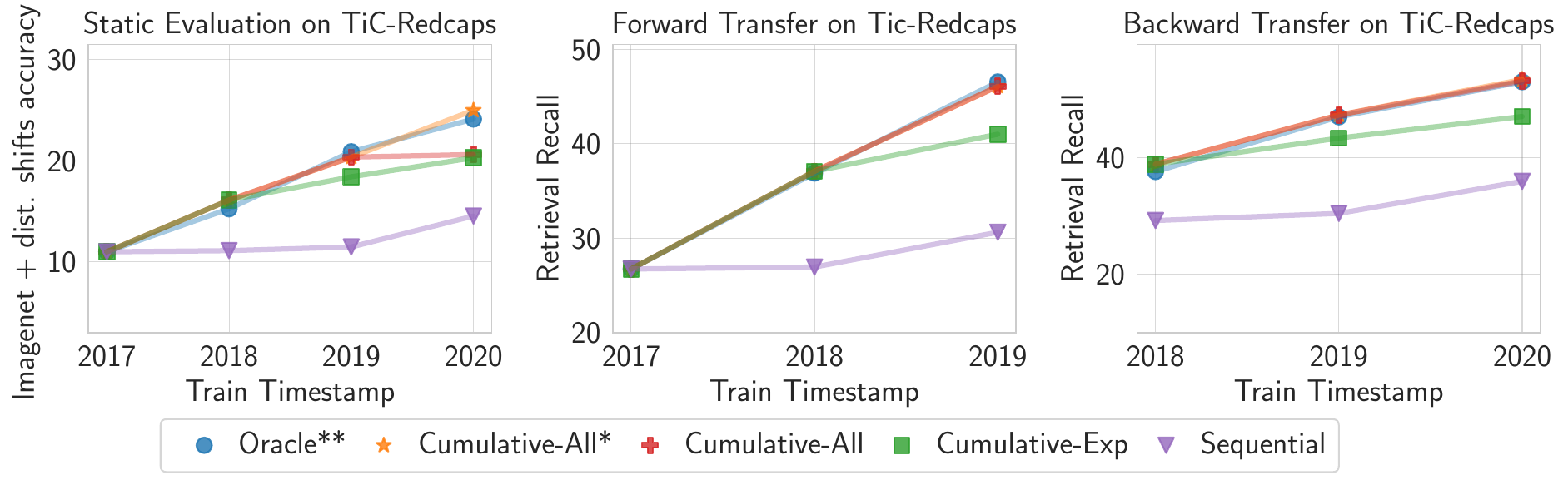}
        \caption{ \credcaps{}.}
    \end{subfigure}

    \vspace{3pt}

    \begin{subfigure}[b]{\textwidth}
        \includegraphics[width=\linewidth]{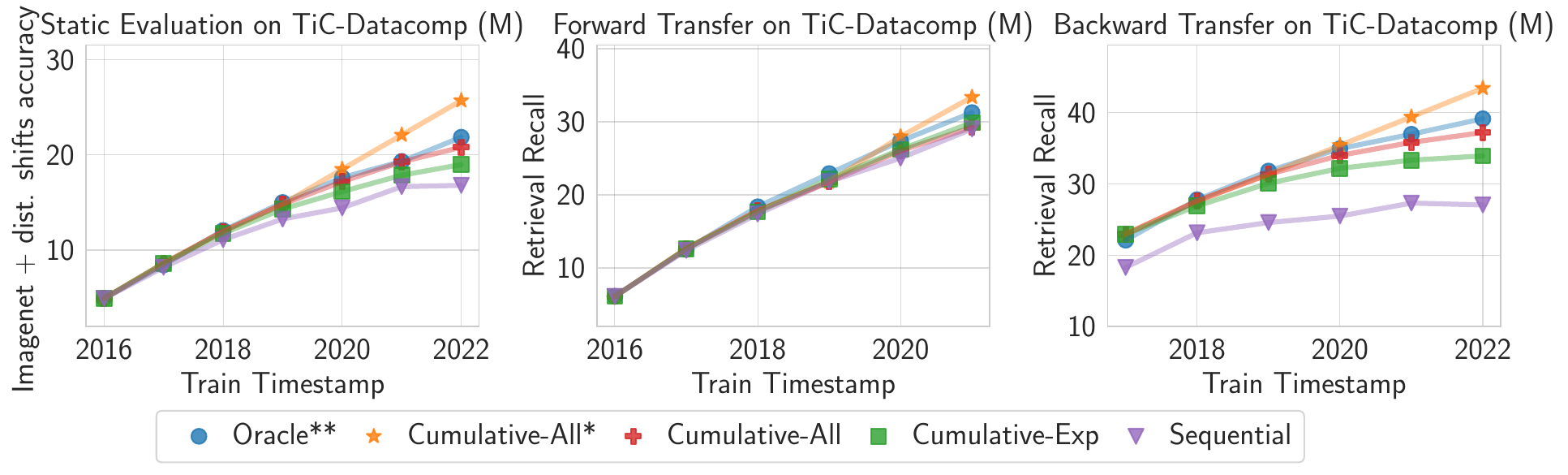}
        \caption{\benchname{} (M).}
    \end{subfigure}

    \vspace{3pt}

    \begin{subfigure}[b]{\textwidth}
        \includegraphics[width=\linewidth]{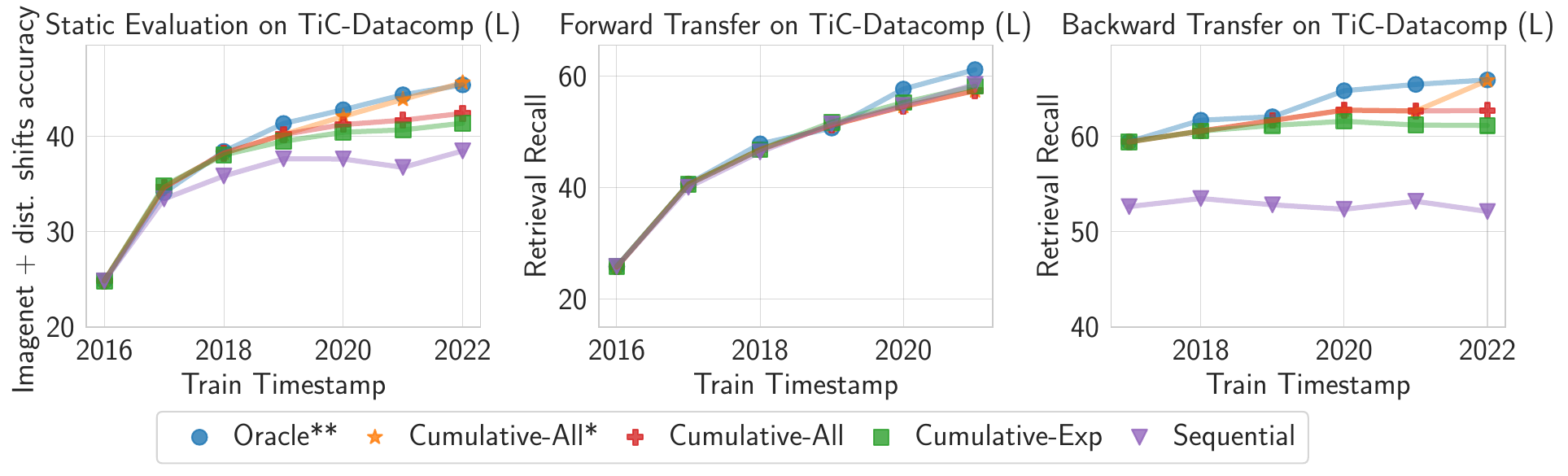}
        \caption{\benchname{} (L).}
    \end{subfigure}
    
    \caption{ \textbf{Static and dynamic evaluation performance over time with selected methods in our testbed.}
    As we get more data, all methods improve on both static and forward transfer on dynamic tasks but methods with limited replay buffer start performing slightly worse for backward transfer. 
    } 
    \label{fig:plots_datasets}
\end{figure}

\begin{figure}
    \centering
    \begin{subfigure}[b]{\textwidth}
        \includegraphics[width=0.23\linewidth]{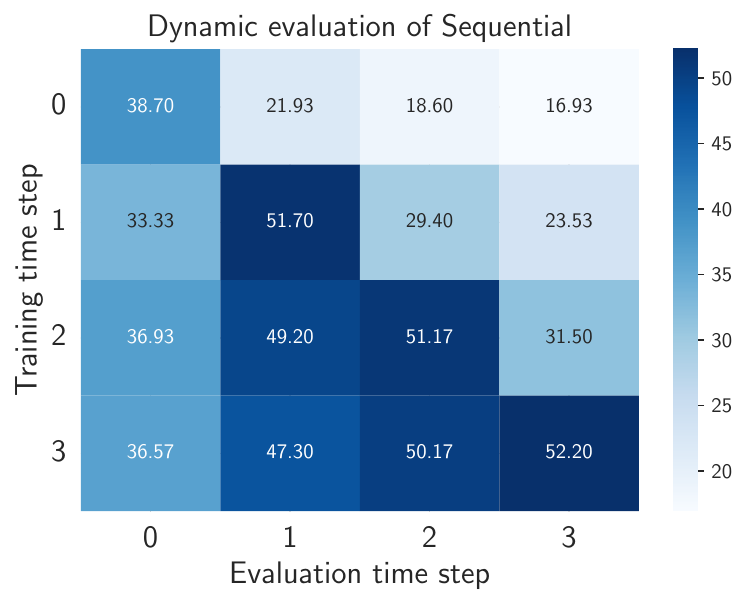}
        \includegraphics[width=0.23\linewidth]{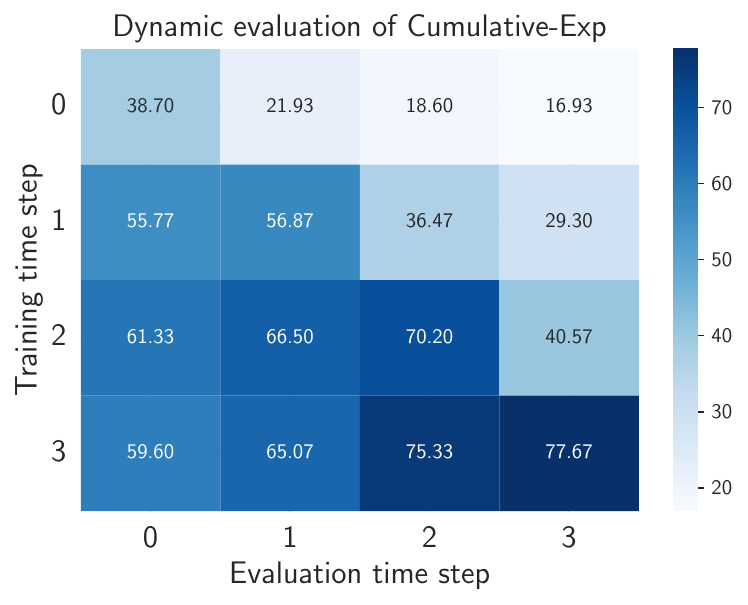}
        \includegraphics[width=0.23\linewidth]{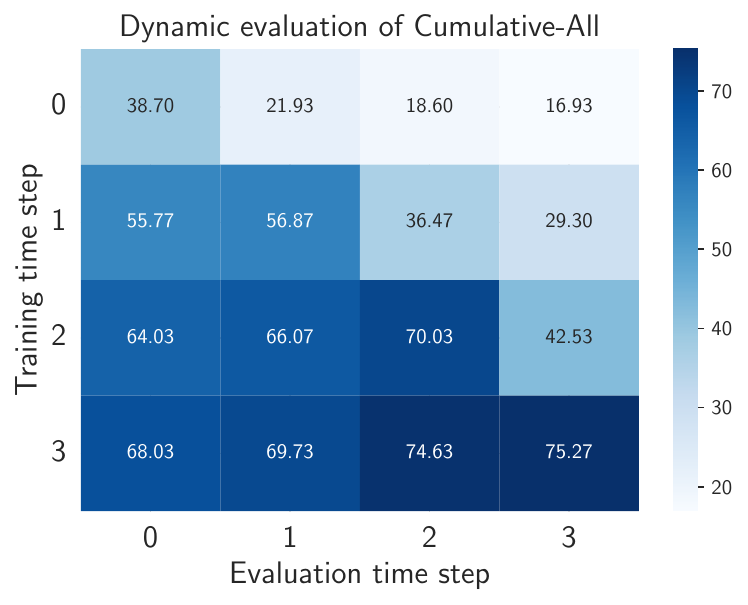}
        \includegraphics[width=0.23\linewidth]{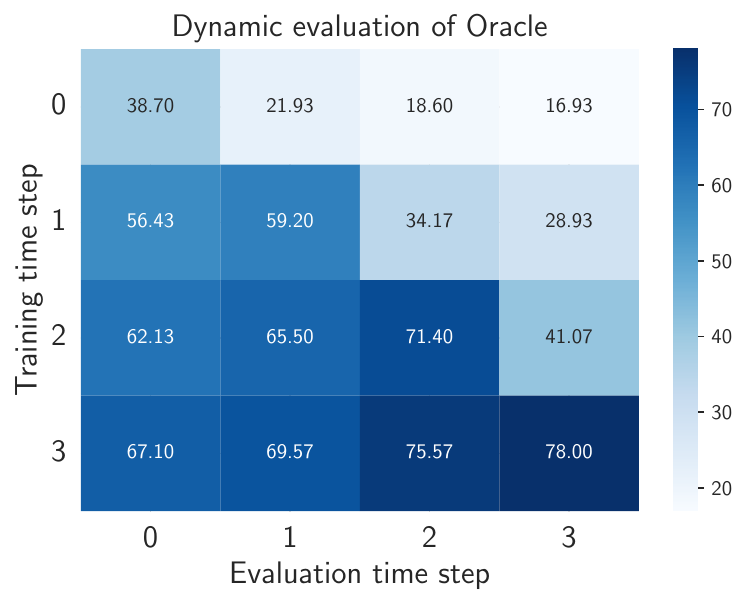}
        \caption{\cyfcc{}.}
    \end{subfigure}

    \begin{subfigure}[b]{\textwidth}
        \includegraphics[width=0.23\linewidth]{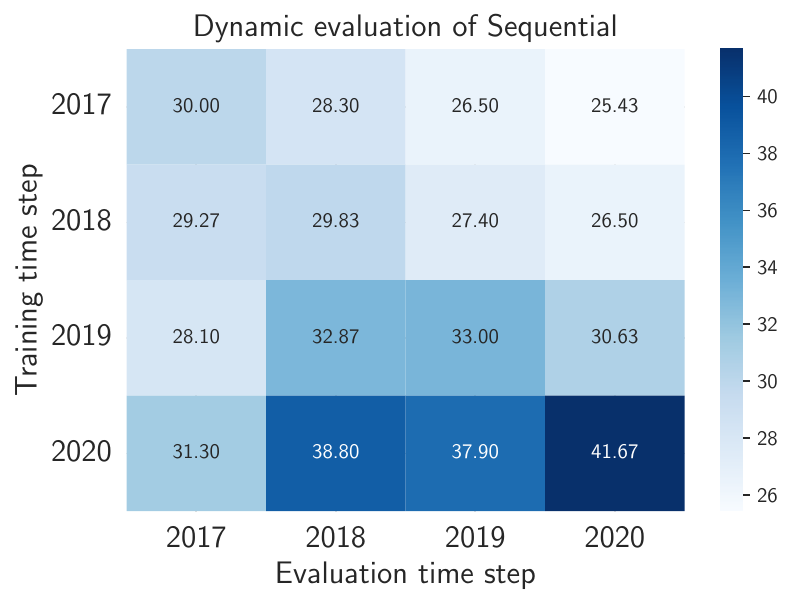}
        \includegraphics[width=0.23\linewidth]{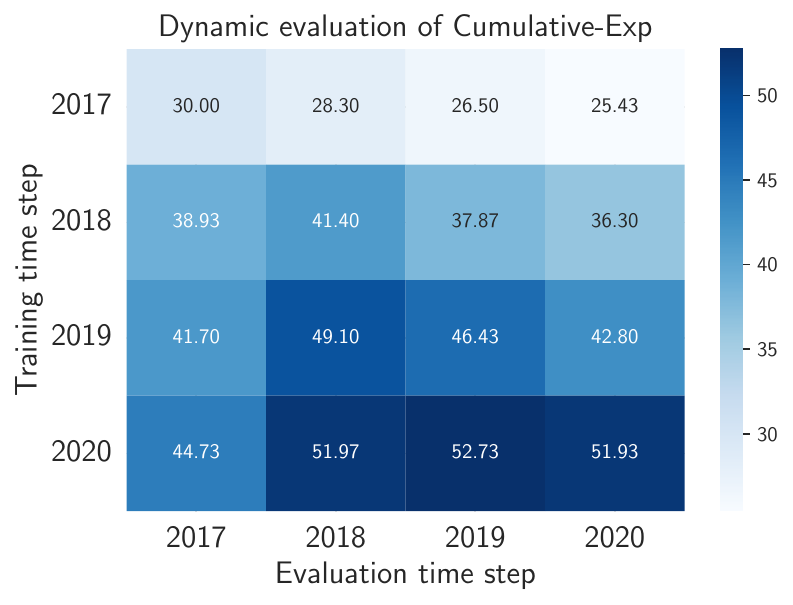}
        \includegraphics[width=0.23\linewidth]{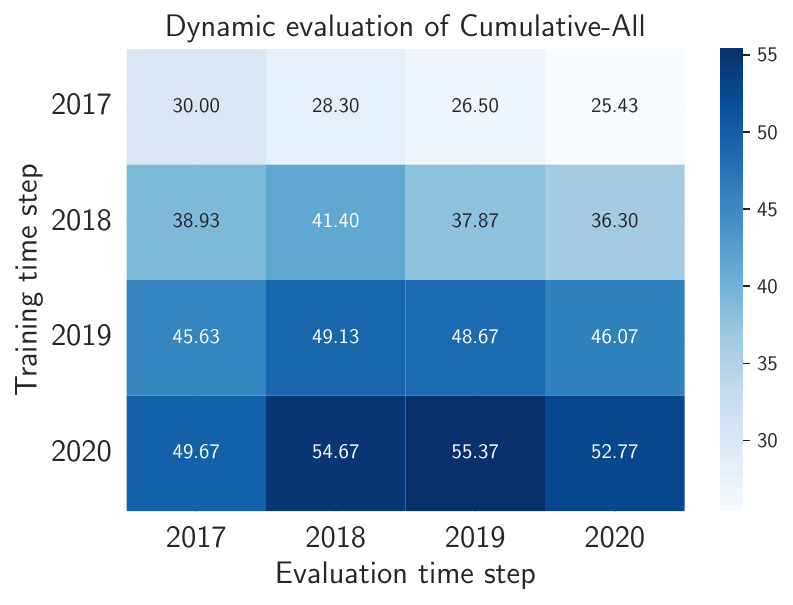}
        \includegraphics[width=0.23\linewidth]{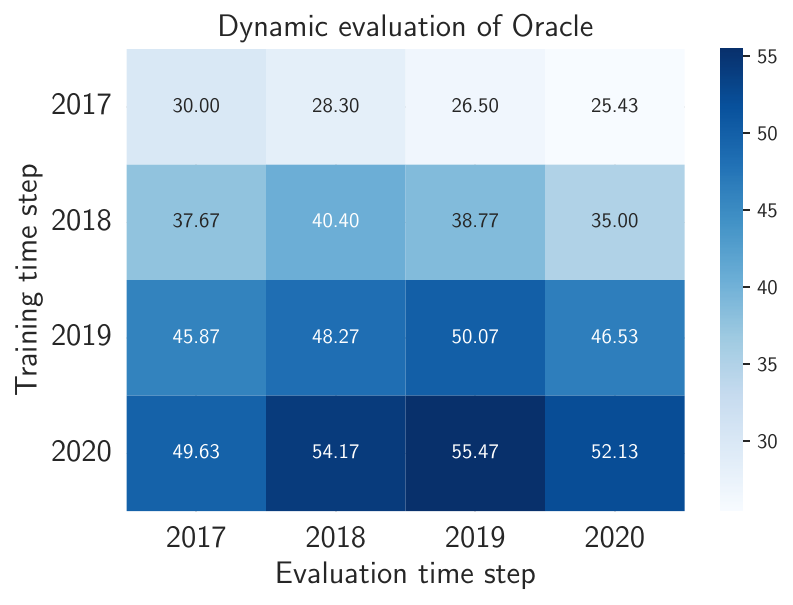}
        \caption{\credcaps{}.}
    \end{subfigure}

    \begin{subfigure}[b]{\textwidth}
        \includegraphics[width=0.23\linewidth]{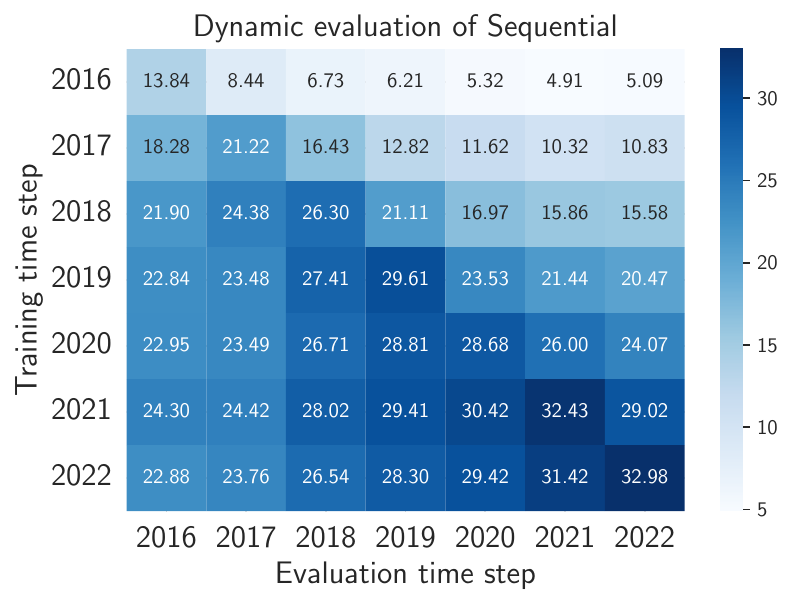}
        \includegraphics[width=0.23\linewidth]{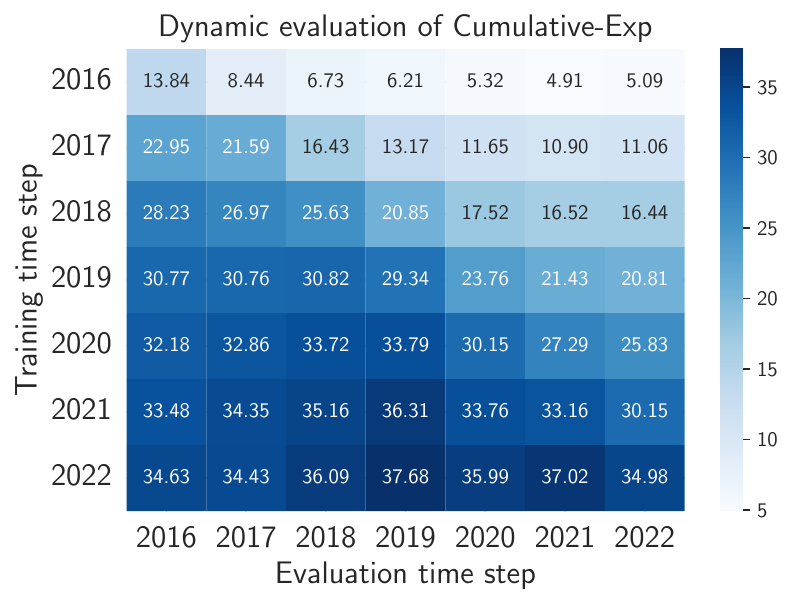}
        \includegraphics[width=0.23\linewidth]{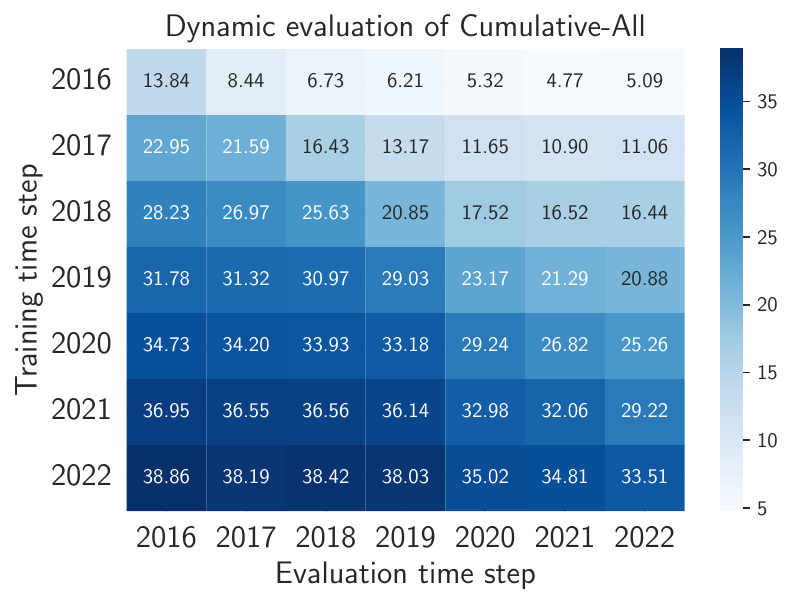}
        \includegraphics[width=0.23\linewidth]{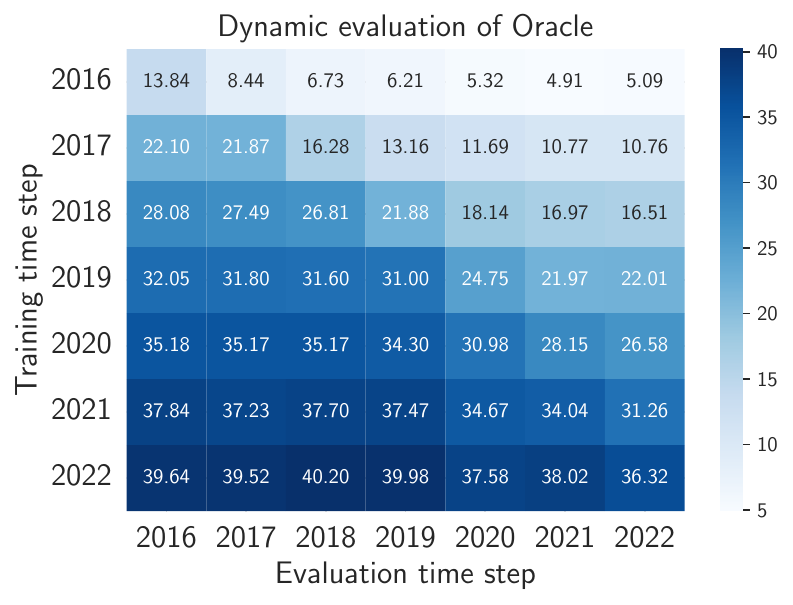}
        \caption{\benchname{} (M).}
    \end{subfigure}

    \begin{subfigure}[b]{\textwidth}
        \includegraphics[width=0.23\linewidth]{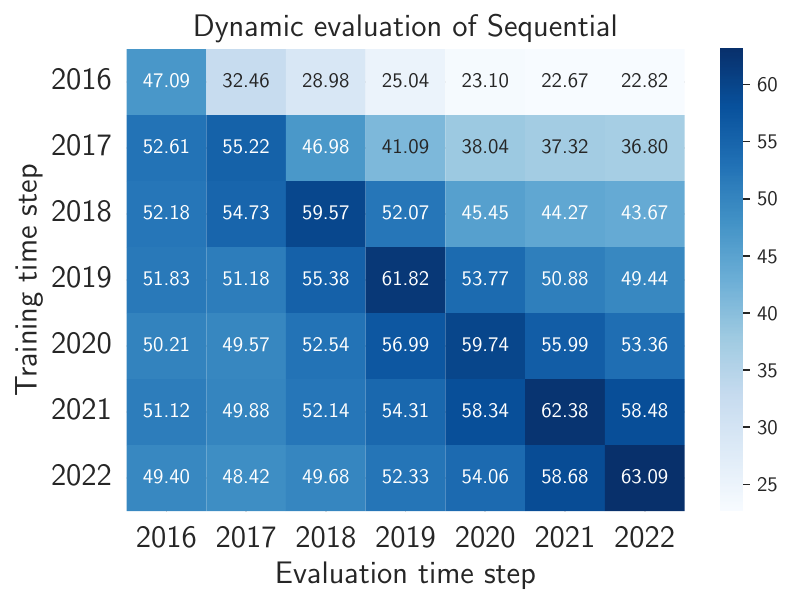}
        \includegraphics[width=0.23\linewidth]{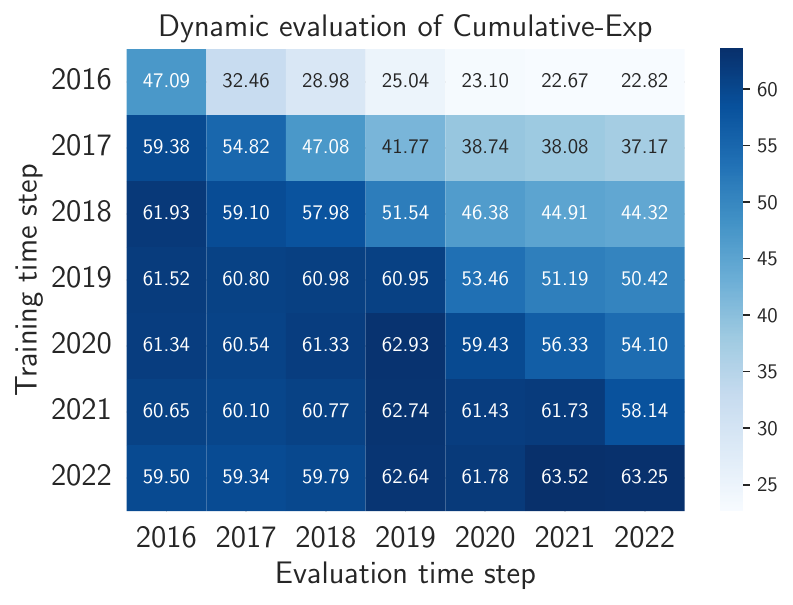}
        \includegraphics[width=0.23\linewidth]{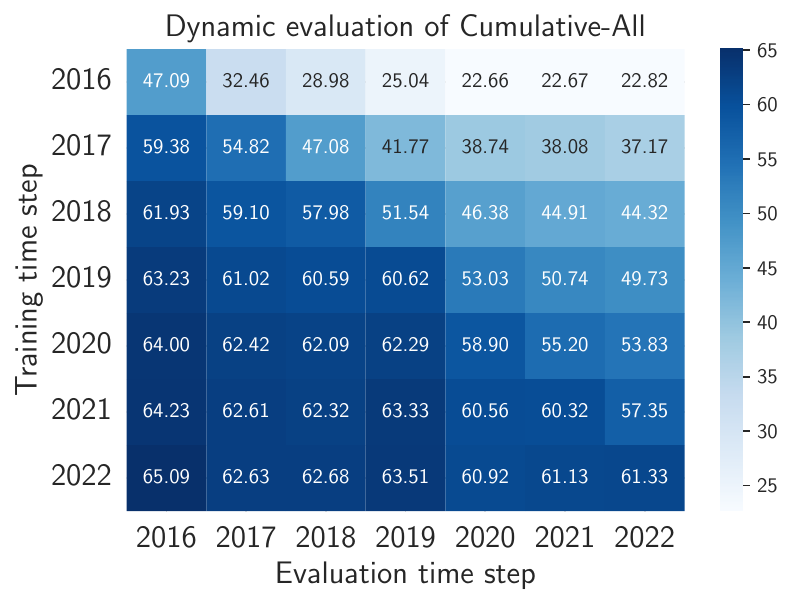}
        \includegraphics[width=0.23\linewidth]{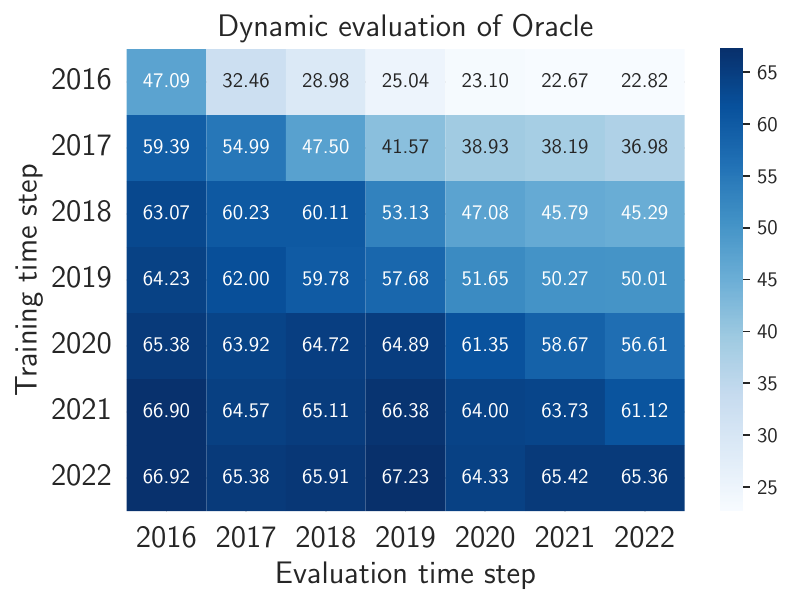}
        \caption{\benchname{} (L).}
    \end{subfigure}
    
    \caption{\update{Dynamic retrieval evaluation results on our benchmarks with \Sequential{}, \Cumulative{}-\Exponential{}, \Cumulative{}-\Full{} and \Oracle{}. These evaluations highlight the catastrophic forgetting observed with \Sequential{} and \Cumulative{}-\Exponential{}. Moreover, by observing new data, we not only benefit on tasks from current time step but also improve performance on tasks from old time steps.}}
    \label{fig:dynamic_evals_heatmap}
\end{figure}

\subsection{Results with Basic Filtering on \benchname{} XL} \label{subsec:basic_filtering_xlarge}

\begin{figure}[H]
    \centering
    \includegraphics[width=0.6\linewidth]{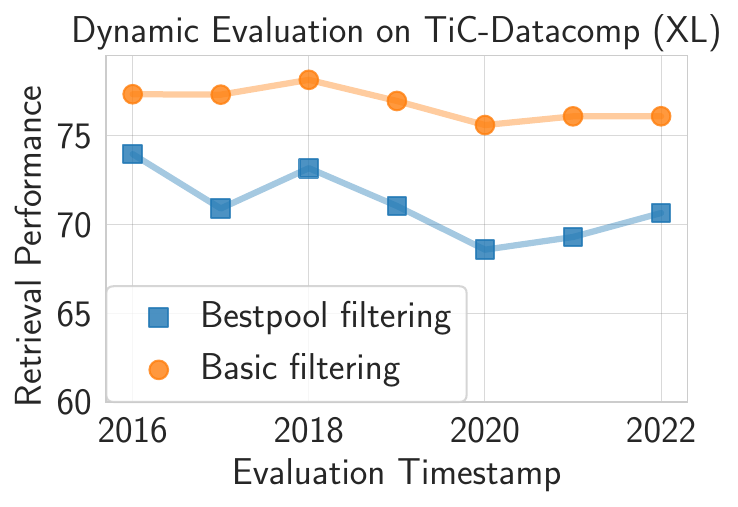}
    \caption{Comparing \Oracle{} models trained on Bestpool and Basic filtering trained on data from all time steps. Our results clearly highlight that Basic filtering performs better than Bestpool filtering on dynamic retrieval task. However, on static tasks, the order is reversed. Moreover, Bestpool filtering shows a drop in retrieval performance from 2016 to 2022 when compared with Basic filtering.}
    \label{fig:bestpool_basic}
\end{figure}

\textbf{Filtering strategy changes the ordering of performance on static and dynamic retrieval tasks.}
We observe that while Bestpool filtering  models
outperform basic filterining models 
on \benchname{} (XL) by 6\%
on static tasks, they underperform by over $5\%$
on dynamic retrieval task (see \figref{fig:bestpool_basic}). 
In the main paper (\tabref{tab:main_results}), we 
included \benchname{} (\mxlarge{}) results with Bestpool 
filtering. In \tabref{tab:basic_xlarge_result}, we include basic 
filtering results.
We observe that while Bestpool filtering models perform better than basic filtering models
on static tasks, the order is flipped on dynamic retrieval tasks. 
Hence, we resort to including results with Basic filtering at smaller scales, but
include Bestpool results for completeness as it achieves better results on static tasks.

\begin{table}[H]
    \centering
    \caption{\textbf{Zero shot performance on our time-continual benchmarks 
    (Basic and \update{Bestpool filtering}).} $^*$ and $^{**}$ denote methods that violate the 
    compute budget and use extra compute.  For static tasks, we tabulate 
    accuracy of the models obtained on the final timestamp.  For dynamic tasks, 
    we tabulate forward transfer, backward transfer and ID performance.  For 
    all metrics, higher is better. \update{Bestpool filtering results are copied from \tabref{tab:main_results}.} %
    }
\resizebox{1.0\textwidth}{!}{%
\begin{tabular}{lcccccccccc}
\toprule[1.2pt]
\multirow{3}{*}{\textbf{Benchmark}} & \multirow{3}{*}{\textbf{Method}} &   \multirow{3}{*}{\parbox{4em}{\centering \textbf{Compute} (MACs)}} & \multicolumn{4}{c}{\textbf{Static Tasks}} &  {} & \multicolumn{3}{c}{\textbf{Dynamic Retrieval Tasks}} \\[3pt]
{} & {} & {} & \multirow{2}{*}{\parbox{4em}{ImageNet}} & \multirow{2}{*}{\parbox{4em}{ImageNet dist. shift}} & \multirow{2}{*}{\parbox{4em}{Flickr30k}} & \multirow{2}{*}{\parbox{5.5em}{\centering Average over 28 datasets}} & {} & \multirow{2}{*}{\parbox{4em}{\centering Backward Transfer}} & \multirow{2}{*}{\parbox{4em}{\centering ID Performance}} & \multirow{2}{*}{\parbox{4em}{\centering Forward Transfer}}  \\ 
{} & {} & \\
\midrule 
\multirow{2}{*}{\parbox{7em}{\textbf{\benchname} (XL; Bestpool)}} & \Sequential{} &  $2.7\times 10^{20}$ & $66.5$ & $54.2$ & $61.2$ & $61.0$ &  & $63.1$ & $68.9$ & $56.8$\\
&  \Cumulative{}-\Full{} &  $2.7\times 10^{20}$ & $71.6$ & $58.8$ & $65.1$ & $64.8$ &  & $\mathbf{70.7}$ & $\mathbf{68.5}$ & $\mathbf{57.1}$\\
\rowcolor{gray!10} &  \Cumulative{}-\Full{}$^{*}$ &  $3.5\times 10^{20}$ & $\mathbf{72.8}$ & $60.4$ & $66.5$ & $\mathbf{66.7}$ &  & $\mathbf{71.0}$ & $\mathbf{68.6}$ & $\mathbf{57.1}$ \\
\rowcolor{gray!10} &  \Oracle{}$^{**}$ &  $1.1 \times 10^{21}$ & $\mathbf{73.3}$ & $\mathbf{61.3}$ & $\mathbf{68.0}$ & $\mathbf{65.8}$ & & - & - & - \\
\midrule
\multirow{2}{*}{\parbox{7em}{\textbf{\benchname} (XL; Basic)}} &  \Cumulative{}-\Full{} &  $2.7\times 10^{20}$ & $63.5$ & $52.0$ & $62.8$ & $58.7$ && $64.6$ & $55.5$ & $47.6$ \\
& \Sequential{} &  $2.7\times 10^{20}$ & $60.2$ & $48.9$ & $62.4$ & $56.6$  & & $51.6$ & $50.3$ & $45.0$ \\
\rowcolor{gray!10} &  \Oracle{}$^{**}$ &  $1.1 \times 10^{21}$ & $66.0$ & $54.0$ & $63.8$ & $59.6$ & & - & - & - \\
\bottomrule[1.2pt]
\end{tabular}
}    
    \label{tab:basic_xlarge_result}
\end{table}

\subsection{Ablations with learning rate warmup and maximum learning rate} \label{app:lr_results}

\begin{figure}[t]
    \centering
    \begin{subfigure}[b]{0.9\textwidth}
        \includegraphics[width=\textwidth]{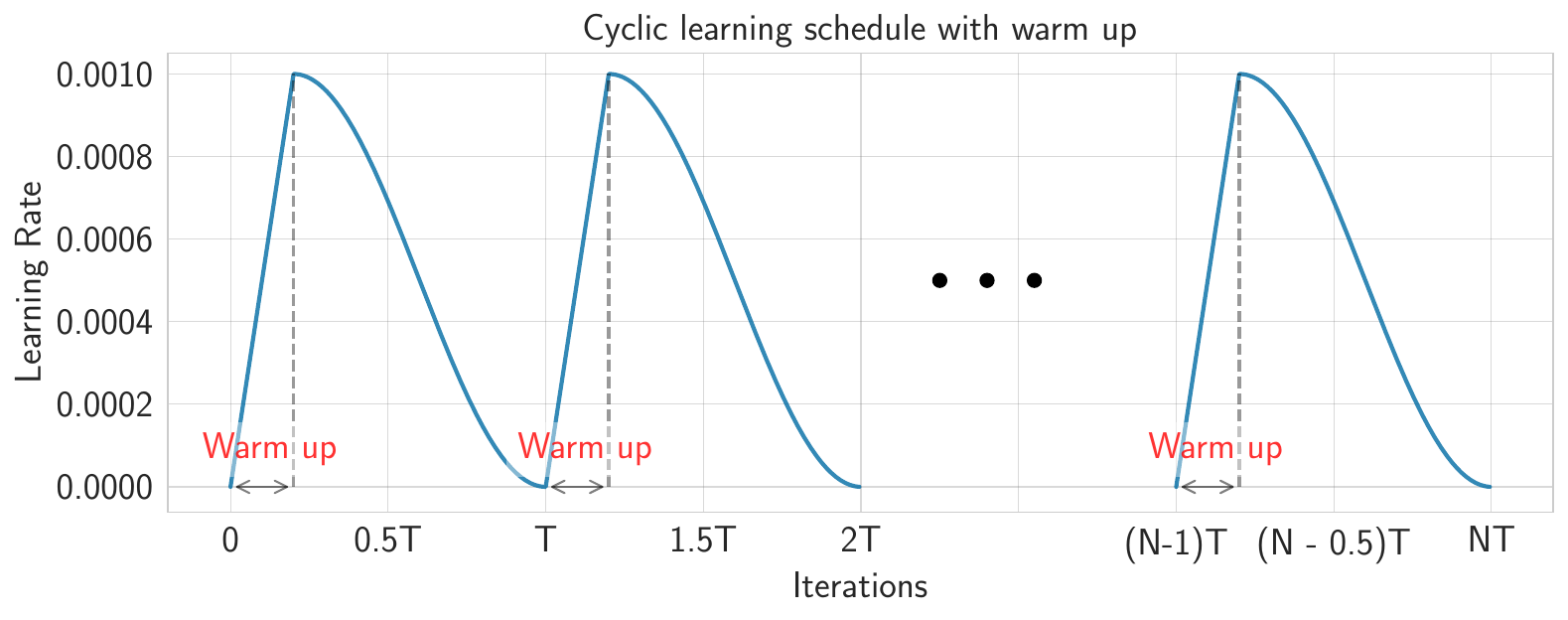}
        \caption{Multiple cycles of standard cosine learning rate schedules which involves warm-up for all subsequent training runs.}
    \end{subfigure}
    \vspace{5pt}
    
    \begin{subfigure}[b]{0.9\textwidth}
        \includegraphics[width=\textwidth]{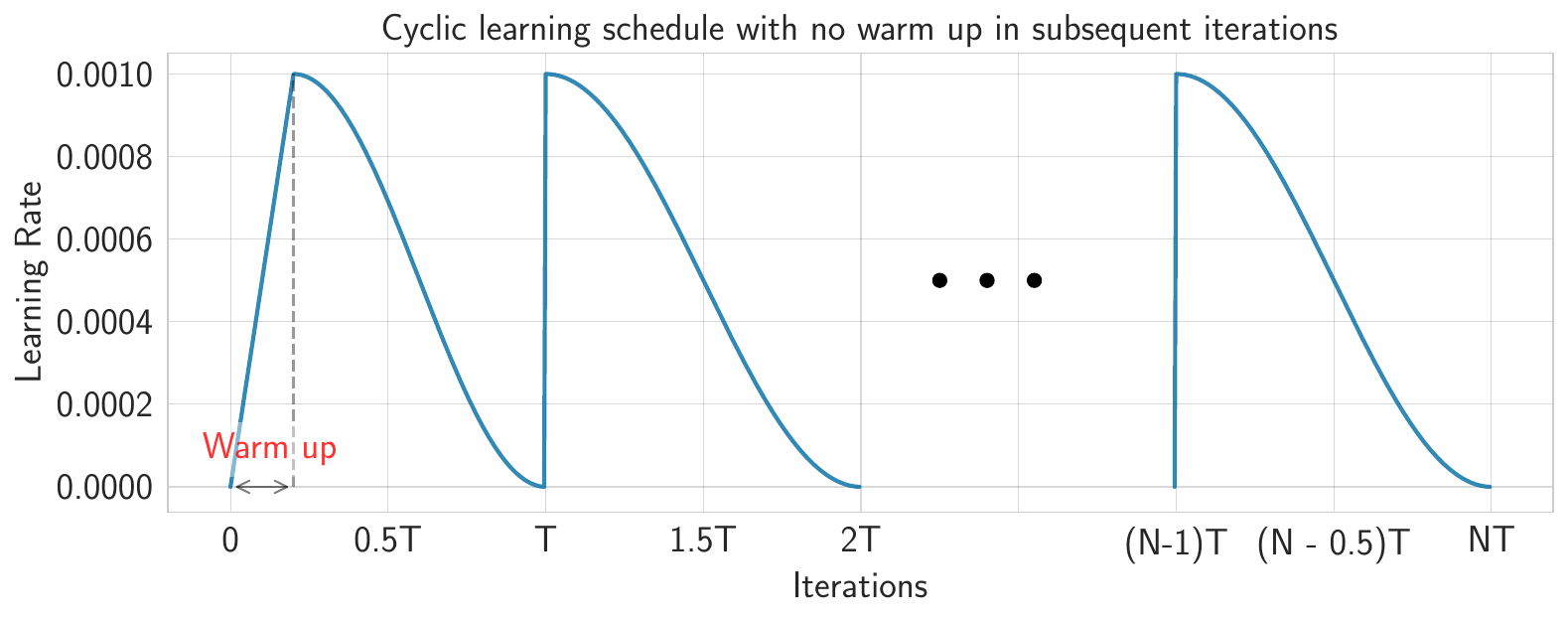}
        \caption{Our proposed cosine learning rate schedule without learning rate warm-up for subsequent training runs.}
    \end{subfigure}
    \vspace{5pt}

    \caption{\textbf{Learning rate schedule ablations.} Schedules vary on how continual training is performed when the training run is initialized with the best previous model. When training with cosine learning schedules for subsequent runs, we observe that keeping the same maximum learning rate as the first run performs the best.  }
    \label{fig:lr_schedules}
\end{figure}

To continually train models as more data arrives sequentially over time, we use multiple cycles of cosine learning rate schedule (\figref{fig:lr_schedules}). There are two crucial design choices: (i) Should we warm up the learning rate for subsequent continual runs? and (ii) How should the maximum learning rate change for sequential training runs?

\begin{table}[H]
    \centering
    \caption{\update{\textbf{Zero shot performance on our time-continual benchmarks with and without initial LR wamrup for subsequent runs.} Using warm up on sequential runs after training on the first time step hurts slightly when compared with not using warm up on sequential runs.} 
    }
\resizebox{1.0\textwidth}{!}{%
\begin{tabular}{lcccccccccc}
\toprule[1.2pt]
\multirow{3}{*}{\textbf{Benchmark}} & \multirow{3}{*}{\textbf{Method}} &  \multicolumn{4}{c}{\textbf{Static Tasks}} &  {} & \multicolumn{3}{c}{\textbf{Dynamic Retrieval Tasks}} \\[3pt]
{} & {} & {} & \multirow{2}{*}{\parbox{4em}{ImageNet}} & \multirow{2}{*}{\parbox{4em}{ImageNet dist. shift}} & \multirow{2}{*}{\parbox{4em}{Flickr30k}} & \multirow{2}{*}{\parbox{5.5em}{\centering Average over 28 datasets}} & {} & \multirow{2}{*}{\parbox{4em}{\centering Backward Transfer}} & \multirow{2}{*}{\parbox{4em}{\centering ID Performance}} & \multirow{2}{*}{\parbox{4em}{\centering Forward Transfer}}  \\ 
{} & {} & \\
\midrule 
\multirow{2}{*}{\textbf{\benchname} (M)}  &  \Cumulative{}-\Full{} (w/o warmup)  & &$24.0$ & $20.2$ & $20.9$ & $17.9$ &  & $33.8$ & $26.4$ & $15.1$\\
&  \Cumulative{}-\Full{} (w warmup) & & 
$23.3$ & $20.1$ & $20.3$ & $17.6$ &  & $33.3$ & $26.1$ & $14.8$ \\
\multirow{2}{*}{\textbf{\benchname} (L)}  &  \Cumulative{}-\Full{} (w/o warmup)   &&  $48.9$ & $41.3$ & $50.9$ & $36.3$ &  & $62.1$ & $57.3$ & $41.2$\\
&  \Cumulative{}-\Full{} (w warmup) && $47.6$ & $40.6$ & $50.0$ & $35.2$ &  & $60.1$ & $53.0$ & $39.5$\\
\bottomrule[1.2pt]
\end{tabular}
}    
    \label{tab:lr_warmup}
\end{table}

\begin{table}[H]
    \centering
\caption{{\Cumulative{} experiments on \benchname{} (M) with different maximum learning rates for subsequent runs with first run fixed at LR 0.00025}. Our default choice for subsequent runs is 0.00025. Performance reported on ImageNet. At maximum learning rate $0.001$, the runs crashed with Nan in loss. }
\resizebox{0.5\textwidth}{!}{%
\begin{tabular}{lccccc}
\toprule[1.2pt]
\multirow{2}{*}{\textbf{Method}} & \multicolumn{4}{c}{\textbf{Max LR}} \\
\cmidrule{2-6}
& 0.00005 & 0.0001 & 0.00025  & 0.0005 & 0.001\\
\midrule
\Cumulative{}-\Full{}  & $16.3$ & $19.0$ & \update{$24.0$} & $10.1$ & --  \\
\bottomrule[1.2pt]
\end{tabular}
}
    \label{tab:max_lr_results}
\end{table}

When training with large batches, linear learning rate warm-up is typically employed to stabilize the start of the training when beginning from a random initialization~\citep{goyal2017accurate, steiner2021train}. However, when training sequentially by initializing models with checkpoints from the previous step, it remains unclear whether we should employ a learning rate warm up or not. Our observations highlight that while warm up is benefits for the first time step, not using warm up on subsequent runs performs better.  In particular, we observe that removing the warm up for the first training run hurts the final performance. On \benchname{} (\mlarge{}), we observe that training a ViT-B/16 with warm up on the first time step (i.e., 2016) gets $29.9$ zero-shot on Imagenet, whereas, without warm up ViT-B/16 achieves only $24.1$  zero-shot performance on Imagenet. 
\tabref{tab:lr_warmup} shows the final performance of models trained with and without warmup on subsequent time steps (after training on the first time step with warmup).  In particular, on \benchname{} (\mlarge{}), we observe $1.5\%$ accuracy gap on Imagenet and $4.3\%$ accuracy gap on dynamic ID retrieval performance on models trained with and without warm up. 

Hence, we default to using warmup when training on the first time step 
and not using it on the subsequent time steps with all methods 
except for training on \benchname{} (XL) where we 
add a smaller warm up (10\% of the warm up iterations used in first step) to stabilize training.

Next, we experiment with different maximum learning rate when training with cosine schedules. 
We ablate on \benchname{} (M) to investigate
how to change LR after training 
on data from the first time step. 
Unlike conventional pretraining and finetuning settings where LR is typically decreased for subsequent training, 
we observe that decaying maximum LR for subsequent steps in our setup hurts on static and dynamic tasks 
and consequently, we use the same 
maximum LR across our runs (see \tabref{tab:max_lr_results}). %

\subsection{\update{Preliminary experiments comparing random subsampling with other strategies to reduce buffer size}}

\update{In our preliminary experiments, we explored the efficacy of subsampling old data based on the alignment between text and image content from previous time steps. Specifically, when training a model at time step $t+1$, we used the model from the end of time step t to assess this alignment. We employed two distinct subsampling methods:}

\update{1. Retaining half of the data with the lowest alignment scores, based on the premise that these data points might be more challenging to learn and require additional gradient steps.}

\update{2. Retaining half of the data with the highest alignment scores, under the assumption that these represent higher quality data, as indicated by the stronger alignment between text and image pairs.}

\update{We applied these methods to the TiC-YFCC dataset and evaluated their performance against a baseline of random sampling. The outcomes revealed minimal differences: less than 0.2\% variation in Imagenet performance and under 0.5\% in dynamic retrieval performance across different time steps. Given that these minor improvements came with a significant computational cost—requiring a full forward pass to compute alignment post each training epoch—they exceeded our compute budget constraints. As a result, we opted for random sampling in our research. We leave investigation on improved subsampling techniques for future work. }

\subsection{\constcosine{}: An alternative learning rate schedule} \label{sec:const_cosine_schedule}

\begin{figure}[H]
    \centering
    \includegraphics[width=0.8\linewidth]{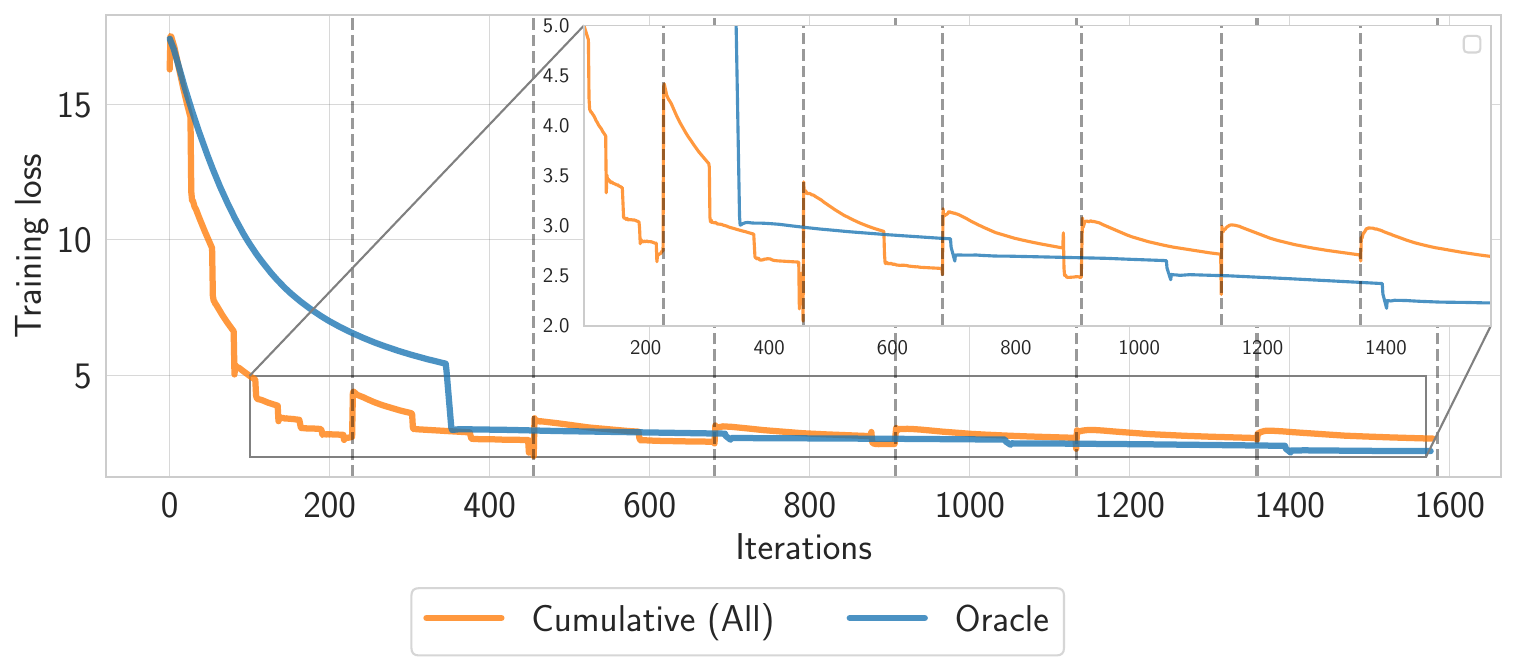}
    \caption{\textbf{Training loss increases every time the LR is reset to maximum LR for Cumulative.} {Loss comparison on training data with \Cumulative{} and \Oracle{} method.} \Cumulative{} is trained with a cyclic cosine schedule without warm up for sequential training runs. For \Cumulative{}, we plot the loss on training data, and as the training progresses, samples from new time steps are added to the training pool. For \Oracle{}, the training data is the union of data from all time steps and remains the same throughout the training.}
    \label{fig:loss_decay}
\end{figure}

The defacto LR schedule for training CLIP models is an initial linear increase to a maximum value, i.e., warm up, followed by a cosine decay
~\citep{radford2021learning, gadre2023datacomp}. 
In the main paper, we default to using cosine LR schedule for each sequential run, resulting in a cyclic schedule. We observe a significant increase in training loss early in subsequent runs when the LR is high. 
Comparing the loss on training data with \Cumulative{} and \Oracle{} methods, we observe that as training progresses the training loss increases every time the learning rate is increased to the maximum LR (\figref{fig:loss_decay}). 

It would be ideal for continual training to employ a learning rate schedule that is ``forward looking'', allowing us to continually train from a previous checkpoint without experiencing a significant increase in training loss. One desirable property of such a learning rate schedule would be its ability to adapt without requiring prior knowledge of the decay period. 

\begin{figure}[H]
    \centering
    \begin{subfigure}[b]{0.9\textwidth}
        \includegraphics[width=\textwidth]{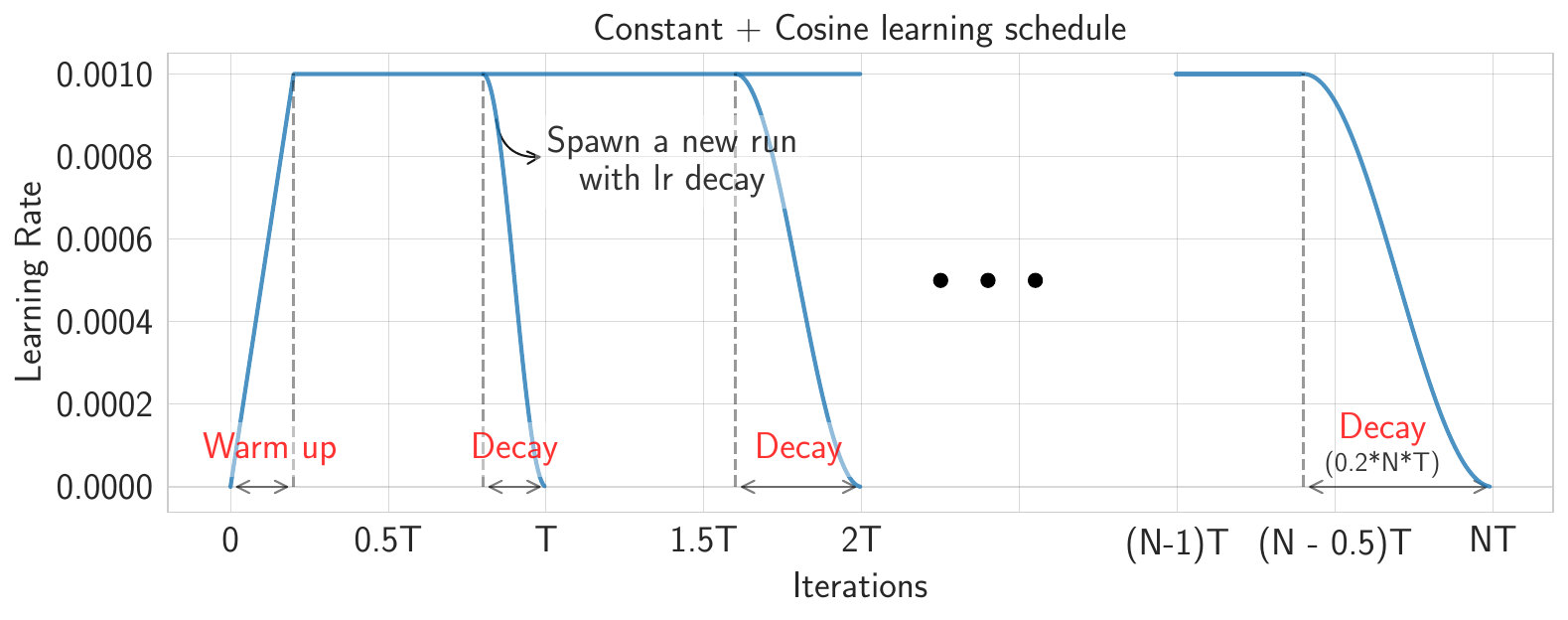}
    \end{subfigure}
    
    \caption{
\constcosine{}: Our proposed alternative forward-looking learning rate schedule schedule which trains one model with constant learning rate and decays the learning rate with cosine schedule only for a fraction of iterations before obtaining a deployable model. \constcosine{} schedule uses an extra compute budget than an \Oracle{} run because an extra training run is launched for the fraction of training when learning rate is decayed. 
}   
    \label{fig:const_cosine_lr}
\end{figure}

In our work, we perform preliminary experiments with the \emph{simplest} alternative, \constcosine{} where after the warm up period, we train with a constant learning rate and decay the learning rate only for a small fraction of training towards the end when we want a deployable model (\figref{fig:const_cosine_lr}).  
This allows us to continue training for subsequent runs from the checkpoint at the end of the constant learning rate schedule and decay the LR only in the end. 
For our experiments, we fix the decay period as $0.2$ of the total training iterations. Due to this, \constcosine{} schedule slightly increases the overall training budget of the \Cumulative{} runs when compared with cyclic cosine schedules. 

For \constcosine{}, we only ablate at relatively smaller scale datasets in our testbed (i.e., \cyfcc{}, \credcaps{}, and \benchname{} (\mmedium{})). For a fair comparison, we also re-run \Oracle{} methods with the same \constcosine{} schedule. 
Note that for \constcosine{} experiments, we use the same maximum LR as with the cosine schedule. 

We observe that training with \constcosine{} schedule significantly improves both \Cumulative{} and \Oracle{} as compared to their counterparts trained with cosine learning rates
\footnote{We also experimented with \constcosine{} schedule for \Oracle{} training on \benchname{} (\mlarge{}) and \benchname{} (\mxlarge{}). We observe that with a decay fraction of $0.2$, \constcosine{} achieves similar results to that of the cosine learning rate schedule. In particular, \constcosine{}  achieves $61.3$ on \mlarge{} and $73.0$ on \mxlarge{} versus Cosine schedule achieves $62.3$ on \mlarge{} and $73.3$ on \mxlarge{}. This highlights the potential of training with \constcosine{} schedule in scenarios where total training duration might be unknown apriori.}. 
Moreover, as expected, we do not observe jumps in training loss when training \Cumulative{} with \constcosine{} schedule. However, the gap between \Oracle{} and \Cumulative{} with  \constcosine{} doesn't decrease when compared with gap between \Oracle{} and \Cumulative{} with cosine learning rate schedules. 
This highlights that the jumps in the training loss observed while training with the cyclic cosine schedule might have benign effects on the final performance.

\begin{table}[H]
    \centering
    \caption{\textbf{Zero shot performance on Imagenet with \constcosine{} LR schedule.} 
    We observe that \constcosine{} improves over cyclic cosine LR schedule. However, the gap between cyclic cosine LR schedule and \constcosine{} for different LR schedules remains the same.   
    $^{**}$ denote methods that violate the compute budget.
    }
    \vspace*{-5pt}
\resizebox{.8\textwidth}{!}{%
\begin{tabular}{lccccccc}
\toprule[1.2pt]
\multirow{3}{*}{\textbf{Benchmark}} & {\textbf{Method}} & \multicolumn{2}{c}{\textbf{Cosine LR Schedule}} &  {} & \multicolumn{2}{c}{\textbf{\constcosine{} LR schedule}} \\[3pt]
{} & {}  & Compute (MACs) & ImageNet & {} & Compute (MACs) & ImageNet \\
\midrule
\multirow{2}{*}{\textbf{\cyfcc{}}} 
&  \Cumulative{}-\Full{} &   $3.4 \times 10^{18}$  & ${29.3}$ & {} & $4.4 \times 10^{18}$  & ${32.8}$ \\
&  \Oracle{}$^{**}$ &  $8.5\times 10^{18}$ & ${29.2}$ & {} & $8.5\times 10^{18}$ & ${33.2}$ \\  
\midrule 
\multirow{2}{*}{\textbf{\credcaps{}}} 
&  \Cumulative{}-\Full{} &   $3.4 \times 10^{18}$  & ${32.2}$ & {} & $4.4 \times 10^{18}$  & ${35.1}$ \\
&  \Oracle{}$^{**}$ &  $8.5\times 10^{18}$ & ${32.7}$ & {} & $8.5\times 10^{18}$ & ${36.2}$ \\  
\midrule 
\multirow{2}{*}{\textbf{\benchname{} (M)}} 
&  \Cumulative{}-\Full{} &   $3.0 \times 10^{18}$  & ${24.0}$ & {} & $3.6 \times 10^{18}$  & ${28.2}$ \\
&  \Oracle{}$^{**}$ &  $1.2\times 10^{19}$ & ${25.5}$ & {} & $1.2\times 10^{19}$ & ${28.9}$ \\  
\bottomrule[1.2pt]
\end{tabular}
}    
    \label{tab:const_cosine_results}
\end{table}

\subsection{OpenCLIP models obtained by retraining after removing any duplicate examples from the test set}

\begin{figure}[H]
    \centering
    \includegraphics[width=0.8\linewidth]{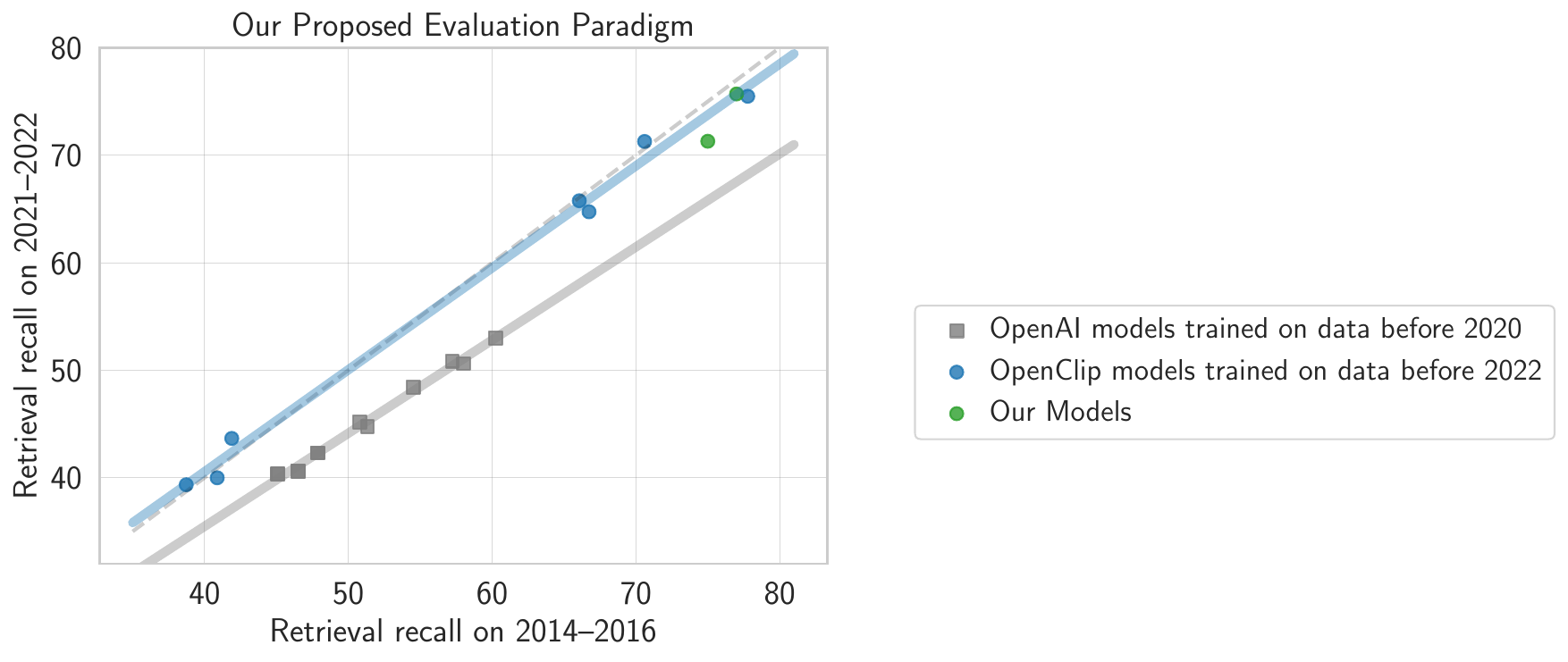}
    \caption{We replicate OpenCLIP models by training from scratch and removing duplicates from the evaluation dataset. We observe that trends continue to hold.}
    \label{fig:openclip_replicate}
\end{figure}

OpenCLIP models (e.g., models trained on Datacomp and LAION-5B) have been 
trained on data curated from Common Crawl. Since the retrieval tasks we 
constructed are built on top of data curated from Common Crawl, one may argue 
there is a possibility of  train/test overlap in our evaluations of OpenCLIP 
models. 
Thus, we retrain OpenCLIP models on DataComp datasets after removing the 
samples in our test sets. \Cref{fig:openclip_replicate} shows that the trends 
observed for OpenCLIP models holds for our retrained models.

\subsection{\update{Results on dynamic classification task}}

\update{In the main paper, we include results on our dynamic retrieval task. For completeness, here we include results on dynamic classification tasks on \benchname{} splits (\tabref{tab:main_results_classification}).  Along with including results on all nodes of ImageNet, we also include results on classification task restricted to classes in the ``motor vehicles'' subtree of ImageNet hierarchy. For the dynamic classification task, we observe trends similar to the dynamic retrieval task.}

\begin{table}[H]
    \centering
    \caption{\update{\textbf{Zero shot performance on our \benchname{}-\imagenet{} classification task.} 
    $^*$ and $^{**}$ denote methods that violate the compute budget.
    We tabulate forward/backward transfer 
    and ID performance on classification tasks (\secref{subsec:exp_protocol}). 
    For \benchname{} (XL), we include results with Bestpool 
    filtering.}}
\resizebox{1.0\textwidth}{!}{%
\begin{tabular}{lccccccccc}
\toprule[1.2pt]
\multirow{3}{*}{\textbf{Benchmark}} & \multirow{3}{*}{\textbf{Method}} &   \multirow{3}{*}{\parbox{4em}{\centering \textbf{Compute} (MACs)}} & \multicolumn{3}{c}{\textbf{Dynamic Retrieval Tasks (All)}} &  {} & \multicolumn{3}{c}{\textbf{Dynamic Retrieval Tasks (`Motor Vehicles')}} \\[3pt]
{} & {} & {} &  \multirow{2}{*}{\parbox{4em}{\centering Backward Transfer}} & \multirow{2}{*}{\parbox{4em}{\centering ID Performance}} & \multirow{2}{*}{\parbox{4em}{\centering Forward Transfer}} & {} &  \multirow{2}{*}{\parbox{4em}{\centering Backward Transfer}} & \multirow{2}{*}{\parbox{4em}{\centering ID Performance}} & \multirow{2}{*}{\parbox{4em}{\centering Forward Transfer}}  \\ 
{} & {} & \\
\midrule
\multirow{7}{*}{\textbf{\benchname} (M)} 
& \Sequential{} &  $3.0 \times 10^{18}$ &  $15.9$ & $13.3$ & $9.9$ & & $34.5$ & $30.0$ & $22.6$\\
& \Patching{} &  $3.0 \times 10^{18}$  &   $15.6$ & $13.1$ & $9.7$ & & $34.4$ & $29.2$ & $22.1$
 \\
& \Cumulative{}-\Exponential{} &  $3.0 \times 10^{18}$ &   $17.6$ & $14.4$ & $10.4$ & & $36.6$ & $30.9$ & $23.5$
 \\
& \Cumulative{}-\Equal &  $3.0 \times 10^{18}$  & $17.5$ & $14.2$ & $10.4$  & & $36.4$ & $31.1$ & $23.5$
 \\
&  \Cumulative{}-\Full{} &   $3.0 \times 10^{18}$ &  $18.3$ & $14.7$ & $10.6$ & & $38.2$ & $31.7$ & $23.7$
\\
\rowcolor{gray!10}  & \LwF{}$^{*}$ &  $3.8 \times 10^{18}$ &   $16.0$ & $13.5$ & $9.9$ & & $35.1$ & $30.7$ & $23.3$

 \\
\rowcolor{gray!10} &  \Cumulative{}-\Full{}$^{*}$ &  $3.9 \times 10^{18}$ &  $20.7$ & $16.0$ & $10.9$ & & $40.4$ & $32.3$ & $23.9$
\\
\rowcolor{gray!10} & \Oracle{}$^{**}$ &  $1.2 \times 10^{19}$ &  $19.2$ & $15.2$ & $10.7$ & & $38.7$ & $31.9$ & $23.5$
\\
\midrule  
\multirow{7}{*}{\textbf{\benchname} (L)}   
& \Sequential{} & $2.7 \times 10^{19}$ &  
$38.3$ & $36.9$ & $33.3$
& & $58.4$ & $55.6$ & $49.7$
\\
& \Patching{} & $2.7 \times 10^{19}$ &
    $38.6$ & $36.8$ & $33.3$
 & & $58.3$ & $54.9$ & $49.3$
\\
& \Cumulative{}-\Exponential{} & $2.7 \times 10^{19}$
& $40.2$ & $37.9$ & $34.2$
 & & $60.7$ & $56.8$ & $51.1$
\\
& \Cumulative{}-\Equal & $2.7 \times 10^{19}$ & 
  $40.6$ & $38.0$ & $34.2$
 & & $60.7$ & $56.8$ & $50.8$
\\
&  \Cumulative{}-\Full{} & $2.7 \times 10^{19}$ &    
  $41.3$ & $38.3$ & $34.4$
 & & $61.4$ & $56.6$ & $50.9$
\\
\rowcolor{gray!10} &  \Cumulative{}-\Full{}$^{*}$ &  $4.1\times 10^{19}$ &  
  $43.0$ & $39.2$ & $34.6$
 & & $62.7$ & $57.5$ & $51.1$
\\
\rowcolor{gray!10}  &  \Oracle{}$^{**}$ & $1.1 \times 10^{20}$ & 
  $43.8$ & $40.0$ & $35.2$
 & & $62.6$ & $56.8$ & $50.7$
\\
\midrule
\multirow{3}{*}{\textbf{\benchname} (XL)} & \Sequential{} &  $2.7\times 10^{20}$ & 
$55.4$ & $55.1$ & $53.3$  & & $67.8$ & $66.0$ & $63.5$

\\
&  \Cumulative{}-\Full{} &  $2.7\times 10^{20}$ & 
$58.5$ & $56.7$ & $54.3$  & & $70.2$ & $67.4$ & $63.8$

\\
\rowcolor{gray!10} &  \Cumulative{}-\Full{}$^{*}$ &  $3.5\times 10^{20}$ & 
 $58.8$ & $56.9$ & $54.3$  & & $70.5$ & $67.5$ & $63.8$
\\
\bottomrule[1.2pt]
\end{tabular}
}    
    \label{tab:main_results_classification}
\end{table}

\subsection{\update{Addressing differences between \Sequential{} and \Cumulative{}-\Full{} between \cyfcc{} and \benchname{}}} 

\update{In \tabref{tab:main_results}, we observe differences in the behavior of \Sequential{} and \Cumulative{}-\Full on \cyfcc{} when compared with \benchname{}. For instance, differences between the ID performance between  \Sequential{} and \Cumulative{}-\Full{} is larger in \cyfcc{} than in \benchname{} (M). Similar observations hold true for backward transfer performance. In this section, we explain the underlying causes for these differences.}
 
\update{We identify two primary reasons: 
\begin{enumerate}[label=(\roman*), leftmargin=*]
    \item the nature of the distribution shift observed in \cyfcc{}. We observe that models trained with \Sequential{} on \cyfcc{} suffer from relatively larger drops on old-time steps than \benchname{} (M) due to catastrophic forgetting  (see 
\figref{fig:dynamic_evals_heatmap}).
\item compute used at each time step per data available at each time step is different for these bencmarks. Overall YFCC is 2x smaller than Tic-Datacomp (M) but the compute we used in both TiC-YFCC and TiC-Datacomp setup is of similar order (in fact, it is slightly higher in TiC-YFCC). We re-ran the experiments for Tic-YFCC by reducing the compute. In the updated runs, we observe that the gap between ID performances of Sequential and Cumulative-All vanishes.  
\end{enumerate}}

\begin{table}[H]
    \centering
    \caption{\update{\textbf{Zero shot retrieval performance on \cyfcc{} with \Sequential{} and \Cumulative{}-\Full{} with reduced compute.} 
    }}
    \vspace*{-5pt}
\resizebox{\textwidth}{!}{%
\begin{tabular}{lcccccccccc}
\toprule[1.2pt]
\multirow{4}{*}{\textbf{Benchmark}} & \multirow{4}{*}{\textbf{Method}} &   \multicolumn{4}{c}{\textbf{Dynamic Retrieval Tasks (original compute)}} & \multicolumn{4}{c}{\textbf{Dynamic Retrieval Tasks (reduced compute)}} \\[3pt]
{} & {} &  \multirow{3}{*}{\parbox{4em}{\centering \textbf{Compute} (MACs)}}  & \multirow{3}{*}{\parbox{4em}{\centering Backward Transfer}} & \multirow{3}{*}{\parbox{4em}{\centering ID Performance}} & \multirow{3}{*}{\parbox{4em}{\centering Forward Transfer}} & \multirow{3}{*}{\parbox{4em}{\centering \textbf{Compute} (MACs)}} & \multirow{3}{*}{\parbox{4em}{\centering Backward Transfer}} & \multirow{3}{*}{\parbox{4em}{\centering ID Performance}} & \multirow{3}{*}{\parbox{4em}{\centering Forward Transfer}}  \\ 
{} & {} & \\
{} & {} & \\
\midrule
\multirow{2}{*}{\textbf{\cyfcc{}}} & \Sequential{} & $3.4 \times 10^{18}$ & $42.2$ & $48.4$ & $23.7$ & $1.5\times 10^{18}$ & $27.0$ & $42.0$ & $15.7$\\
&  \Cumulative{}-\Full{} &   $3.4 \times 10^{18}$  & ${66.4}$ & ${60.2}$ & ${27.6}$ & $1.5\times 10^{18}$ & $46.3$ & $38.7$ & $17.3$ \\
\bottomrule[1.2pt]
\end{tabular}
}    
    \label{tab:main_results_reduced_compute}
    \vspace*{-8pt}
\end{table}

\begin{figure}
    \centering
    \begin{subfigure}[b]{\textwidth}
        \centering
    
        \includegraphics[width=0.23\linewidth]{figures/appendix_figures/Sequential_dist_shift_results_yfcc15m.pdf}
        \includegraphics[width=0.23\linewidth]{figures/appendix_figures/Cumulative-All_dist_shift_results_yfcc15m.pdf}
        \caption{\cyfcc{} (original compute).}
    \end{subfigure}

        \begin{subfigure}[b]{\textwidth}
        \centering
        \includegraphics[width=0.23\linewidth]{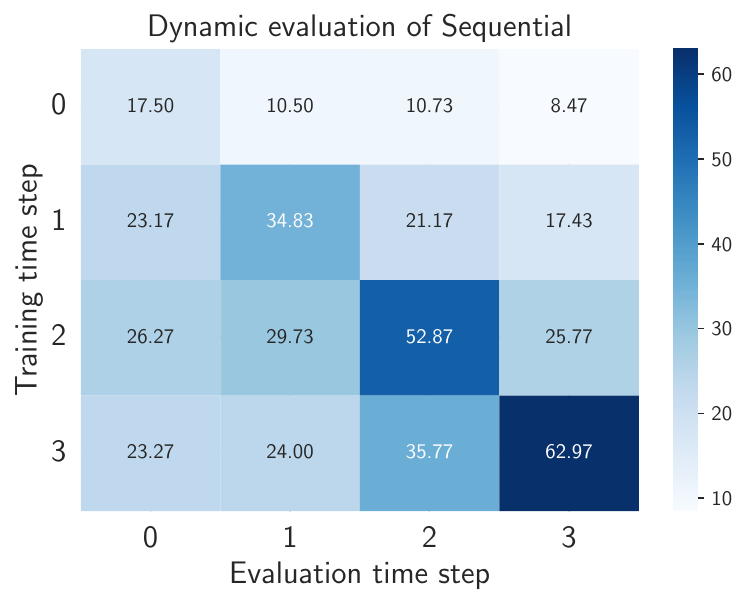}
        \includegraphics[width=0.23\linewidth]{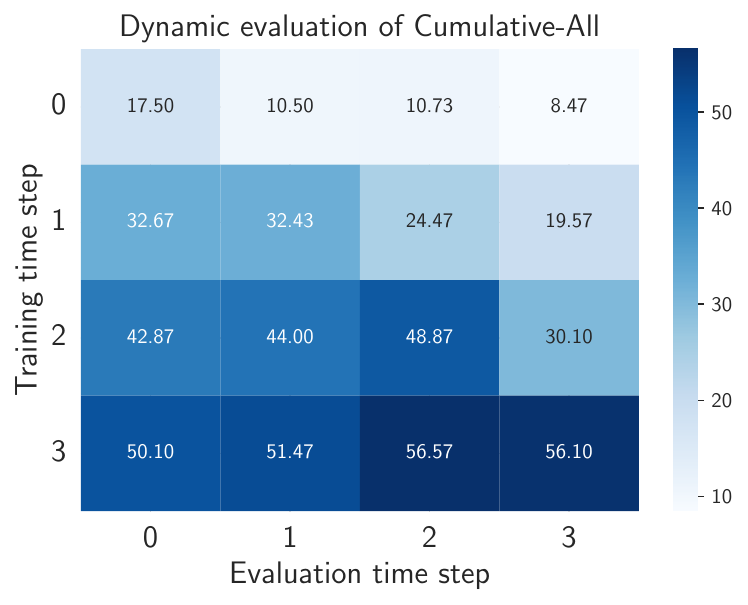}
        \caption{\cyfcc{} (reduced compute).}
    \end{subfigure}

    \caption{\update{Dynamic retrieval evaluation results with \Sequential{}, \Cumulative{}-\Full{} on \cyfcc{} with reduced compute.}}
    \label{fig:dynamic_evals_heatmap_compute}
\end{figure}

\section{Additional Benchmark Details}

\subsection{Filtering ablations on \benchname{}} \label{app:tic_datacomp_filtering}

\begin{figure}[H]
    \centering
    \begin{subfigure}{0.4\linewidth}
        \centering
        \includegraphics[width=\linewidth]{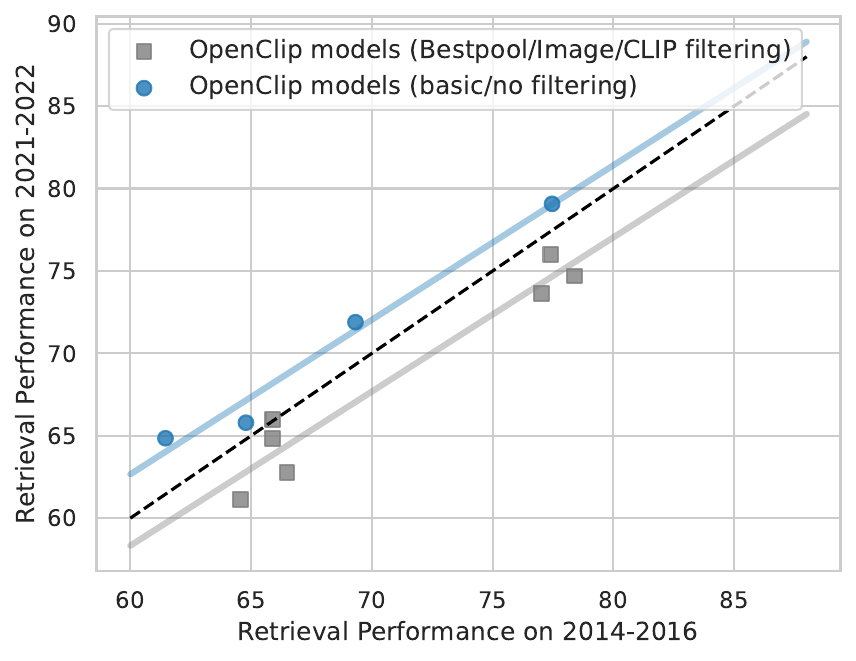}
    \end{subfigure}
    \begin{subfigure}{0.4\linewidth}
        \centering
        \includegraphics[width=\linewidth]{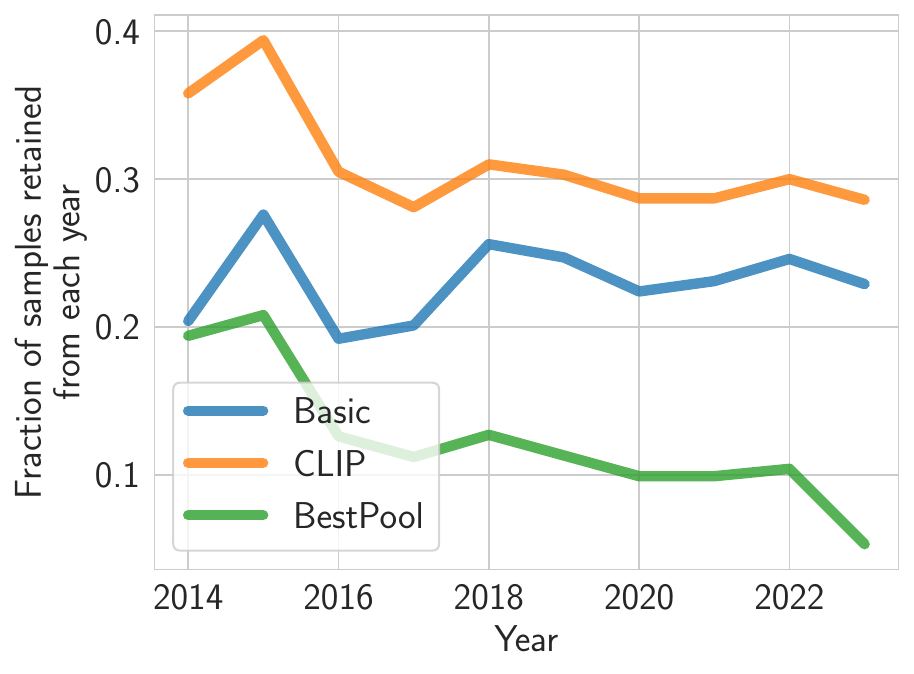}
    \end{subfigure}
    \caption{(Left) Gap in retrieval performance for different OpenCLIP models that use different filtering techniques. (Right) Reduction in \benchname{} data at different times with different filtering techniques. This clearly highlights that there is a selection bias towards retaining more old data for CLIP/BestPool filtering. No such bias exists for basic filtering. }
    \label{fig:filtering_compare}
\end{figure}

For Basic Filtering, \citet{gadre2023datacomp} performs the following three 
steps: filter by English language (using fasttext~\citep{joulin2017bag}), 
filter by caption length over two words and 5 characters, and filter by image 
sizes with smallest dimensions over 200 pixels and aspect ratio above 3.  We do 
not default to other filtering techniques that use off-the-shelf CLIP models 
from \citet{gadre2023datacomp} to avoid biasing dataset selection from each 
time step. In \figref{fig:filtering_compare}, we show that ``Bestpool'' 
filtering (which filters image-text pairs with CLIP scores and ImageNet image 
embeddings) biases dataset selection to preferring old time step data over new 
timestamp data. Moreover, we also show that models trained with Bestpool 
filtering is less robust when evaluated on our dynamic tasks from 2021-2022 (\figref{fig:filtering_compare}).  
Nevertheless, for completeness and to highlight the significance of our findings 
even for state-of-the-art filtering techniques, we perform continual 
learning experiments with Bestpool filtering at \mxlarge{} scale which is included in the main paper. In \appref{subsec:basic_filtering_xlarge}, we include results with Basic filtering at \mxlarge{}.

\subsection{Static Datasets considered for evaluation} \label{app:static_datasets}

\begin{table}[H]
\caption{Evaluation tasks borrowed from \citet{gadre2023datacomp}.}
\setlength\tabcolsep{4.5pt}
    \renewcommand{\arraystretch}{1.1}
\resizebox{\textwidth}{!}{
\centering
\begin{tabular}{lllrrrc}
\toprule
Task type & Dataset & Task & Test set size & Number of classes & Main metric \\ \midrule
\cellcolor{white} & Food-101 \cite{food101} & Food recognition & 25,250 & 101 & accuracy \\
\cellcolor{white} & GTSRB \cite{gtsrb} & Traffic sign recognition & 12,630 & 43 & accuracy \\
\cellcolor{white} & ImageNet 1k \cite{deng2009imagenet} & Visual recognition & 50,000 & 1,000 & accuracy \\
\cellcolor{white} & ImageNet Sketch \cite{imagenetsketch} & Visual recognition & 50,889 & 1,000  & accuracy \\
\cellcolor{white} & ImageNet V2 \cite{imagenetv2} & Visual recognition & 10,000 & 1,000  & accuracy \\
\cellcolor{white} & ImageNet-A \cite{imageneta_and_imageneto} & Visual recognition & 7,500 & 200 & accuracy  \\
\cellcolor{white} & ImageNet-O \cite{imageneta_and_imageneto} & Visual recognition & 2,000 & 200  & accuracy \\
\cellcolor{white} & ImageNet-R \cite{imagenetr} & Visual recognition & 30,000 & 200  & accuracy \\
\cellcolor{white} & KITTI distance \cite{kitti,vtab} & Distance prediction & 711 & 4  & accuracy\\
\cellcolor{white} & MNIST \cite{lecun1998mnist} & Digit recognition & 10,000 & 10 & accuracy  \\
\cellcolor{white} & ObjectNet \cite{objectnet} & Visual recognition & 18,574 & 113  & accuracy \\
\cellcolor{white} & Oxford Flowers-102 \cite{flowers102} & Flower recognition & 6,149 & 102 & mean per class \\
\cellcolor{white} & Oxford-IIIT Pet \cite{pets,vtab} & Pet classification & 3,669 & 37 & mean per class \\
\cellcolor{white} & Pascal VOC 2007 \cite{pascal-voc-2007} & Object recognition & 14,976 & 20 & accuracy \\
\cellcolor{white} & PatchCamelyon \cite{patchcamelyon,vtab} & Metastatic tissue cls. & 32,768 & 2  & accuracy\\
\cellcolor{white} & Rendered SST2 \cite{vtab} & Sentiment classification & 1,821 & 2  & accuracy \\
\cellcolor{white} & RESISC45 \cite{resisc45,vtab} & Satellite imagery recognition & 6,300 & 45 & accuracy  \\
\cellcolor{white} & Stanford Cars \cite{cars} & Vehicle recognition & 8,041 & 196 & accuracy  \\
\cellcolor{white} & STL-10 \cite{stl10} & Visual recognition & 8,000 & 10  & accuracy \\
\cellcolor{white} & SUN-397 \cite{sun397} & Scene recognition & 108,754 & 397 & accuracy  \\
\cellcolor{white} & SVHN \cite{svhn,vtab} & Digit recognition & 26032 & 10 & accuracy  \\
\cellcolor{white} & iWildCam \cite{beery2020iwildcam,wilds2021} & Animal recognition & 42,791 & 182 & macro F1 score  \\
\cellcolor{white} & Camelyon17 \cite{bandi2018detection,wilds2021} & Metastatic tissue cls. & 85,054 & 2 & accuracy \\
\cellcolor{white} \multirow{-25}{*}{Classification}&  FMoW \cite{christie2018functional,wilds2021} & Satellite imagery recognition & 22,108 & 62 & worst-region acc.  \\
\midrule
 \cellcolor{white} \multirow{1}{*}{Retrieval} & Flickr30k \cite{flickr30k} & Image and text retrieval & 31,014 & N/A & R@1  \\
\bottomrule
\end{tabular}
}
\label{tab:eval-sets}
\end{table}

\subsection{Our Benchmark Statistics} \label{app:dataset_stats}

\begin{figure}[H]
    \centering
    \includegraphics[width=0.3\linewidth]{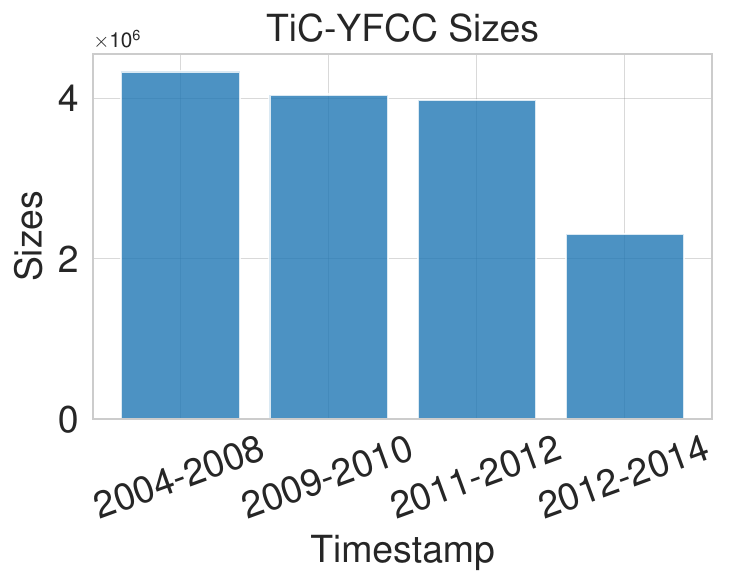}
    \includegraphics[width=0.3\linewidth]{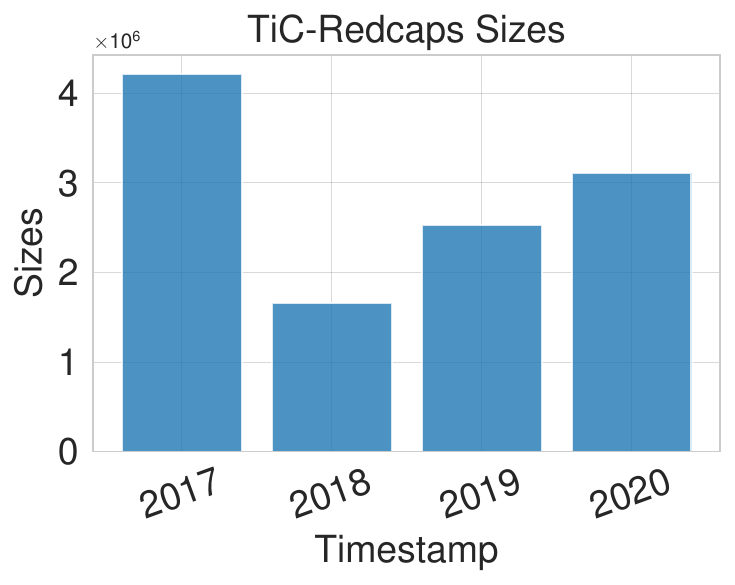}
    \includegraphics[width=0.37\linewidth]{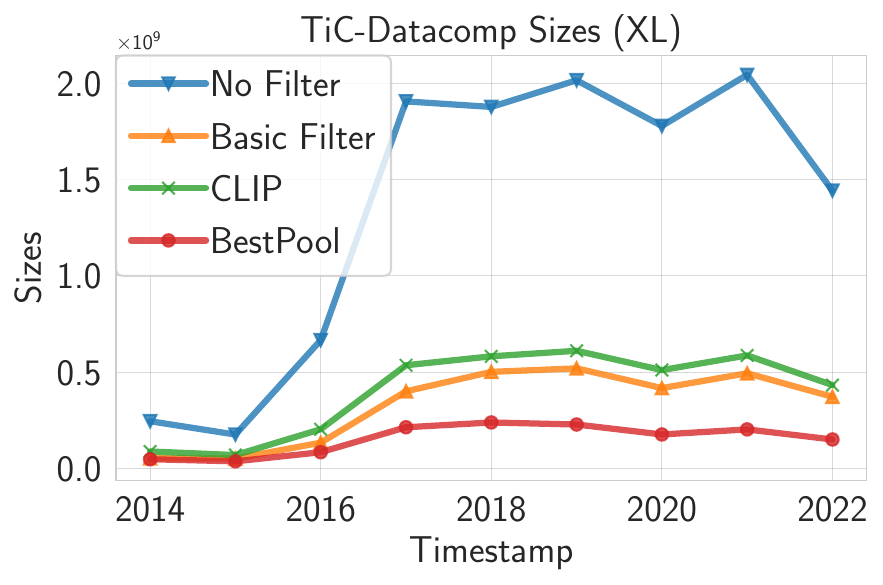}
    
    \caption{Number of examples in each year in our benchmarks. }
    \label{fig:datset_sizes}
\end{figure}

\begin{figure}[H]
    \centering
    \includegraphics[width=0.5\linewidth]{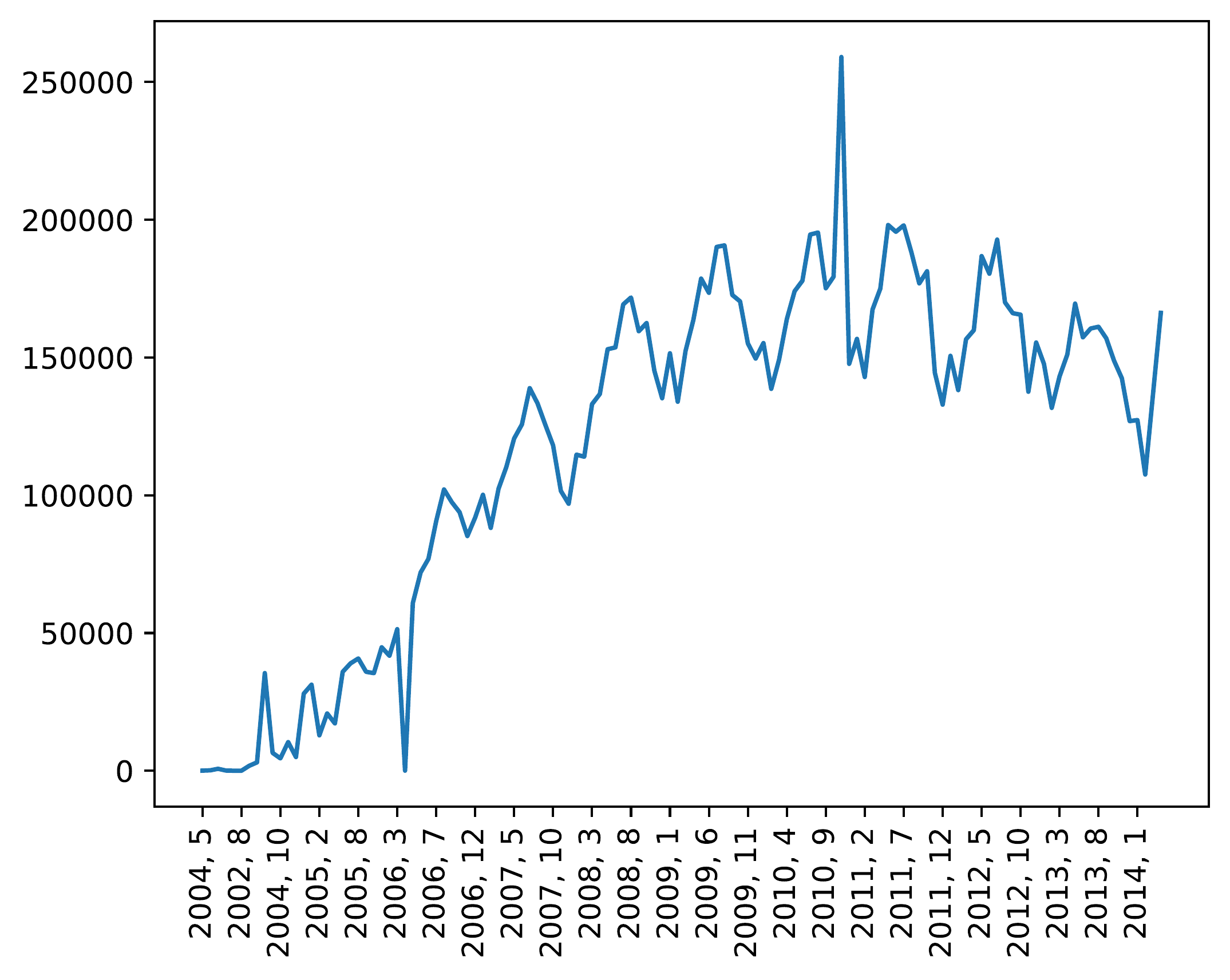}
    \caption{Number of examples in each year in original YFCC 15M. X-axis the upload month and y-axis is the number of examples in that month.}
    \label{fig:yfcc_original}
\end{figure}

In this section, we discuss statistics of our constructed benchmarks. \figref{fig:datset_sizes} summarizes \credcaps{}, \cyfcc{} and \benchname{} dataset sizes. \figref{fig:yfcc_original} summarizes original YFCC dataset  sizes. \tabref{tab:redcaps_table}, \tabref{tab:yfcc_table} and \tabref{tab:datacomp_table} present the exact numbers for these datasets. For \benchname{}, we only discuss the sizes at \mxlarge{} scale.

\begin{table}[H]
\caption{Number of examples in \credcaps{} in each year.} 
    \vspace*{-7pt}
\centering
    
\resizebox{0.5\textwidth}{!}{%
\begin{tabular}{lcccc}
\toprule[1.2pt]
\multirow{2}{*}{\textbf{Dataset}} & \multicolumn{4}{c}{\textbf{Year}} \\
\cmidrule{2-5}
& 2017 & 2018 & 2019 & 2020\\
\midrule
\credcaps{} & 4,220,262 & 1,660,003 & 2,526,575 & 3,115,715 \\
\bottomrule[1.2pt]
\end{tabular}
}
    \label{tab:redcaps_table}
    \vspace*{-8pt}
\end{table}

\begin{table}[H]
\caption{Number of examples in \cyfcc{} in each year.} 
    \vspace*{-7pt}
\centering
    
\resizebox{0.5\textwidth}{!}{%
\begin{tabular}{lcccc}
\toprule[1.2pt]
\multirow{2}{*}{\textbf{Dataset}} & \multicolumn{4}{c}{\textbf{Year}} \\
\cmidrule{2-5}
& 2004--2008 & 2009--2010 & 2011--2012 & 2012--2014\\
\midrule
\cyfcc{} & 4,337,727 & 4,050,166 & 3,976,339 & 2,312,753 \\
\bottomrule[1.2pt]
\end{tabular}
}
    \label{tab:yfcc_table}
    \vspace*{-8pt}
\end{table}

\begin{table}[H]
\caption{Number of examples in \benchname{} in each year before filtering.} 
    \vspace*{-7pt}
\centering
    
\resizebox{1.\textwidth}{!}{%
\begin{tabular}{lccccccccc}
\toprule[1.2pt]
\multirow{2}{*}{\textbf{Dataset}} & \multicolumn{9}{c}{\textbf{Year}} \\
\cmidrule{2-10}
& 2014 & 2015 & 2016 & 2017 & 2018 & 2019 & 2020 & 2021 & 2022\\
\midrule
\benchname{} (no filter) & 244,802,598 & 175,648,045 & 666,019,511 & 1,906,357,755 & 1,877,561,875 & 2,016,011,588 & 1,778,751,066 & 2,044,463,701 & 1,442,233,121\\
\benchname{} (basic filter) & 52,764,775 & 50,757,898 & 133,333,267 & 400,225,598 & 501,347,511 & 519,575,760 & 417,067,014 & 494,038,122& 371,748,613 \\
\bottomrule[1.2pt]
\end{tabular}
}
    \label{tab:datacomp_table}
    \vspace*{-8pt}
\end{table}

\update{Next, we tabulate the number of examples in our retrieval evaluation datasets. Since the evaluation dataset sizes are different at different time steps, we subsample the dataset to a fixed size before performing retrieval evaluations.  On \cyfcc{} and \credcaps{}, we randomly sampled 1000 image-text pairs from these evaluation datasets. For \benchname{}, we randomly sample 4000 image-text pairs.  We repeat this process for 3 seeds and report the aggregated performance.}

\begin{table}[H]
\caption{Number of retrieval evaluation examples in \credcaps{} in each year.} 
    \vspace*{-7pt}
\centering
    
\resizebox{0.5\textwidth}{!}{%
\begin{tabular}{lcccc}
\toprule[1.2pt]
\multirow{2}{*}{\textbf{Dataset}} & \multicolumn{4}{c}{\textbf{Year}} \\
\cmidrule{2-5}
& 2017 & 2018 & 2019 & 2020\\
\midrule
\credcaps{} & 31,316 & 42,539 & 16,738 & 25,565 \\
\bottomrule[1.2pt]
\end{tabular}
}
    \label{tab:redcaps_table_evals}
    \vspace*{-8pt}
\end{table}

\begin{table}[H]
\caption{Number of retrieval evaluation examples in \cyfcc{} in each year.} 
    \vspace*{-7pt}
\centering
    
\resizebox{0.5\textwidth}{!}{%
\begin{tabular}{lcccc}
\toprule[1.2pt]
\multirow{2}{*}{\textbf{Dataset}} & \multicolumn{4}{c}{\textbf{Year}} \\
\cmidrule{2-5}
& 2004--2008 & 2009--2010 & 2011--2012 & 2012--2014\\
\midrule
\cyfcc{} & 43,820 & 40,909  & 40,165 & 23,354 \\
\bottomrule[1.2pt]
\end{tabular}
}
    \label{tab:yfcc_table_evals}
    \vspace*{-8pt}
\end{table}

\begin{table}[H]
\caption{Number of retrieval evaluation examples in \benchname{} in each year before filtering.} 
    \vspace*{-7pt}
\centering
    
\resizebox{0.8\textwidth}{!}{%
\begin{tabular}{lccccccccc}
\toprule[1.2pt]
\multirow{2}{*}{\textbf{Dataset}} & \multicolumn{7}{c}{\textbf{Year}} \\
\cmidrule{2-8}
& 2016 & 2017 & 2018 & 2019 & 2020 & 2021 & 2022\\
\midrule
\benchname{}  & 23,085 & 39,289 & 50,450 & 53058 & 42,239 & 49,841 & 38,051\\
\bottomrule[1.2pt]
\end{tabular}
}
    \label{tab:datacomp_table_evals}
    \vspace*{-8pt}
\end{table}

\subsection{Compute Constraints for Different Datasets} \label{app:compute_constraints}

We closely follow compute budget constraints from \citet{gadre2023datacomp}. In particular, on \benchname{}, we restrict to using exactly the same amount of overall compute as fixed in \citet{gadre2023datacomp}. Below we list exact total MACs on each dataset: 

\begin{itemize}
    \item \cyfcc{}: Total MACs:  $3.4 \times 10^{18}$
    \item \credcaps{}: Total MACs:  $3.4 \times 10^{18}$
    \item \benchname{} \mmedium : Total MACs:  $3.0 \times 10^{18}$
    \item \benchname{} \mlarge:  Total MACs:   $2.7 \times 10^{19}$
    \item \benchname{} \mxlarge:  Total MACs: $2.7\times 10^{20}$
\end{itemize}

For a ViT-B architecure, these values correspond to 20k iterations on  \cyfcc{} (batch size: 8192), \credcaps{}  (batch size: 8192), 35k iterations on  \benchname{} (M)  (batch size: 4096),  157k iterations on  \benchname{} (L)  (batch size: 8192), and  143.5k iterations on  \benchname{} (XL)  (batch size: 90100). We divide these iterations equally among all time steps.

\subsection{Creation Pipeline for Evaluation Datasets} \label{app:evaluation_examples}

\textbf{\benchname{}-\retreival{} {} {}} To create a retrieval task, we sample 
a batch of IID image-text pairs 
from different timestamps and evaluate text retrieval performance given the 
corresponding image (similarly, image retrieval given the corresponding text).  
Alongside general evaluations, we also construct datasets from specific 
domains, e.g., Covid-19 subset and Flickr subset. To create Covid-19, we filter 
the dataset to only retain pairs where the caption contains a mention of 
"covid".  This search process restricts the data to time only after 2019. For 
the Flickr subset, we filter the dataset to only retain pairs where the 
corresponding ``url'' contains data from Flickr. 

\textbf{\benchname-\imagenet{} {} {}} We create our dynamic classification dataset \benchname-\imagenet{} with ImageNet classes from the CommonPool data augmented with temporal information. Our construction process draws inspiration from the LAIONet construction process described in \citet{shirali2023makes}. In particular, we first filter examples where the corresponding caption contains one and only one of the synsets of ImageNet-1K.
We also apply additional basic filtering~\citep{gadre2023datacomp} to make sure that images are of at least 200 size in smallest dimension and the caption contains at least 2 words and 5 characters. 
After filtering for examples with ImageNet synsets, we only retain examples where the similarity---as evaluated by an off-the-shelf sentence embedding model~\citep{reimers-2019-sentence-bert}---between imagenet synset definition and the caption exceeds a threshold of $0.5$. 
The goal of this filtering step is to restrict examples with ``high'' alignment between caption and imagenet synset definition.  
This last step differs from the LAIONet construction. 
Crucially, unlike LAIONet, we do not filter the image-text pairs with CLIP similarity scores to avoid biasing the dataset selection process.

\subsection{Distribution Shift Analysis on Proposed benchmarks} \label{app:dist_shift_analysis}

\begin{figure}[H]
    \centering
    \begin{subfigure}{0.4\linewidth}
        \centering
        \includegraphics[width=\linewidth]{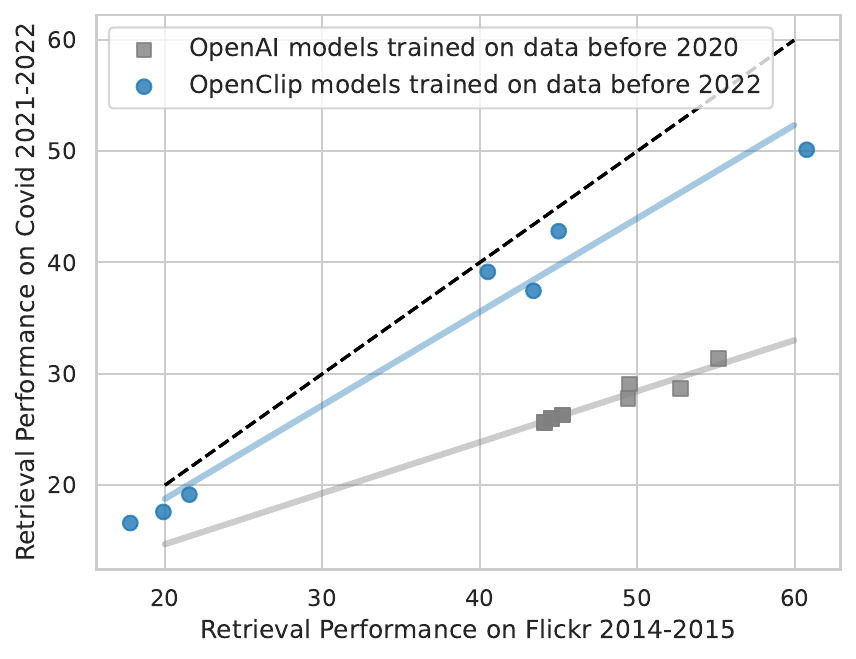}
    \end{subfigure}
    \begin{subfigure}{0.4\linewidth}
        \centering
        \includegraphics[width=\linewidth]{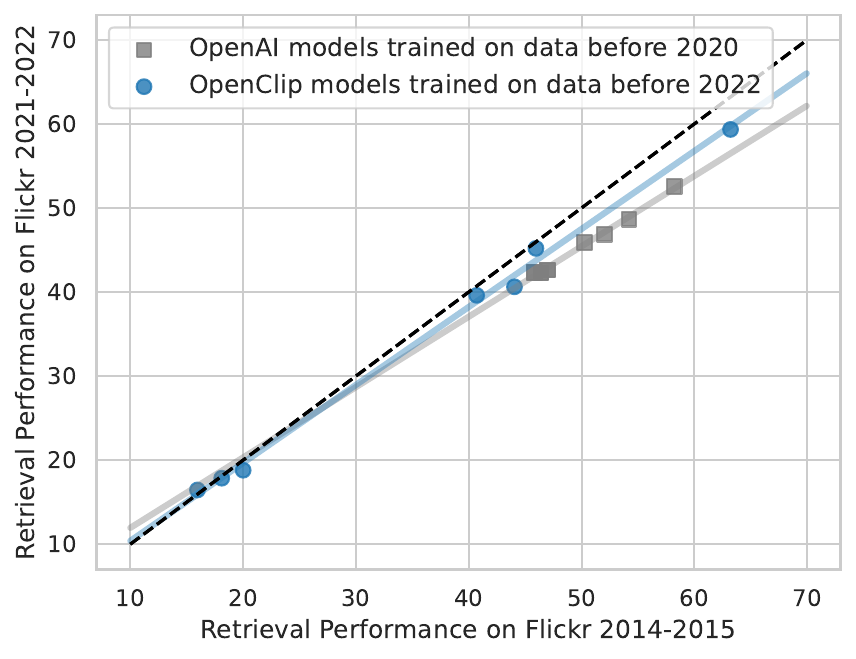}
    \end{subfigure}
    \caption{(Left) Comparison of retrieval performance on COVID queries versus Flickr queries (construction described in \appref{app:evaluation_examples}). (Right) Comparison on old Flickr versus new Flickr data. Clearly, we observe that while gap on old versus new flickr data is small, the gap is significantly larger on Covid queries.}
    \label{fig:covid_flickr_compare}
\end{figure}

\begin{figure}[H]
    \centering
    \begin{subfigure}{0.4\linewidth}
        \centering
        \includegraphics[width=\linewidth]{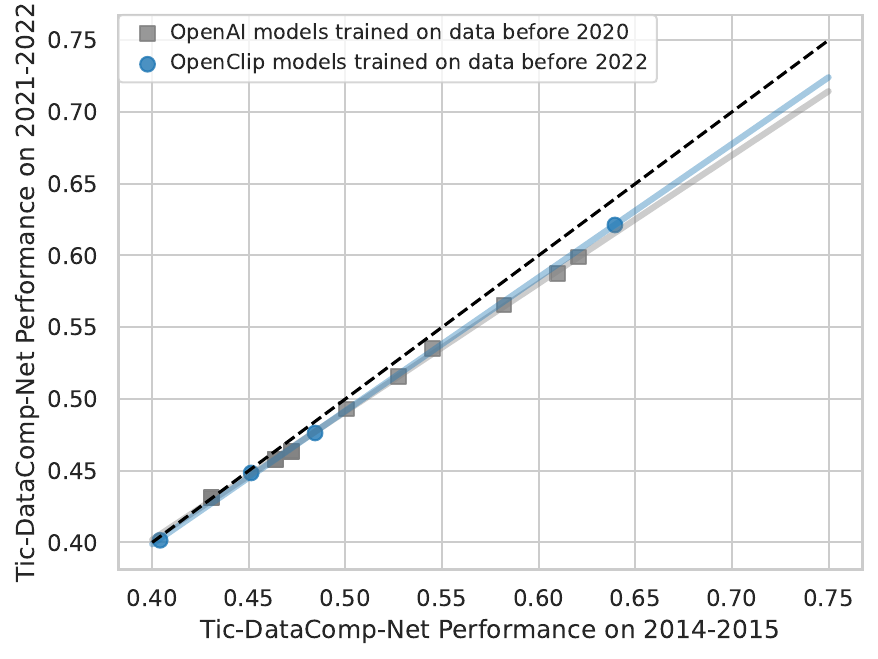}
    \end{subfigure}
    \begin{subfigure}{0.4\linewidth}
        \centering
        \includegraphics[width=\linewidth]{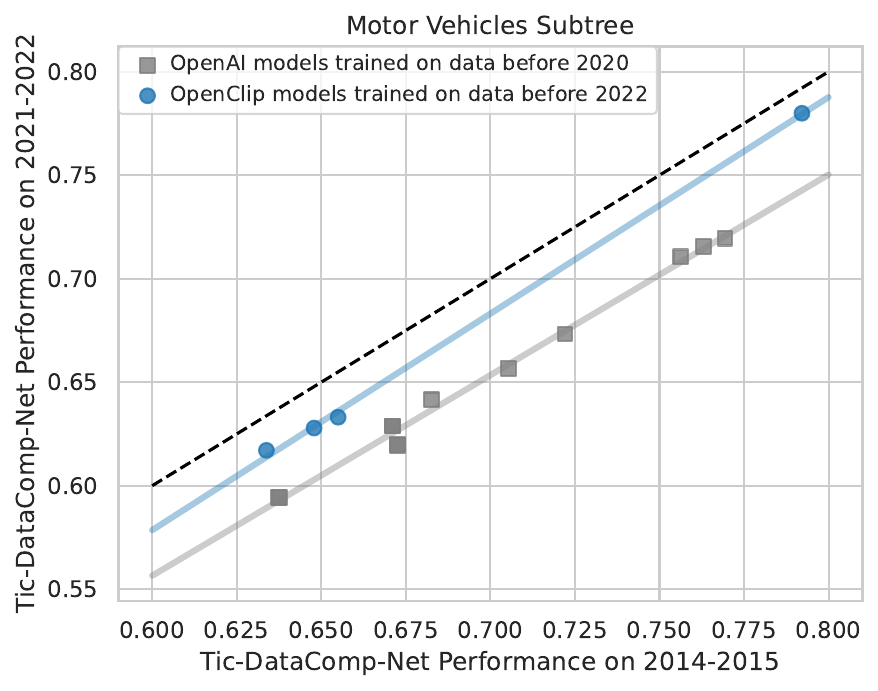}
    \end{subfigure}
    \caption{(Left) Comparison on old versus new data from \benchname{}-Net. (Right) Comparison on motor vehicles node from  \benchname{}-Net. For our classification task, we observe a very small drop ($\approx 1\%$) when averaged across all categories. However, we observe a substantial gap  on classes in ``motor vehicle'' subtree, when comparing OpenAI  and OpenCLIP models. These findings highlight that while overall ImageNet classes  may remain timeless, certain categories tend to evolve faster than others.  }
    \label{fig:imagemet_compare}
\end{figure}

\textbf{\benchname{} analysis through the lens of constructed evaluation tasks {} {}}
Here, we compare performance of 
OpenAI and OpenCLIP models on our datasets.
We observe a 
significant performance gap between 
OpenAI and OpenCLIP models 
on our dynamic retrieval task (\figref{fig:intro}). 
This gap widens notably on retrieval queries where captions mention COVID-19.
On the other hand,  OpenAI and 
OpenCLIP models exhibit similar robustness for 
retrieval on data coming from Flickr highlighting 
that data from some domains do not exhibit shifts 
that cause performance drops.
For our classification task, we observe a very small drop ($\approx 1\%$) when averaged across all categories.
However, we observe a substantial gap 
on specific subtrees in ImageNet. For example,  
classes in ``motor vehicle'' subtree 
show an approximate $7\%$ performance drop, when comparing OpenAI 
and OpenCLIP models. These findings highlight that while overall ImageNet classes 
may remain timeless, certain categories tend to evolve faster than others.  
Our qualitative and quantitative analysis on \benchname{} 
clearly highlights evolution of distributions and captures different properties 
than standard benchmarks.

\textbf{Quantitative analysis on \cyfcc{} {} {}}
We analyze \cyfcc{} using off-the-shelf sentence and image encoders.
For off-the-shelf sentence embedder, we used an existing sentence transformer from Hugging Face~\citep{reimers-2019-sentence-bert}. 
For the image encoder, we use a CLIP pretrained ViT-B-16 model~\citep{radford2021learning, ilharco2021openclip}. 

\begin{figure*}[h]
    \centering
    \begin{subfigure}[b]{0.8\linewidth}
    \includegraphics[width=\linewidth]{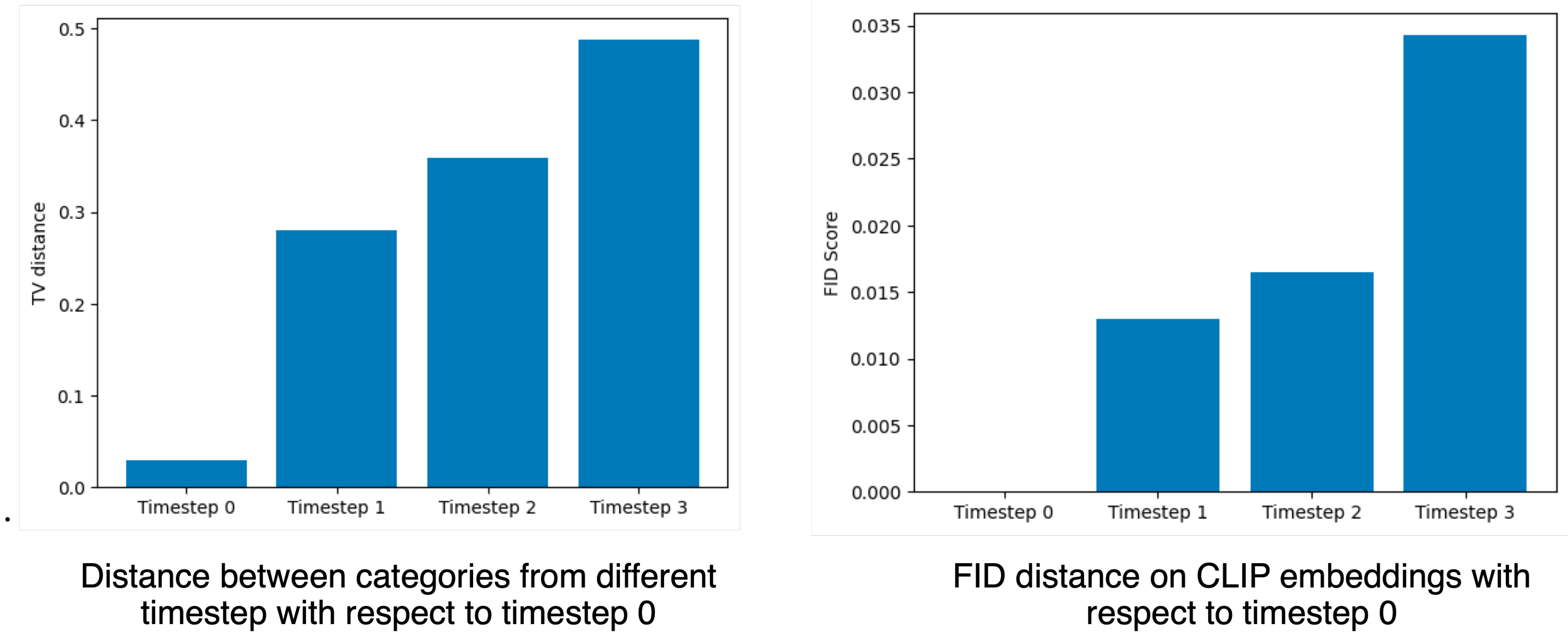} 
    \caption{\cyfcc{}}
    \end{subfigure}

     \begin{subfigure}[b]{0.8\linewidth}
    \includegraphics[width=\linewidth]{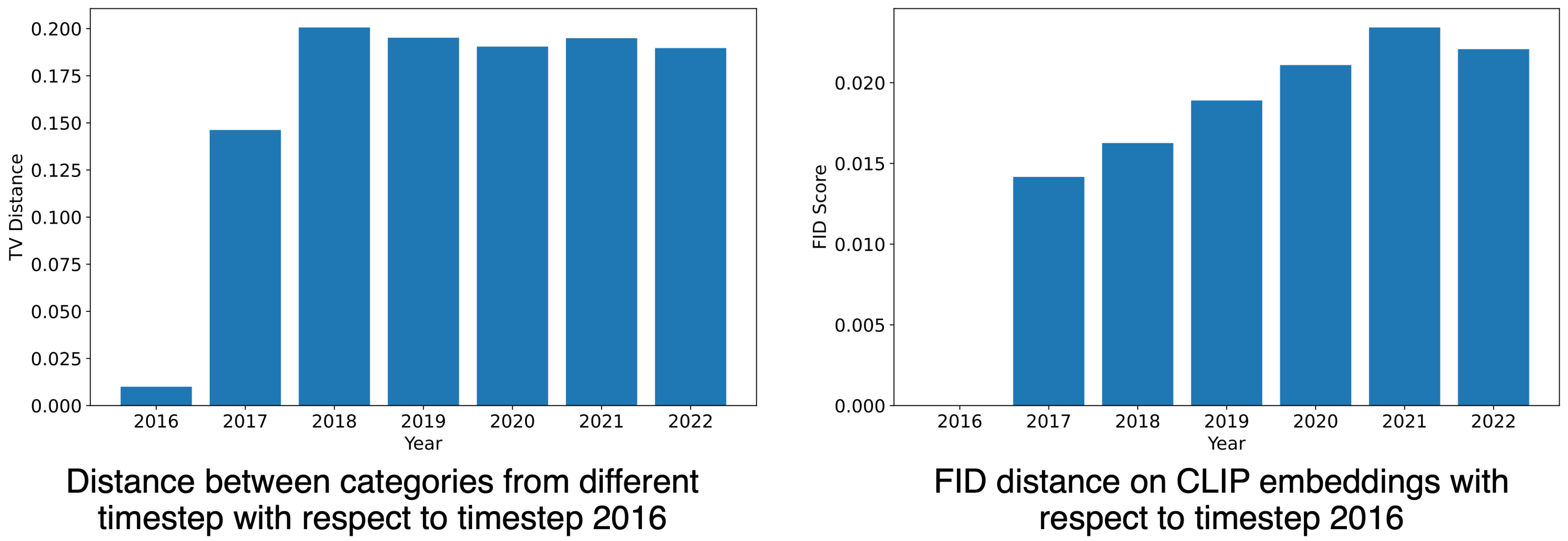} 
    \caption{\benchname{} (M)}
    \end{subfigure}
    \caption{Distribution shift results. Analysis on \cyfcc{} and \benchname{} (M) using off-the-shelf sentence and image encoders.  We first embed images from different time steps with an  OpenAI CLIP encoder and then compute Frechet Inception Distance (FID; \citet{Seitzer2020FID}).  As time progresses, we observe that FID distance increases with respect to data from first time step. Similarly TV distance over categorical distribution on Wordnet Noun synsets also increases with time when compared to categorical distribution on first timestep. }
    \label{fig:yfcc_dist_shift}
\end{figure*}

We first embed images from different time steps with an 
OpenAI CLIP encoder and then compute Frechet Inception Distance (FID; \citet{Seitzer2020FID}). 
As time progresses, we observe that FID distance increases 
with respect to data from first time step (\figref{fig:yfcc_dist_shift}). 
Similarly, we use the pretrained sentence transformer to 
extract top-5 categories from Wordnet Nouns  
for each caption. We then obtain a distribution over these Nouns for each time step. 
We observe that the TV distance 
over the distribution of WordNet nouns evolves over time
when compared to data from the first time step.

\begin{figure*}[h]
    \centering
    \begin{subfigure}[b]{\linewidth}
    \includegraphics[width=0.48\linewidth]{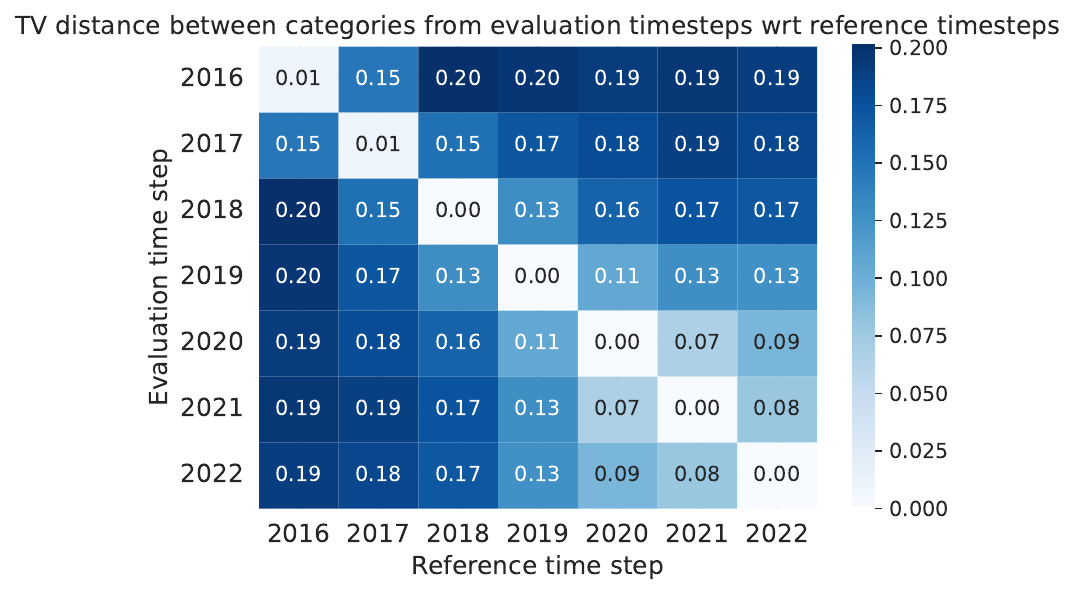}
    \includegraphics[width=0.45\linewidth]{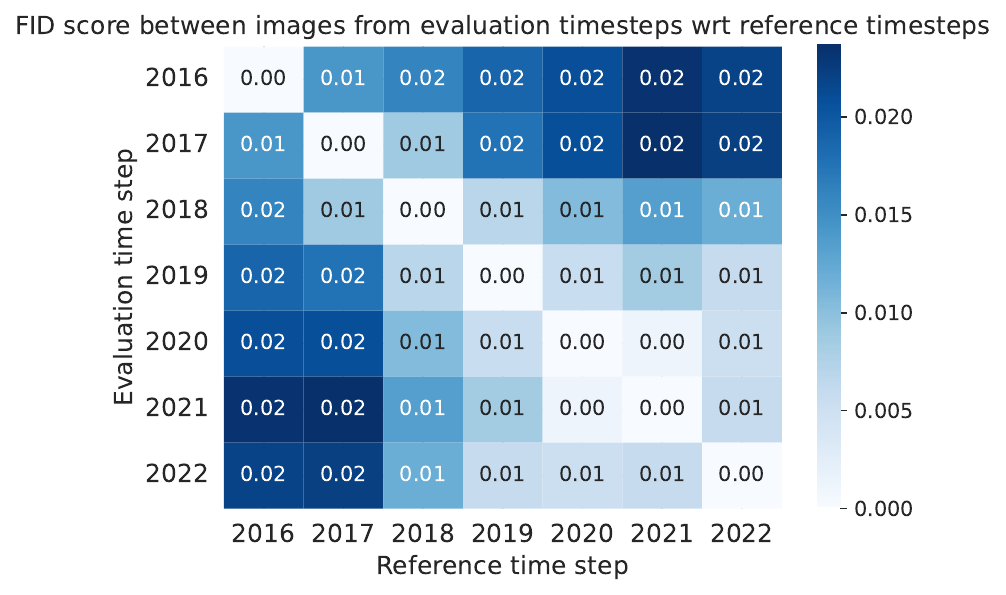} 
    \end{subfigure}

    \caption{\update{Distribution shift analysis on \benchname{} (M) using off-the-shelf sentence and image encoders. We first embed images from different time steps with an  OpenAI CLIP encoder and then compute Frechet Inception Distance (FID; \citet{Seitzer2020FID}).  As time progresses, we observe that FID distance increases with respect to data from first time step. Similarly TV distance over categorical distribution on Wordnet Noun synsets also increases with time when compared to categorical distribution on first timestep.} }
    \label{fig:datacomp_dist_shift_heatmap}
\end{figure*}

\subsection{Creation Pipiline for \benchname{}} \label{app:tic_datacomp_construction}

We collect timestamps for the 
CommonPool dataset introduced in \datacomp{}. 
We repeat the crawling process described in \citet{gadre2023datacomp}
to download WARC files from Common Crawl. 
In particular, we follow the same multistep process which involved: (i) parsing URLs and alt-text from Common Crawl dumps and downloading these images; (ii) tagging images with meta data and id of the common crawl batch; and (iii) conducting evaluation set duplication and safety content filtering. 
After downloading the WARC files, we perform a join with the datacomp 
12.8B examples. During this join, we lost approximately 0.1B of examples that are no longer available online. 
Moreover, while performing this join, we only retain examples 
with their first occurrence.
This is done before running any de-duplication on image-text pairs for 
exact matches as done in \citet{gadre2023datacomp}. 

The source of \datacomp{} is Common Crawl, which periodically 
releases web-crawled data snapshots, typically on a monthly basis since 2014 
with new and updated webpages.
This process provides timestamps at the granularity of 
months, spanning years 2014--2022.

\begin{figure}[t]
    \centering
    \includegraphics[width=0.6\linewidth]{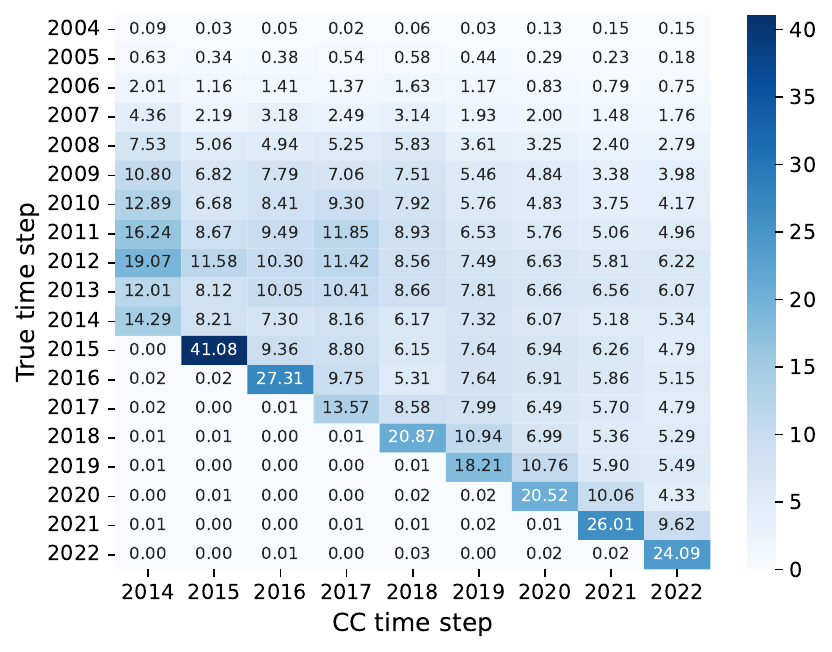}
    \caption{Comparison of Common Crawl assigned timestamp and true timestamp on a subset of 10k examples containing image-text pairs from Flickr. We observe a clear trend where CC timestamps correlate with true timestamps. }
    \label{fig:flickr_stats}
\end{figure}

We note that while this augmented time information may contain some noise, on 
average, we find it to be a reasonably accurate proxy for the upload time of 
web pages. To perform an initial check, we note that our data contains images from flickr which 
provides an API to query for true upload timestamp. So we extract 10k examples from our benchmark \benchname{}
and query Flickr for their true timestamp. \figref{fig:flickr_stats} summarizes true timestamps 
with timestamps extracted from CC. 

\section{Additional Experimental Details}

\subsection{Additional details on ImageNet IID split continual learning experiment} \label{app:imagenet_exp} 

With ImageNet data, we consider 2, 4 and 8 splits including the full dataset. 
This design is inspired by \citet{ash2020warm}. We consider ViT-B/16 
architecture trained for 300 epochs on full data and split the iterations 
corresponding to 300 epochs equally among k splits when training sequentially. 
We keep all other hyperparameters, such as learning rate, optimizer, and batch size, set to the standard values typically employed for training ViT-B/16 on the ImageNet dataset~\citep{dosovitskiy2020image}. We also employ $\ell_2$ regularization and augmentation on ImageNet training data. 
We evaluate the models on IID ImageNet test set.

\update{Our Imagenet experiments were primarily inspired by the ``loss of plasticity'' phenomenon described in \citet{ash2020warm}. Their study demonstrates that models sequentially trained on two splits of CIFAR-10 data (initially on 50\%, followed by 100\% of data) exhibit poorer generalization compared to models trained from scratch on the entire dataset. Since we do not observe this behavior for continual training of CLIP, we investigated the existence of such behaviors on up to 8 splits of Imagenet. Our findings reveal that the simple cumulative baseline (with no extra budget) remains competitively close to the Oracle model (that benefits from using the full compute budget on the entire pooled training data from the beginning).}

\update{Prior works~\citep{prabhu2023computationally, hu2021one} performed continual learning experiments on Imagenet to compare different methods and highlight the effectiveness of continual training on synthetic continual learning setups derived from ImageNet. While these papers include results with an \Oracle{} method, differences in the settings considered in these studies limit direct comparisons. }

\update{In particular, 
we show the performance gap of less than 1\% in the same setup used otherwise in the paper when using SOTA training procedures achieving 81\% validation performance. Comparitively the referenced \citet{hu2021one} does not show whether the 65\% to 77\% performance gap in their Table 1 can be bridged by increasing the compute for their method. Instead, authors show that if they restrict the compute for Oracle in Table 2, the Oracle performance drops to 68\% (with $\approx 3\%$ gap). 
}

\update{Moreover, in \citet{prabhu2023computationally}, authors perform experiments on DI-Imagenet-2k where they start with an initial memory of Imagenet-1k 1.2 M samples and sequentially observe data for the same classes 1k classes from Imagenet-21k pool. This makes comparing streaming accuracy (or  Imagenet-1k accuracy) for different methods incomparable with our setup (with a gap of over 7\% in streaming accuracy even at step 8 as compared to less than 1\% in our setup). 
}

\subsection{Training and Hyperparameter Details} \label{app:hyperparam_details}

We create a common experimental setup by fixing the training procedure for sequential runs. 
Unless specified otherwise, we closely follow the CLIP training recipe proposed in \citep{ilharco2021openclip, radford2021learning} where we train
models with a contrastive objective over images and captions. Given a set of image-text pairs, we train an image encoder and a text encoder such that the similarity between the
representations of images and their corresponding text is maximized relative to unaligned pairs.  Only \LwF\ deviates from this standard training procedure. 
For each benchmark, we pick Vision Transformers (ViTs) as the image encoder, in particular, we fix the model architecture to ViT-B/16~\citep{dosovitskiy2021an}.
We fix the Adam optimizer and its hyperparameters to values suggested in \citep{ilharco2021openclip}. 

We primarily ablate over only two things: maximum learning rate with cosine learning schedule and warm up iterations for sequential training. 
For choosing other hyperparameters,
we follow the OpenCLIP library~\citep{ilharco2021openclip}.

\subsection{Replay sizes with \Exponential{} and \Equal{} strategies} \label{app:replay_sizes}

We default to using 2D size of data where D represents incoming data size from new time step.
As described in the main text, for -\Exponential{}, we reduce the buffer size by half of what we used at old time step and use rest of the half as data from previous time step. \appref{app:dataset_stats} lists the dataset sizes for each benchmark which dictate the exact buffer sizes.

\section{\update{Results with Other Continual Learning Methods}}
\label{app:other_methods}

\subsection{\update{Results with EWC Method}}

\update{As proposed in the original work \citet{kirkpatrick2017overcoming}, we implement EWC method where we optimize the following loss: 
\begin{align*}
    \calL_{EWC} (\theta) = \calL (\theta) + \sum_{i} \frac{\lambda_{EWC}}{2} F_i (\theta_i - \theta_{t-1, i})^2 \,,
\end{align*}
where  $\calL (\theta)$ is the standard contrastive loss on data from time step $t$, $F_i$ is the $i$-th diagonal entry of the fisher information matrix, and $\theta_{t-1}$ are the frozen parameters from previous time step. We perform experiments with different values of $\lambda_{EWC} \in \{1,10,100,400\}$ (see \tabref{tab:ewc_results}).}

\begin{table}[H]
    \centering
    \caption{\update{\textbf{Zero shot performance on our time-continual benchmarks with EWC.} 
    $^*$ and $^{**}$ denote methods that violate the compute budget.
    For static tasks, we tabulate accuracy of the models obtained on the final timestamp. 
    For dynamic tasks, we tabulate forward/backward transfer 
    and ID performance on retrieval tasks (\secref{subsec:exp_protocol}). 
    We observe that EWC performs worse than \Sequential{}, \Patching{} and \LwF{}.   
    }
    }
    \vspace*{-5pt}
\resizebox{1.0\textwidth}{!}{%
\begin{tabular}{lcccccccccc}
\toprule[1.2pt]
\multirow{3}{*}{\textbf{Benchmark}} & \multirow{3}{*}{\textbf{Method}} &   \multirow{3}{*}{\parbox{4em}{\centering \textbf{Compute} (MACs)}} & \multicolumn{4}{c}{\textbf{Static Tasks}} &  {} & \multicolumn{3}{c}{\textbf{Dynamic Retrieval Tasks}} \\[3pt]
{} & {} & {} & \multirow{2}{*}{\parbox{4em}{ImageNet}} & \multirow{2}{*}{\parbox{4em}{ImageNet dist. shift}} & \multirow{2}{*}{\parbox{4em}{Flickr30k}} & \multirow{2}{*}{\parbox{5.5em}{\centering Average over 28 datasets}} & {} & \multirow{2}{*}{\parbox{4em}{\centering Backward Transfer}} & \multirow{2}{*}{\parbox{4em}{\centering ID Performance}} & \multirow{2}{*}{\parbox{4em}{\centering Forward Transfer}}  \\ 
{} & {} & \\
\midrule 
\multirow{2}{*}{\textbf{\benchname} (M)}  
& \Sequential{} &  $3.0 \times 10^{18}$  & $19.2$ & $16.4$ & $16.4$ & $15.0$ &  & $25.7$ & $26.4$ & $14.9$ \\
& \Patching{} &  $3.0 \times 10^{18}$  & $19.3$ & $16.8$ & $18.5$ & $14.7$ &  & $26.9$ & $25.4$ & $14.5$ \\
\rowcolor{gray!10}  & \LwF{}$^{*}$ &  $3.8 \times 10^{18}$ &  $19.2$ & $16.5$ & $17.7$ & $14.3$ &  & $25.6$ & $26.6$ & $14.9$ \\ 
\rowcolor{gray!10} & EWC ($\lambda_{EWC}$ = 1)$^{*}$ &  $3.6 \times 10^{18}$  & $18.7$ & $16.3$ & $16.2$ & $15.1$ &  & $25.5$ & $26.4$ & $14.8$ \\
\rowcolor{gray!10} & EWC ($\lambda_{EWC}$ = 10)$^{*}$ &  $3.6 \times 10^{18}$  & $18.1$ & $15.8$ & $16.8$ & $14.7$ &  & $24.8$ & $25.7$ & $14.4$ \\
\rowcolor{gray!10} & EWC ($\lambda_{EWC}$ = 100)$^{*}$ &  $3.6 \times 10^{18}$  & $17.6$ & $15.4$ & $16.3$ & $14.8$ &  & $24.4$ & $25.4$ & $14.3$\\
\rowcolor{gray!10} & EWC ($\lambda_{EWC}$ = 400)$^{*}$ &  $3.6 \times 10^{18}$  & $17.0$ & $15.0$ & $16.4$ & $14.3$ &  & $24.1$ & $24.9$ & $14.0$ \\
\bottomrule[1.2pt]
\end{tabular}
}    
    \label{tab:ewc_results}
    \vspace*{-8pt}
\end{table}

\subsection{\update{Results with Oversampling + Counting Based Sampling Method}}

\update{In this section, we perform ablation on  \Cumulative{}-\Equal{}. In particular, we made the following two modifications: (i) \emph{Count based sampling}: Instead of random sampling, we implemented the count-based subsampling that prioritizes not/less used examples; (ii) \emph{Oversampling}: We oversampled data from old timesteps with ratio inversely proportional to the ratio of examples, i.e., if the old data is of size D/2 and the new data is of size D, then we upsample old data with 2:1 ratio.}

\update{However, we observe that this method doesn’t improve performance over \Cumulative{}-\Equal{} and in fact hurts the performance slightly (see \tabref{tab:oversampling_results}). We hypothesize that this can be due to a decreasing marginal utility of labeled data as highlighted in ~\citet{cui2019class}. Their work argues that due to information overlap among data, as the number of samples increases, the marginal benefit a model can extract from the data diminishes. As a result, \citet{cui2019class} proposed using of ``effective sample size'' instead of the actual number of samples to obtain the ratio used to perform re-sampling or re-weighting. In particular, the expression of ``effective sample size'' is given by $E_n = \frac{1-\beta^n}{1-\beta}$ where $n$ is the original sample size and $\beta$ is a hyperparameter that \citet{cui2019class} selects from $\beta \in \{0.9, 0.99, 0.999, 0.9999\}$.}

\update{For different time steps, we leverage this expression of $E_n$ to calculate the effective number of samples. 
In our settings (even at small scales), our datasets contain an order of 100k image-text pairs even after subsampling data from old time step. For example, with -\Equal{} baseline, when training on the last time step (i.e., 2022), the smallest dataset (i.e., 2016) is of approximately 400k samples.
Plugging in the expression for effective sample size from \citet{cui2019class}, we observe that for all 
$\beta \in (0, 0.99999)$, the ratio of effective sample sizes for different time steps remains close to 1. This may highlight why our naive over-sampling strategy doesn’t improve over no-oversampling. }

\begin{table}[H]
    \centering
    \caption{\update{\textbf{Zero shot performance on our time-continual benchmarks with oversampling and counting-based sampling.} 
    $^*$ and $^{**}$ denote methods that violate the compute budget.
    For static tasks, we tabulate accuracy of the models obtained on the final timestamp. 
    For dynamic tasks, we tabulate forward/backward transfer 
    and ID performance on retrieval tasks (\secref{subsec:exp_protocol}). 
    }
    }
    \vspace*{-5pt}
\resizebox{1.0\textwidth}{!}{%
\begin{tabular}{lcccccccccc}
\toprule[1.2pt]
\multirow{3}{*}{\textbf{Benchmark}} & \multirow{3}{*}{\textbf{Method}} &   \multirow{3}{*}{\parbox{4em}{\centering \textbf{Compute} (MACs)}} & \multicolumn{4}{c}{\textbf{Static Tasks}} &  {} & \multicolumn{3}{c}{\textbf{Dynamic Retrieval Tasks}} \\[3pt]
{} & {} & {} & \multirow{2}{*}{\parbox{4em}{ImageNet}} & \multirow{2}{*}{\parbox{4em}{ImageNet dist. shift}} & \multirow{2}{*}{\parbox{4em}{Flickr30k}} & \multirow{2}{*}{\parbox{5.5em}{\centering Average over 28 datasets}} & {} & \multirow{2}{*}{\parbox{4em}{\centering Backward Transfer}} & \multirow{2}{*}{\parbox{4em}{\centering ID Performance}} & \multirow{2}{*}{\parbox{4em}{\centering Forward Transfer}}  \\ 
{} & {} & \\
\midrule 
\multirow{2}{*}{\textbf{\benchname} (M)}  
& \Sequential{} &  $3.0 \times 10^{18}$  & $19.2$ & $16.4$ & $16.4$ & $15.0$ &  & $25.7$ & $26.4$ & $14.9$ \\
& \Cumulative{}-\Equal{} (Counts + OS) &  $3.0 \times 10^{18}$  & $18.1$ & $15.3$ & $14.3$ & $16.5$ &  & $28.9$ & $23.7$ & $14.2$ \\
& \Cumulative{}-\Equal{} &  $3.0 \times 10^{18}$  & $22.1$ & $18.4$ & $19.2$ & $17.1$ &  & $31.8$ & $26.8$ & $15.1$\\
\bottomrule[1.2pt]
\end{tabular}
}    
    \label{tab:oversampling_results}
    \vspace*{-8pt}
\end{table}

\section{\update{Results With New Evaluation Metrics on Dynamic Tasks}} \label{app:updated_metrics}

\update{Recall, $T$ represent the number of time 
steps for which we have data.
For each training 
method, we generate a total of $T$ models, each corresponding to the end of 
training at a particular time step. 
For each model and a dynamic evaluation task, we 
obtain $T$ performance values. We represent these values using the performance 
matrix $\calE$, where each entry $\calE_{i,j}$ signifies the performance of the 
model obtained after observing training data at time step $i$ when evaluated on 
a dataset from time step $j$. 
Defining backward metrics as in \secref{subsec:dynamic_eval_datasets} involves averaging the entries in the upper and lower diagonal of our performance matrix $\calE$, i.e., it was calculated as the average of time steps before each training step (i.e., the lower triangular of $\calE$), i.e., $\frac{\sum_{i \ge j } \calE_{ij}}{(T(T-1))/2}$. This backward transfer metric has been used in prior works \citet{lin2021clear}. 
However, this approach inadvertently resulted in the backward transfer metric being influenced by later evaluation time steps 
resulting in backward transfer performance numbers slightly larger than ID performance. } 

\update{To address this issue, we've revised our metric calculation method to metric as in \citet{diaz2018don}. Now, we normalize the data in each row, which corresponds to evaluation time steps by subtracting the ID performance. This adjustment ensures a more balanced and accurate representation across all training time steps. In particular, our updated forward and backward transfer metrics can be summarized as:}

\update{\begin{itemize}[leftmargin=*]
\item \emph{Backward transfer}: Let $\calB_i$ denote the average performance on evaluation tasks before time $i$, then we define backward transfer as average of $\calB_i$ across each training step, i.e., $\sum_{i=2}^T \frac{\sum_{i\ge j} {\calE_{ij} - \calE_{ii}} }{T(T-1)/2}$%
\item \emph{Forward transfer}: Let $\calF_i$ denote the average performance on evaluation tasks after time $i$, then we define forward transfer as average of $\calF_i$ across each training step, i.e., $\sum_{i=1}^{T-1} \frac{\sum_{i\le j} \calE_{ij} - \calE_{ii} }{T(T-1)/2}$
\end{itemize}}

\begin{table}[t]
    \centering
    \caption{\update{\textbf{Zero shot performance on our time-continual benchmarks.} 
    $^*$ and $^{**}$ denote methods that violate the compute budget.
    For dynamic tasks, we tabulate forward/backward transfer 
    and ID performance on retrieval tasks with updated metrics as defined in \appref{app:updated_metrics}. 
    }}
    \vspace*{-5pt}
\resizebox{.8\textwidth}{!}{%
\begin{tabular}{lcccccccccc}
\toprule[1.2pt]
\multirow{4}{*}{\textbf{Benchmark}} & \multirow{4}{*}{\textbf{Method}} &   \multirow{4}{*}{\parbox{4em}{\centering \textbf{Compute} (MACs)}}  & \multicolumn{3}{c}{\textbf{Dynamic Retrieval Tasks}} \\[3pt]
{} & {} & {} & \multirow{3}{*}{\parbox{4em}{\centering Backward Transfer}} & \multirow{3}{*}{\parbox{4em}{\centering ID Performance}} & \multirow{3}{*}{\parbox{4em}{\centering Forward Transfer}} & \multirow{3}{*}{\parbox{5em}{\centering \update{Relative Backward Transfer}}}  & \multirow{3}{*}{\parbox{4em}{\centering \update{Relative Forward Transfer}}} \\ 
{} & {} & \\
{} & {} & \\
\midrule
\multirow{8}{*}{\textbf{\cyfcc{}}} & \Restart{} & $3.4 \times 10^{18}$  & $13.2$ & $41.4$ & $18.6$ &  $-29.8$ & $-21.2$\\
& \Sequential{} & $3.4 \times 10^{18}$ & $42.2$ & $48.4$ & $23.7$ & $-9.5$ & $-21.5$\\
& \Patching{} & $3.4 \times 10^{18}$  & $44.7$ & $53.4$ & $24.5$ & $-15.6$ & $-22.0$ \\
& \Cumulative{}-\Exponential{} &   $3.4 \times 10^{18}$  & $60.4$ & $60.1$ & $27.1$ & $-9.8$ & $-23.0$ \\
& \Cumulative{}-\Equal &  $3.4 \times 10^{18}$   & $60.4$ & $60.4$ & $27.1$ & $-10.3$ & $-23.0$ \\
&  \Cumulative{}-\Full{} &   $3.4 \times 10^{18}$  & $\mathbf{66.4}$ & $\mathbf{60.2}$ & $\mathbf{27.6}$ & $-4.1$ & $-22.4$\\
\rowcolor{gray!10} & \LwF{}$^{*}$ & $4.1 \times 10^{18}$   & $36.6$ & $56.0$ & $23.2$  & $-27.4$ & $-24.9$ \\ 
\rowcolor{gray!10} &  \Cumulative{}-\Full{}$^{*}$ &   $3.6\times 10^{18}$ & $\mathbf{66.8}$ & $\mathbf{60.3}$ & $\mathbf{27.6}$ & $-3.9$ & $-22.4$\\
\rowcolor{gray!10} & \Oracle{}$^{**}$ &  $8.5\times 10^{18}$ & $\mathbf{66.1}$ & $\mathbf{61.8}$ & $\mathbf{26.9}$ & $-6.6$ & $-24.0$\\
\midrule 
\multirow{8}{*}{\textbf{\credcaps{}}} 
& \Restart{} & $3.4 \times 10^{18}$ & $21.3$ & $25.4$ & $22.4$ & $-4.5$ & $-2.7$ \\
& \Sequential{} &  $3.4 \times 10^{18}$ & $33.0$ & $33.6$ & $27.5$  &  $-3.8$ & $-3.0$\\
& \Patching{} &  $3.4 \times 10^{18}$ & $34.8$ & $34.8$ & $27.8$& $-3.9$ & $-3.0$ \\
& \Cumulative{}-\Exponential{} &  $3.4 \times 10^{18}$  & $44.5$ & $42.0$ & $32.6$ & $-3.0$ & $-4.0$ \\
& \Cumulative{}-\Equal &  $3.4 \times 10^{18}$ & $44.4$ & $42.0$ & $32.6$& $-3.0$ & $-4.0$ \\
&  \Cumulative{}-\Full{} & $3.4 \times 10^{18}$ & $\mathbf{48.9}$ & $\mathbf{43.2}$ & $\mathbf{33.4}$ & $-0.6$ & $-3.5$\\
\rowcolor{gray!10} & \LwF{}$^{*}$ &  $4.1 \times 10^{18}$ & $35.4$ & $36.0$ & $28.4$ & $-4.6$ & $-3.7$\\
\rowcolor{gray!10} &  \Cumulative{}-\Full{}$^{*}$ &  $3.6\times 10^{18}$ & $\mathbf{49.0}$ & $\mathbf{43.4}$ & $\mathbf{33.4}$ & $-1.0$ & $-3.5$\\
\rowcolor{gray!10} & \Oracle{}$^{**}$ &  $8.5\times 10^{18}$  & $\mathbf{48.5}$ & $\mathbf{43.1}$ & $\mathbf{33.4}$ & $-1.0$ & $-3.4$\\

\midrule 
\multirow{6}{*}{\textbf{\benchname} (M)} 
& \Sequential{} &  $3.0 \times 10^{18}$  & $25.7$ & $26.4$ & $14.9$ & $-4.7$ & $-7.6$ \\
& \Patching{} &  $3.0 \times 10^{18}$ & $26.9$ & $25.4$ & $14.5$ & $-1.9$ & $-7.4$ \\
& \Cumulative{}-\Exponential{} &  $3.0 \times 10^{18}$ & $31.7$ & $27.1$ & $\mathbf{15.2}$  & $0.3$ & $-7.6$ \\
& \Cumulative{}-\Equal &  $3.0 \times 10^{18}$  & $31.8$ & $26.8$ & $15.1$ & $0.9$ & $-7.6$ \\
&  \Cumulative{}-\Full{} &   $3.0 \times 10^{18}$& $33.8$ & $26.4$ & $15.1$ &$3.5$ & $-7.3$\\
\rowcolor{gray!10}  & \LwF{}$^{*}$ &  $3.8 \times 10^{18}$ & $25.6$ & $26.6$ & $14.9$ & $-4.8$ & $-8.0$\\
\rowcolor{gray!10} &  \Cumulative{}-\Full{}$^{*}$ &  $3.9 \times 10^{18}$ & $\mathbf{36.7}$ & $\mathbf{28.3}$ & $\mathbf{15.5}$ & $3.0$ & $-7.3$ \\
\rowcolor{gray!10} & \Oracle{}$^{**}$ &  $1.2 \times 10^{19}$  & $34.9$ & $27.8$ & $\mathbf{15.6}$ & $2.5$ & $-7.7$\\
\midrule 
\multirow{5}{*}{\textbf{\benchname} (L)} 
& \Sequential{} & $2.7 \times 10^{19}$ & $52.6$ & $\mathbf{58.4}$ & $41.1$ & $-8.7$ & $-14.4$ \\
& \Patching{} & $2.7 \times 10^{19}$ & $55.2$ & $57.5$ & $40.9$ & $-4.9$ & $-13.9$\\
& \Cumulative{}-\Exponential{} & $2.7 \times 10^{19}$ &  $60.4$ & $\mathbf{58.4}$ & $\mathbf{41.4}$ & $-1.1$ & $-13.8$\\
& \Cumulative{}-\Equal & $2.7 \times 10^{19}$ &  $60.9$ & $\mathbf{58.2}$ & $\mathbf{41.4}$ & $-0.3$ & $-13.8$\\
&  \Cumulative{}-\Full{} & $2.7 \times 10^{19}$   & $62.1$ & $57.3$ & $41.2$ & $2.2$ & $-13.5$\\
\rowcolor{gray!10} &  \Cumulative{}-\Full{}$^{*}$ &  $4.1\times 10^{19}$ & $63.0$ & $57.8$ & $41.2$ & $2.1$ & $-13.5$ \\
\rowcolor{gray!10}  &  \Oracle{}$^{**}$ & $1.1 \times 10^{20}$ & $\mathbf{64.3}$ & $\mathbf{58.6}$ & $\mathbf{41.8}$ &$2.2$ & $-13.3$\\
\midrule 
\multirow{3}{*}{\textbf{\benchname} (XL)} & \Sequential{} &  $2.7\times 10^{20}$ & $63.1$ & $68.9$ & $56.8$ & $-5.6$ & $-12.3$\\
&  \Cumulative{}-\Full{} &  $2.7\times 10^{20}$ & $\mathbf{70.7}$ & $\mathbf{68.5}$ & $\mathbf{57.1}$ & $2.5$ & $-11.7$
\\
\rowcolor{gray!10} &  \Cumulative{}-\Full{}$^{*}$ &  $3.5\times 10^{20}$ & $\mathbf{71.0}$ & $\mathbf{68.6}$ & $\mathbf{57.1}$ & $2.5$ & $-11.7$
\\
\bottomrule[1.2pt]
\end{tabular}
}    
    \label{tab:main_results_updated}
    \vspace*{-8pt}
\end{table}

\end{document}